%% file: neurips_2026.tex
\documentclass{article}

\PassOptionsToPackage{numbers, compress}{natbib}
\newif\ifpreprint
\preprinttrue

\ifpreprint
\usepackage[preprint]{neurips_2026}
\else
\usepackage{neurips_2026}
\fi

\usepackage[utf8]{inputenc} % allow utf-8 input
\usepackage[T1]{fontenc}    % use 8-bit T1 fonts
\usepackage{hyperref}       % hyperlinks
\usepackage{url}            % simple URL typesetting
\usepackage{booktabs}       % professional-quality tables
\usepackage{amsfonts}       % blackboard math symbols
\usepackage{nicefrac}       % compact symbols for 1/2, etc.
\usepackage{microtype}      % microtypography
\usepackage{xcolor}         % colors
\usepackage{graphicx}
\usepackage{amsmath}
\usepackage{subcaption}
\usepackage{wrapfig,lipsum}
\usepackage{pdfpages}
\usepackage{algorithm}
\usepackage{algpseudocode}

\usepackage[
  separate-uncertainty = true,
  multi-part-units = repeat
]{siunitx}

\algnewcommand{\singlecomment}[1]{\State \textcolor{gray}{// #1}}

\newcommand{\ourmethod}{POP}

\title{Bootstrapping Post-training Signals for Open-ended Tasks via Rubric-based Self-play on Pre-training Text}

\author{%
  Chengyu Huang\quad \textbf{Sheng-Yen Chou}\quad \textbf{Zhengxin Zhang}\quad \textbf{Claire Cardie}\\
  Department of Computer Science, Cornell University \\
  \texttt{\{ch2263, sc3379, zz865, ctc9\}@cornell.edu} \\
}

\begin{document}

\maketitle

\begin{abstract}

Self-play has recently emerged as a promising paradigm for post-training Large Language Models (LLMs). In self-play, the target LLM creates the task input (e.g., a question), which it then addresses itself by producing a task output (e.g., an answer). A reward model evaluates the output, and the rewards are used to train the LLM, typically via Reinforcement Learning (RL). A key benefit of self-play for post-training LLMs is its minimal supervision costs: self-play avoids the need for high-quality input-output pairs traditionally constructed by humans or expensive proprietary models. Existing work, however, explores self-play only for verifiable tasks, such as math and coding, for which objective ground truth is available and easily checkable. In this paper, we seek to extend self-play to more realistic open-ended tasks. We propose \ourmethod, a self-play framework that uses the same LLM to synthesize evaluation rubrics along with each input-output pair. The rubric is used to evaluate outputs and train the model. Crucially, we ground the framework on a content-rich pretraining corpus to (1) enable an exploitable generation-verification gap and reduce reward hacking, and (2) prevent mode collapse. On Qwen-2.5-7B, \ourmethod~increases performance of both the pretrained base model and instruction-tuned model on multiple tasks ranging from long-form healthcare QA to creative writing and instruction following.
\ifpreprint
GitHub:  \url{https://github.com/HCY123902/POP}.
\fi

\end{abstract}

\input{Introduction}

\input{related_work}

\input{method}

\input{setup}

\input{results}

\vspace{-2pt}
\section{Limitations and Future Directions}
\label{sec:limitations_and_future_directions}
\vspace{-2pt}
Future work might investigate raising the level of automation by one more level. That is, we can ask $\pi_{ref}$ to select tasks suitable for the pre-training text and generate the query synthesis prompt. 

Cost and compute constraints limited our ability to create larger synthesized datasets. Since the design of \ourmethod~is general and not specific to any task, we believe synthesizing datasets larger in magnitude on a general pre-training corpus (e.g., OpenWebText) is a promising direction to enable cheap, effective, and general post-training.

Our pipeline also requires a strong enough reference model. Otherwise, the model may not be able to follow instructions,  synthesize queries, responses, rubrics, or conduct evaluations. Investigating the minimum level of model competency for \ourmethod~to be functional should be investigated in the future.
We discuss other limitations in Appendix~\ref{app:limitations_and_future_directions}.

\vspace{-5pt}
\section{Conclusion}
\vspace{-5pt}
We introduce \ourmethod, a rubric-based self-play post-training framework on open-ended tasks. \ourmethod~synthesizes query-response-rubric triples and grades each response using the rubrics. To mitigate reward hacking, we give privileged access to pretraining text during rubric generation and only take the responses with the highest and lowest scores for DPO training. Experiment results suggest that \ourmethod~bootstraps effective training signals across different tasks. The design of \ourmethod~is meant to be generalizable, and
we hope it highlights a direction toward cheap and effective post-training for LLMs.

\bibliography{custom}
\bibliographystyle{plainnat}

\medskip

%%%%%%%%%%%%%%%%%%%%%%%%%%%%%%%%%%%%%%%%%%%%%%%%%%%%%%%%%%%%

\newpage
\appendix

\input{appendix}

%%%%%%%%%%%%%%%%%%%%%%%%%%%%%%%%%%%%%%%%%%%%%%%%%%%%%%%%%%%%

\ifpreprint
\else
\clearpage
\input{checklist.tex}
\fi

\end{document}

%% file: introduction.tex
\section{Introduction}

Recent Large Language Models (LLMs) have become increasingly capable of handling complex tasks, from math problem solving \citep{gdm2025} to agentic workflows \citep{agentsurvey2025}. However, continual post-training of LLMs 
% to achieve high performance on desired tasks 
requires high-quality data from humans or stronger LLMs, which is often expensive or unavailable. In fact, the cost and scarcity of high-quality training data is becoming a bottleneck \citep{villalobos2024}.

To this end, self-play has emerged as a promising direction for post-training LLMs with minimal external supervision. In essence, in each iteration of self-play (i) the LLM synthesizes task input (e.g., asks a question); (ii) the same LLM responds with task output (e.g., gives an answer); (iii) the output (and optionally the input) is then scored by a task-dependent reward model that can take various forms (a rule-based verifier, neural network, etc.); and (iv) the reward is used to update the LLM, to make it produce both better task-specific output (e.g., answer better) and (optionally) better task input (e.g., ask better questions). In contrast to standard RL and alternative self-improvement methods \citep{huang2023, yuan2024}, self-play avoids the need for supervision not only of the task output, but also the task input.

Self-play for post-training has thus far only been explored in "verifiable" domains such as math \cite{huang2025rzero, liu2025, zhang2025, chen2025} and coding \cite{absolutezero2025, zhou2025self}, where verifying the correctness of task output is easy. Extending self-play to more realistic open-ended tasks is important, but challenging. Recent work \cite{gunjal2025, viswanathan2025, bi2025, zhou2025, he2025, rezaei2025, huang2025rlra, shao2025}, however, shows that rubric-based rewards can provide reliable signals for open-ended tasks. Inspired by this, we propose a rubric-based self-play framework that post-trains LLMs for open domains. In particular, after the model generates responses to a query, we use the \textit{same} model to generate an evaluation rubric specific to that query, and then grade each response based on the rubric.

The use of the same model to synthesize queries, responses, rubrics, and rewards engenders several risks: (1) query diversity might be limited, causing mode collapse; (2) response quality might be poor, limiting positive training signals and learning effectiveness; (3) the generation-verification gap \cite{gap2025} --- ensuring that response verification is easier than response generation --- might be insufficient, resulting in reward hacking. We address (1) by grounding the entire framework on diverse text sampled from a content-rich pre-training corpus. We ask the model to synthesize queries that draw from the text, and, in tasks for which ground truth exists, to construct queries whose answers are derivable from the text. We address (2) by using Direct Preference Optimization (DPO) \cite{dpo2023}, to train the LLM via contrastive signals from the response pairs. We use two methods to mitigate (3). First, we give privileged access to the pre-training text to the model when it evaluates its generated responses. Second, we select only the highest and lowest scored responses for DPO training.

\begin{figure}
    \centering
    \includegraphics[width=\linewidth]{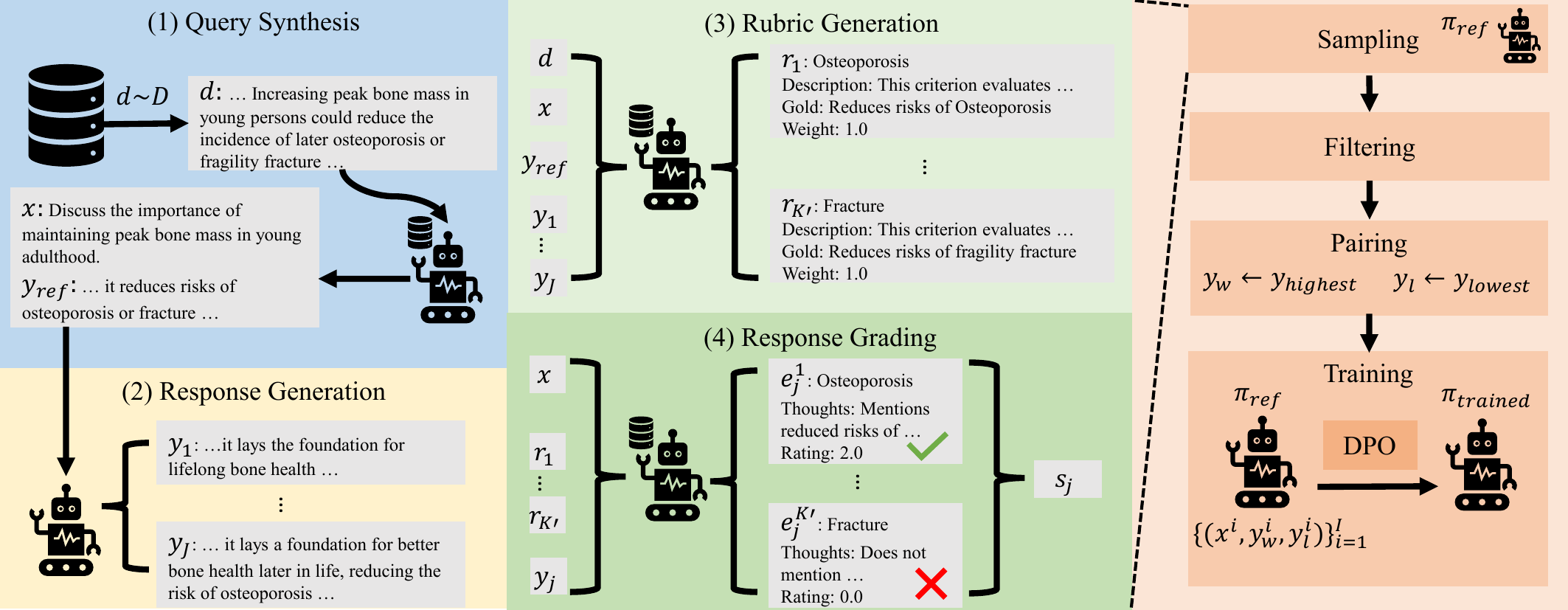}
    \caption{Training pipeline for \ourmethod. We sample a set of examples from the base model and create response pairs from them, which are then used to train the model via DPO. To create each example, (1) we sample a piece of text from a pre-training corpus and, grounded on that, the LLM synthesizes a query and a reference response; (2) the LLM produces a set of candidate responses to the query; (3) the LLM generates an evaluation rubric, conditioned on the pre-training text; (4) the LLM scores each response according to the rubric. After sampling, we filter out invalid examples and take the responses with the highest and lowest scores as the preferred and dispreferred responses, respectively. The resulting dataset is used for DPO.}
    \label{fig:pipeline}
    \vspace{-10pt}
\end{figure}

Our resulting approach \textbf{P}ost-trains LLMs for \textbf{O}pen domains with \textbf{P}retraining text (\textbf{\ourmethod}). \ourmethod~uses the same model in three roles: the Proposer, which synthesizes queries, grounded on pre-training text; the Solver, which answers the queries; and the Verifier, which generates the rubric and grades each Solver's response, again grounded on pre-training text. We evaluate \ourmethod~on three open-ended tasks: long-form healthcare QA, creative writing, and general instruction following. For each task, we change only the pre-training corpus and the prompts for the proposer and solver, while keeping other parts of the framework fixed.

Our main contributions are as follows. (i) We propose \ourmethod, the first self-play framework for post-training LLMs on open-ended tasks, using rubric-based evaluations; (ii) We propose the use of pre-training text to ground \ourmethod~to enable an exploitable generation-verification gap without introducing strong supervision; (iii) Experiment results show that \ourmethod~increases performance of pretrained base models and instruction-tuned models on three quite distinct tasks --- long-form Healthcare QA (up to 3\% on HealthBench500 \cite{arora2025}), creative writing (up to 3\% on Creative Writing V3 \cite{creative-writing-bench-v3}), and instruction following (up to 5\% on IFEval \cite{zhou2023instructionfollowing} and 4\% on ArenaHard \cite{zheng2023}); (iv) Further analysis suggests that while our rubric evaluations are still imperfect, pairing the highest and lowest scored responses produces mostly correct training signals.

%% file: related_work.tex
\vspace{-10pt}
\section{Related Work}
% \vspace{-10pt}
\paragraph{Self-Play.} Prior work applies self-play almost exclusively to verifiable reasoning tasks: \cite{absolutezero2025} and \cite{zhou2025} apply self-play to coding tasks; \cite{chen2025, huang2025rzero, liu2025, zhang2025} focus on math reasoning; \cite{cheng2024} works with verifiable commonsense reasoning. \cite{deepseek2025} and \cite{karl2026} synthesize question-answer pairs on various verifiable agentic tasks, with access to tools that interact with environments. However, these verifiable tasks are only a small part of what is needed to train highly capable LLMs \citep{chung2022}.  Real-world use cases often require long-form open-ended output, such as document-assisted writing (e.g., summarization), creative writing (e.g., storytelling), information extraction, open-ended QA, etc. POP focuses on self-play in the context of open-ended tasks.

\paragraph{Rubric as Rewards.} There is a growing body of literature on extending existing Reinforcement Learning (RL) methods to post-train LLMs in non-verifiable domains. One major branch uses rubrics to score open-ended text from the model, and uses the score as the reward for RL. Without exception, these methods require access to humans or strong LLMs to generate rubric or evaluate answers, both of which are expensive and, in some cases, unavailable.

% Specifically, \cite{dineen2025} employs strong LLMs to synthesize hierarchical rubrics, which are then used subsequently for SFT and RL. \cite{gunjal2025, viswanathan2025} generate question-specific rubrics using strong LLMs, adopting a binary rating scale. \cite{huang2025rlra} curate a set of rubrics in advance, from human experts or strong LLMs, and use these rubrics to synthesize questions, which are then fed into their rubric-based RL pipeline. \cite{rezaei2025} ask a strong LLM to compare responses from the current model against those from a reference model to curate new rubric criteria during training; they then use another strong LLM to grade the responses. \cite{zhou2025} use rubrics to guide the solver to generate high-quality responses, and gradually limit access to rubrics during training for curriculum learning. \cite{he2025} hire humans to generate the rubrics and score responses; these are used to train their LLM-based rubric generators and verifiers. \cite{bi2025} ask humans or strong LLMs to synthesize rubrics from reference answers and use them for RL training. Similar to \cite{zhou2025}, they selectively refine the sampled responses using rubrics to enhance RL training efficiency. 

Specifically, \cite{dineen2025} rely on strong LLMs to synthesize hierarchical rubrics according to given principles. \cite{huang2025rlra} synthesize the rubrics in advance, from human or strong LLMs, and use these rubrics to synthesize questions for training. \cite{gunjal2025, viswanathan2025} use strong LLMs to generate question-specific rubrics. \cite{rezaei2025} ask strong LLMs to compare responses from the current model against those from a reference model to curate new rubric criteria during training. \cite{he2025} use human-annotated rubrics and scores to train their LLM-based rubric generators and verifiers. \cite{zhou2025, bi2025} use human or strong-LLM-synthesized rubrics to guide the solver to generate high-quality responses, to either enable curriculum learning or boost training efficiency. In contrast to all of the above, POP does not rely on strong supervision for rubric evaluation.

%% file: method.tex
\section{Methodology}
\label{sec:method}

The full training pipeline of \ourmethod~is shown in Figure~\ref{fig:pipeline}; for pseudo code, see Algorithm~\ref{ago:pipeline}. \ourmethod~is a rubric-based self-play framework that uses the \textit{same} model to (1) synthesize queries, (2) generate responses, (3) generate rubrics, and (4) grade the responses. Steps (1), (3), (4) are grounded on the pre-training text. The rubric scores are used to construct response pairs for DPO training.

% \textcolor{red}{[You'll need to summarize the entire approach here. The caption has some of this (and you can borrow from it to put here) but doesn't provide any motivation for the approach.  The intro has the high-level idea, which you can restate here...starting with the goals of the approach (perhaps in contrast to the related work) and maybe using the Proposer/Solver/etc. terminology but linking it to specific parts of the figure and to the subsection titles. Just need to summarize the steps. Needs to be clear in this summary that you will be using DPO (and why) so maybe the "Preliminaries" and the "Training" paragraphs from 3.2 will migrate here.]}

\subsection{Sampling}

We presume access to a pre-training corpus $D$ with content relevant to the target task $t$ (e.g., a corpus of medical articles or clinical records for health QA; fantasy books for creative writing; general text for instruction following). Given $D$, we prompt a base LLM $\pi_{ref}$ to synthesize $I$ examples of $t$. Each example is synthesized according to the following four steps.

\paragraph{Query Synthesis (Proposer).} First,  a document $d$ is sampled from $D$. To ensure that $d$ is informative, we force it to be at least 50 words long. For efficiency reasons, we also truncate $d$ to the first 1024 words. Next, the LLM is prompted to synthesize a new query $x$ and a reference response $y_{ref}$ conditioned on $d$: $(x,y_{ref}) \sim \pi_{ref}(\cdot |P_{t}^{proposer}(d))$, where $P_{t}^{proposer}$ is a task-specific query synthesis prompt (see Appendix~\ref{app:prompts} for all prompts). $y_{ref}$ will be used later for rubric generation.

For tasks where ground truth answers exist (e.g., long-form health/medical QA), we ask in the prompt that the answer to $x$ be derivable from $d$. For tasks where ground truth does not exist (e.g., creative writing; instruction-following), we relax the condition and ask that the query be drawn from $d$. 

\paragraph{Response Generation (Solver).} A set of candidate responses to $x$ is sampled from $\pi_{ref}$ according to a task-specific response-generation prompt $P_{t}^{solver}$: $\forall j\in\{1,\cdots, J\}, y_j \sim \pi_{ref}(\cdot |P_{t}^{ans}(x))$. 

\paragraph{Rubric Generation.} A \textit{query}-specific rubric $r$ is generated by $\pi_{ref}$ according to a rubric generation prompt $P^{rubric}$ and conditioned on the pre-training document $d$, query $x$, reference response $y_{ref}$, and candidate responses $\{y_j\}_{j=1}^{J}$: $r\sim \pi_{ref}(\cdot | P^{rubric}(d, x, y_{ref}, \{y_j\}_{j=1}^{J}))$.

$r$ consists of $K'$ criteria $r_1, \cdots, r_{K'}$, where $K'$ is upper bounded by a constant $K$. Each criterion $r_k$ has a name, a description of the characteristics of good and bad responses, an optional gold label $g_k$ extracted from the document $d$, and a weight $w_k$ that indicates how important the criterion is.

Importantly, we enforce several principles via the rubric generation prompt. First, the rubric must be \textbf{grounded on the pre-training document $d$}. For criteria that have ground truth or gold standard answers (e.g., correctness criteria), the prompt requests the model to also extract the gold label $g_k$ from $d$ for that criterion. This provides \textit{explicit grounding} on the pre-training corpus. For criteria where there is no ground truth (e.g., creativity criteria), we prompt the model to utilize $d$ when generating the criterion description, which serves as \textit{implicit grounding} on the pre-training corpus.

In addition, the rubric must be \textbf{discriminative}: the model is prompted to generate only criteria that meaningfully distinguish high-quality responses from low-quality ones in the given response pool $\{y_{ref}\}\cup \{y_j\}_{j=1}^{J}$. In contrast to \cite{shao2025}, note that \ourmethod~includes a $d$-privileged reference response $y_{ref}$ in the rubric prompt. The rationale is that we want to discard examples where all candidate responses are highly similar and provide no meaningful contrastive signals. For these, $y_{ref}$ serves as an extra reference point to avoid generating meaningless criteria that focus on local differences between candidate responses but do not correlate with global quality. Consequently, these examples will receive the same score for all their candidate responses and be filtered out in the filtering stage.

\paragraph{Response Grading (Grader).} $\pi_{ref}$ is asked via prompt $P^{grader}$ to grade each candidate response $y_j$ and produce an evaluation report $e_j$ using the rubric $r$: $e_j\sim \pi_{ref}(\cdot | P^{grader}(x, r, y_j))$. $e_j$ consists of individual evaluations $e_{j}^{k}$ of $y_j$ for each criterion $r_k$. For each $e_{j}^{k}$, we ask the model to give its thoughts on how well the model does according to $r_k$'s description and the criterion-specific gold label $g_k$. Then, we ask it to produce a rating $s_{j}^{k}$ of either 0 (Bad/Does not match gold), 1 (Medium/Partially matches gold), or 2 (Good/Fully matches gold), depending on the extent to which $y_j$ satisfies $r_k$. If no valid score can be extracted for a criterion, we set its score to 0. Finally, we aggregate the scores into a final score $s_j\in[0.0,2.0]$ as follows.
\begin{align}
    s_j = \frac{\sum_{k=1}^{K} w_k \cdot s_{j}^{k}}{\sum_{k=1}^{K} w_k}
\end{align}
\subsection{Filtering, Pairing, and Training}

\paragraph{Preliminaries.} Direct Preference Optimization \cite{dpo2023} is an offline version of traditional policy gradient RL methods such as PPO \cite{ppo2017}. It avoids the need for an extra reward model, and trains the main policy to learn the reward landscape directly. Given a dataset $\{(x^{i}, y_w^{i}, y_l^{i})\}_{i=1}^{N}$, where each example has a prompt $x$, a preferred response $y_w$, and a dispreferred response $y_l$, it initializes the main policy $\pi_\theta$ from the reference policy $\pi_{ref}$ and trains it using the following objective.
\begin{align}
    L_{DPO}(\pi_\theta,\pi_{ref}) = -E_{(x, y_w, y_l)\sim S} \left[log \sigma\left(\beta log\frac{\pi_\theta(y_w|x)}{\pi_{ref}(y_w|x)} - \beta log\frac{\pi_\theta(y_l|x)}{\pi_{ref}(y_l|x)}\right) \right]
\end{align}
\paragraph{Filtering.} After synthesizing the dataset, for each example, we filter out candidate responses that are malformed or have a malformed evaluation report. In addition, we filter out examples for which the query or rubric cannot be extracted, have an incorrect format, or have zero valid candidate responses.

\paragraph{Pairing.} For each example that passes filtering, we pick the candidate response with the highest score as the preferred response $y_w \leftarrow argmax_{y_j}(s_j)$ and the one with the lowest score as the dispreferred response $y_l \leftarrow argmin_{y_j}(s_j)$. Following \cite{huang2025}, we want $y_w$ and $y_l$ to be maximally different in terms of quality but minimally different in terms of other irrelevant features such as length. Therefore, we filter out pairs where $y_w$ and $y_l$ differ by more than 100 words in length. We also filter out pairs where $y_w$ and $y_l$ have the same score. This gives the DPO training set $\{(x^{i}, y_w^{i}, y_l^{i})\}_{i=1}^{N}$.

\paragraph{Training.} We initialize the main policy from $\pi_{ref}$ and then train it using the DPO objective in equation (2) on the synthesized dataset. We denote the trained model as $\pi_{trained}$.

%% file: setup.tex
\section{Experiment Setup}
\label{sec:setup}

\textbf{Tasks.} We experiment with three tasks: Long-form Healthcare QA, Creative Writing, and Instruction Following. For \textbf{Long-form Healthcare QA}, we use HC4 \cite{maslenkova2025} as our pre-training corpus, which consists of more than 9.7 million medical articles. We evaluate on HealthBench \cite{arora2025}, which assesses LLMs' ability to address queries from either patients or health professionals. Each example is accompanied by rubrics created by human experts. To reduce evaluation cost, we randomly sample 500 examples from it and denote it as HealthBench500. For \textbf{Creative Writing}, we use 2 million book abstracts scraped from the Internet \cite{bookabstracts} as the pre-training corpus. We evaluate on Creative Writing V3 \cite{creative-writing-bench-v3}, which assesses LLMs' creative writing skills via pre-defined rubrics. For \textbf{Instruction Following}, we use OpenWebText \cite{Gokaslan2019OpenWeb} as the pre-training corpus, which consists of 8 million documents in the general domains. For evaluation, we use IFEval \cite{zhou2023instructionfollowing}, a verifiable benchmark that measures strict adherence to writing constraints (e.g., respond within 100 words), and ArenaHard \cite{zheng2023}, an LLM-as-a-judge benchmark measuring general instruction following abilities for diverse user queries (write emails, suggest weekend fun ideas, etc.). We use GPT-4.1-mini \cite{gpt42023} as the judge model for LLM-as-a-judge benchmarks.

\textbf{OOD Evaluations.} To ensure that our trained model's performance does not degrade on other tasks, especially on verifiable tasks, we evaluate \ourmethod~on ten out-of-distribution (OOD) benchmarks: on scientific reasoning (GPQA-Diamond \cite{rein2024gpqa}); math reasoning (GSM8K \cite{cobbe2021gsm8k}; Math500 \cite{hendrycksmath2021}; AIME2024; AIME2025); and factoid QA (NaturalQueries \cite{kwiatkowski2019}; TriviaQA \cite{2017arXivtriviaqa}; TruthfulQA \cite{stephanie2022}; MMLU-Pro \cite{mmlupro2024}); MedMCQA \cite{pmlr-v174-pal22a}. We use 0-shot prompting for the benchmarks.

\begin{table}[]
    \centering
    \small
    \begin{tabular}{llccccccc}
        \toprule
         Task & Model & \#Qry (I) & \#Rsp (J) & \#Criteria (K) & $s$ mean & $s$ std & $s_{y_w}$ & $s_{y_l}$ \\
         \midrule
         Healthcare QA & Base & 1381 & 10.69 & 4.48 & 1.19 & 0.47 & 1.78 & 0.45 \\
          & Inst & 2667 & 24.45 & 2.71 & 1.01 & 0.29 & 1.51 & 0.54 \\
        \midrule
        Creative Writing & Base & 1822 & 8.91 & 4.84 & 1.41 & 0.58 & 1.96 & 0.47 \\
          & Inst & 2521 & 28.95 & 2.96 & 1.66 & 0.21 & 1.95 & 1.21 \\
        \midrule
        Instruction Following & Base & 1665 & 10.53 & 4.47 & 1.20 & 0.50 & 1.80 & 0.43 \\
          & Inst & 2502 & 24.68 & 3.04 & 1.35 & 0.30 & 1.79 & 0.80 \\ 
        \bottomrule
    \end{tabular}
    \caption{Statistics of the synthesized datasets, after filtering. Base: Qwen-2.5-7B; Inst: Qwen-2.5-7B-Instruct; \#Qry: Number of queries; \#Rsp: Number of valid candidate responses per query; \#Crtieria: Number of rubric criteria per query; $s$ mean: Average response score per query; $s$ std: Answer score standard deviation per query; $s_{y_w}$: Average score of preferred response; $s_{y_l}$: Average score of dispreferred response. Score range is from 0.0 to 2.0.}
    \label{tab:stat}
\vspace{-10pt}
\end{table}

\textbf{Models.} We experiment with both a pretrained base model (Qwen-2.5-7B \cite{qwen252024}) and an instruction-finetuned model (Qwen-2.5-7B-Instruct \cite{qwen252024}) as our reference model $\pi_{ref}$.

\textbf{Sampling.} We sample $I=4096$ queries (except for Creative Writing, where we set $I=8192$), each with $J=32$ initial candidate responses, and ask the model to create at most $5$ rubrics per query (i.e., $K=5$). The statistics of the resulting synthesized datasets are shown in Table~\ref{tab:stat}.

\textbf{Training.} We train the model for 1 epoch, with learning rate $1e^{-6}$, $\beta=0.01$, and batch size $16$. We use AdamW optimizer \cite{adamw2019} with a cosine learning scheduler. See details in Appendix~\ref{app:hyperparameters}.

\textbf{Baselines.} We compare $\pi_{trained}$ against (1) the reference model $\pi_{ref}$; (2) a model that is continuously pretrained on the same pre-training text used to ground our pipeline (\textbf{CPT on $D$}); (3) a model that is supervise-finetuned on the same queries, but with reference responses (\textbf{SFT on $y_{ref}$}). Prior rubric-based RL methods \cite{gunjal2025, viswanathan2025, bi2025, zhou2025, he2025, rezaei2025, huang2025rlra} use strong supervision from large frontier models or humans to obtain the input queries, rubrics, or scores, and are not comparable to \ourmethod.

%% file: results.tex
\vspace{-5pt}
\section{Results and Analysis}
\vspace{-5pt}
We first discuss the main results in \S~\ref{sec:main_results}, then evaluate the trained models on OOD benchmarks in \S~\ref{sec:ood_results}. We also analyze the synthesized training datasets in \S~\ref{sec:stat} and conduct various ablation studies in \S~\ref{sec:ablation_study}. Finally, we analyze the correctness of response ranking by our rubric evaluation in \S~\ref{sec:ranking_analysis}.

\vspace{-2pt}
\subsection{Main Results}
\label{sec:main_results}
\vspace{-2pt}

We show the main results in Table~\ref{tab:main_results}. Across different domains, we observe a consistent performance increase for \ourmethod~over the reference Qwen models. In addition, both continuous pre-training on the grounding documents $D$ (CPT on $D$) and supervised fine-tuning on the reference response $y_{ref}$ (SFT on $y_{ref}$) underperform our methods across all models and tasks, except for Qwen-2.5-7B on IFEval.

% \begin{table}[htp!]
%     \centering
%     \begin{tabular}{lccccc}
%         \toprule
%          & Healthcare QA & \multicolumn{2}{c}{Creative Writing} & \multicolumn{2}{c}{Instruction Following} \\
%          \cmidrule(lr){2-2}
%         \cmidrule(lr){3-4}
%         \cmidrule(lr){5-6}
%          & HealthBench500 & \multicolumn{2}{c}{Creative Writing V3} & IFEval & ArenaHard \\
%          & Rubric & Rubric & Elo &  & Win Rate \\
%          \midrule
%          Qwen-2.5-7B & 26.36 & 41.18 & 1200.00 & 32.90 & 12.53 \\
%          % ~CPT on $D$ & 23.89 & 30.10 & 766.79 & 23.48 & 35.67 & 10.63 \\
%          ~CPT on $D$ & 27.26 & 42.21 & 1100.32  & 36.04 & 13.66 \\
%          ~SFT on $y_{ref}$ & 25.70 & 42.11 & 1079.37  & 35.49 & 12.63 \\
%          ~\ourmethod & \textbf{29.97} & \textbf{42.77} & \textbf{1284.66} & \textbf{38.26} & \textbf{15.98} \\
%          \midrule
         
%          Qwen-2.5-7B-Inst & 40.83 & 59.96 & 1524.83 & 71.71 & 50.17 \\
%          % ~CPT on $D$ & 26.14 & 40.38 & 1026.52 & 52.87 & 63.43 & 41.30 \\
%          ~CPT on $D$ & 39.74 & 54.33 & 1357.76 & 70.06 & 48.20 \\
%          ~SFT on $y_{ref}$ & 39.71 & 58.67 & \textbf{1638.03} & 70.24 & 45.30 \\
%          ~\ourmethod & \textbf{41.16} & \textbf{64.71} & 1577.32 &  \textbf{73.57} & \textbf{54.09} \\
%          \bottomrule
%     \end{tabular}
%     \caption{Main results. Rubric: Standard rubric score from original benchmarks; Elo: Elo score . IFEval: We report the strict-prompt subset results, according to common practice.}
%     \label{tab:main_results}
% \vspace{-10pt}
% \end{table}

\begin{table}[htp!]
    \small
    \centering
    \begin{tabular}{lcccc}
        \toprule
         & Healthcare QA & Creative Writing & \multicolumn{2}{c}{Instruction Following} \\
         \cmidrule(lr){2-2}
        \cmidrule(lr){3-3}
        \cmidrule(lr){4-5}
         & HealthBench500 & Creative Writing V3 & IFEval & ArenaHard \\
         % & Rubric & Rubric &  & Win Rate \\
         \midrule
         Qwen-2.5-7B & 26.71 ($\pm 0.57$) & 42.31 ($\pm 0.81$)  & 31.79 ($\pm 0.91$) & 12.26 ($\pm 0.33$) \\
         % ~CPT on $D$ & 23.89 & 30.10 & 23.48 & 35.67 & 10.63 \\
         ~CPT on $D$ & 27.49 ($\pm 0.64$) & 41.42 ($\pm 0.89$) & 33.64 ($\pm 1.71$) & 13.99  ($\pm 0.28$) \\
         ~SFT on $y_{ref}$ & 25.85 ($\pm 0.42$) & 41.93 ($\pm 0.83$) & \textbf{37.46} ($\pm 1.52$) & 13.73 ($\pm 0.78$)\\
         ~\ourmethod & \textbf{29.64} ($\pm 0.75$) & \textbf{44.31} ($\pm 1.23$) & 36.72 ($\pm 1.21$) & \textbf{16.20} ($\pm 0.29$)\\
         \midrule
         
         Qwen-2.5-7B-Inst & 40.35 ($\pm 0.43$) & 59.82 ($\pm 0.65$) & 70.30 ($\pm 1.02$) & 50.08 ($\pm 1.43$)\\
         % ~CPT on $D$ & 26.14 & 40.38 & 52.87 & 63.43 & 41.30 \\
         ~CPT on $D$ & 39.31 ($\pm 0.37$) & 53.91 ($\pm 0.45$) & 69.87 ($\pm 0.99$) & 47.34 ($\pm 0.92$)\\
         ~SFT on $y_{ref}$ & 39.38 ($\pm 0.66$)& 58.25 ($\pm 0.42$) & 70.42 ($\pm 0.42$) & 45.87 ($\pm 1.02$)\\
         ~\ourmethod & \textbf{41.12} ($\pm 0.14$) & \textbf{62.64} ($\pm 1.57$) &  \textbf{72.34} ($\pm 0.87$) & \textbf{53.78} ($\pm 0.39$)\\
         \bottomrule
    \end{tabular}
    \caption{Main results. HealthBench and Creative Writing V3: we report standard rubric scores from original benchmarks; IFEval: We report the strict-prompt results, according to common practice \cite{qwen252024, zhou2025}. ArenaHard: We report the win rate against GPT-4. We repeat all evaluations three times and report the standard errors.}
    \label{tab:main_results}
%\vspace{-8pt}
\end{table}

\begin{figure}
\begin{subfigure}[h]{0.5\linewidth}
\includegraphics[width=\linewidth]{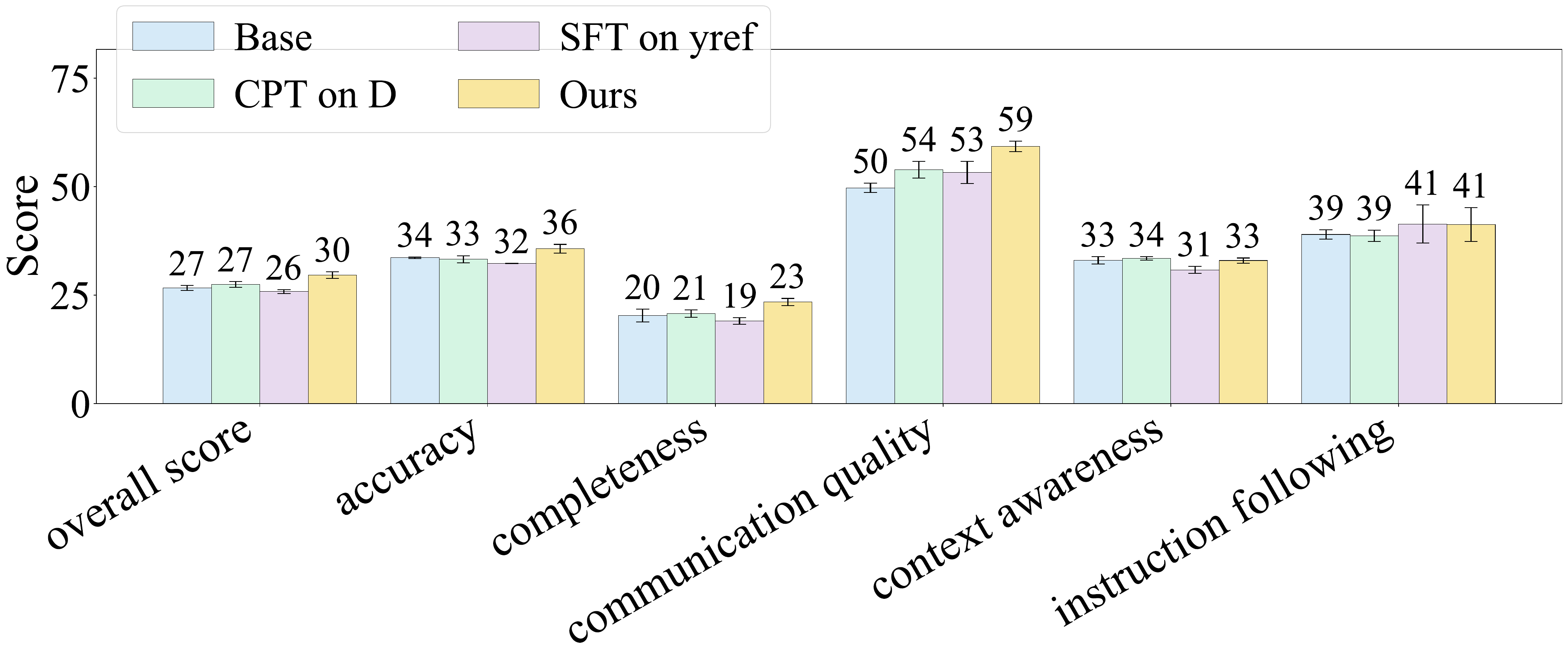}
\end{subfigure}
\hfill
\begin{subfigure}[h]{0.5\linewidth}
\includegraphics[width=\linewidth]{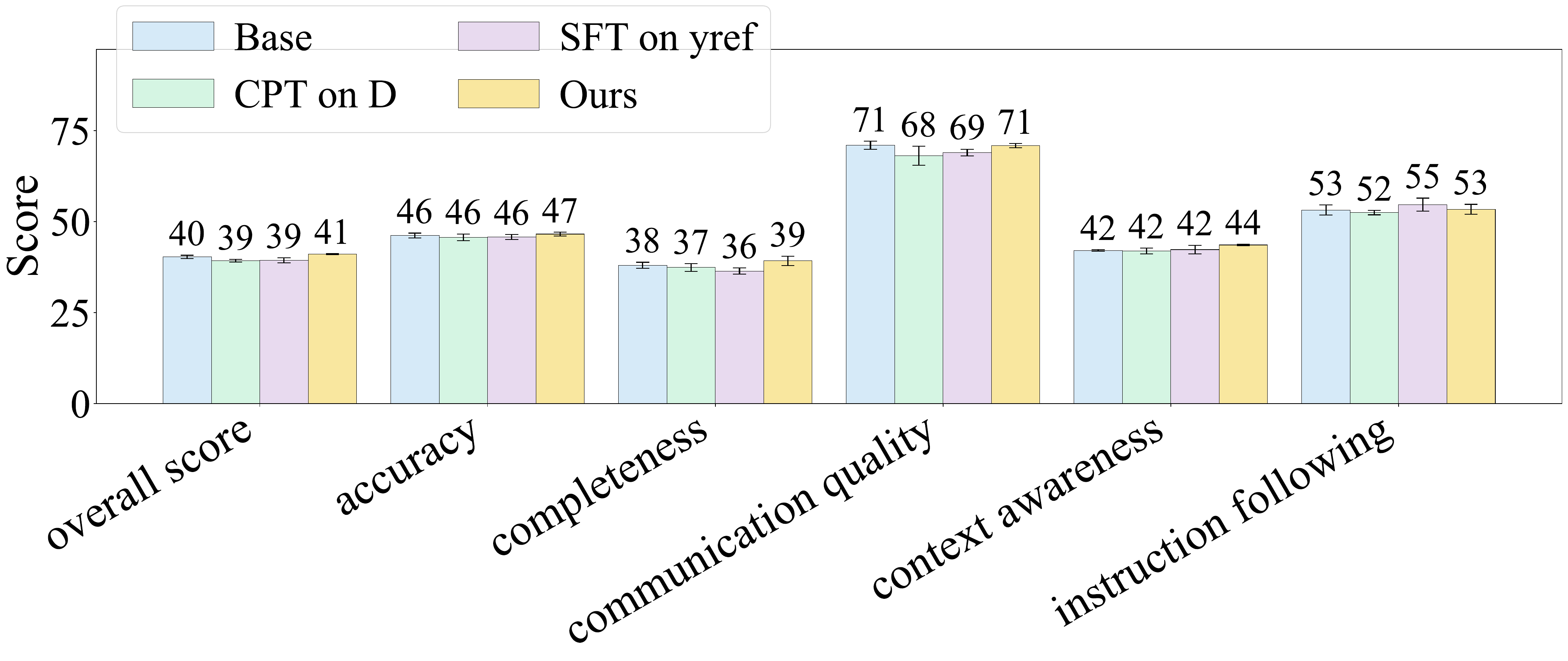}
\end{subfigure}%
\caption{Per-aspect results on HealthBench500. Left: Qwen-2.5-7B; Right: Qwen-2.5-7B-Inst.}
\label{fig:hc_per_axis}
\vspace{-10pt}
\end{figure}

\paragraph{Healthcare QA.} On the pretrained model (Qwen-2.5-7B), we observe a 3\% increase on HealthBench500. The performance gains on the instruction-tuned model (Qwen-2.5-7B-Inst) are only 1\%. We suspect this is due to misalignments between our synthesized training data and the evaluation data from HealthBench. In Figure~\ref{fig:hc_per_axis}, we show finer-grained results on benchmark-defined answer aspects. For Qwen-2.5-7B, performance increases notably on all axes except context awareness. For Qwen-2.5-7B-Inst, performance slightly increases in accuracy, completeness, and context awareness, but remains the same in communication quality and instruction following.

\paragraph{Creative Writing.} On Creative Writing V3, \ourmethod~gets a 2\% increase for Qwen-2.5-7B and a 3\% increase for Qwen-2.5-7B-Inst. Per-criteria results can be found in Appendix~\ref{app:res_cw}.

\begin{figure}[htp!]
\begin{subfigure}[h]{0.5\linewidth}
\includegraphics[width=\linewidth]{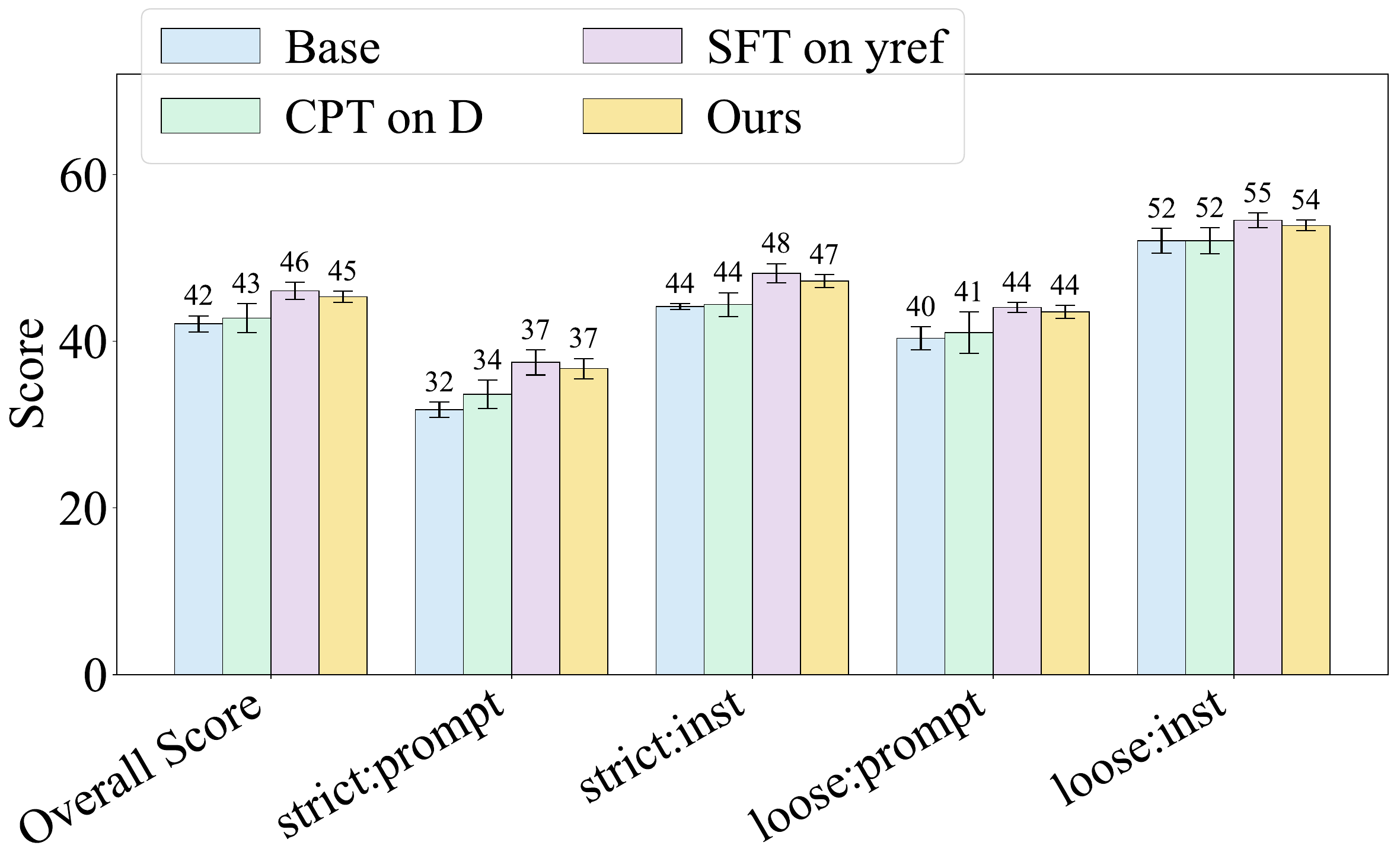}
\end{subfigure}
\hfill
\begin{subfigure}[h]{0.5\linewidth}
\includegraphics[width=\linewidth]{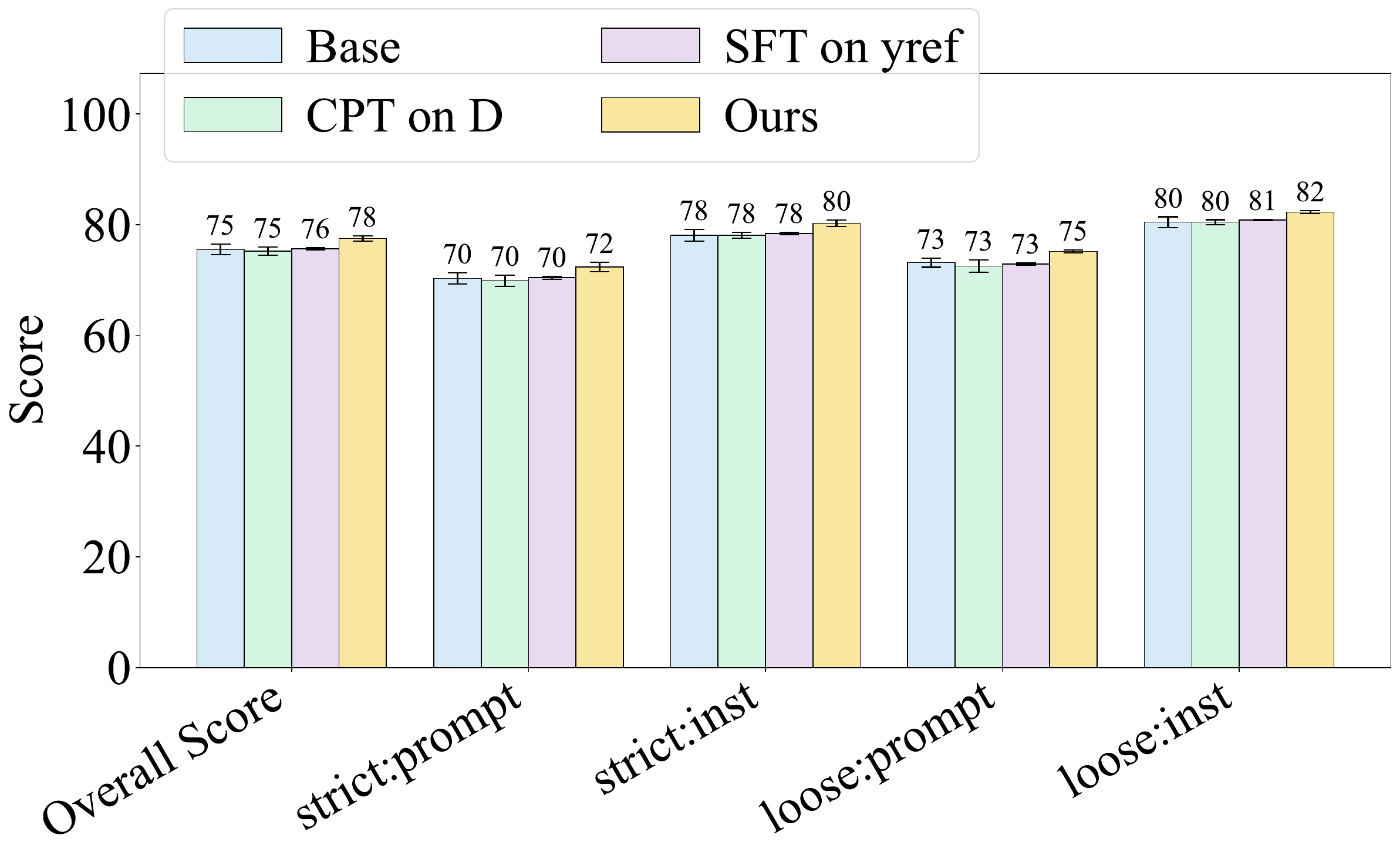}
\end{subfigure}%
\caption{Per-axis results on IFEval. Left: Qwen-2.5-7B; Right: Qwen-2.5-7B-Inst.}
\label{fig:if_per_axis}
\vspace{-10pt}
\end{figure}

\paragraph{Instruction Following.} We first discuss results on IFEval. We observe a 5\% increase for Qwen-2.5-7B and 2\% increase for Qwen-2.5-7B-Inst. Detailed breakdown in Figure~\ref{fig:if_per_axis} indicates a consistent performance increase under both strict and loose evaluations.
% For Qwen-2.5-7B, \ourmethod~gives a 9\% increase on prompt-level metrics and a 7.5\% increase on instruction-level metrics. For Qwen-2.5-7B-Inst, increases are milder. We observe a 3.5\% increase on the prompt-level and a 3\% increase on the instruction-level. 
On ArenaHard, \ourmethod~improves the win rate against GPT-4 by 4\% for Qwen-2.5-7B and 4\% for Qwen-2.5-7B-Inst.

\vspace{-2pt}
\subsection{Results on OOD Benchmarks}
\label{sec:ood_results}
\vspace{-2pt}

% \begin{wrapfigure}{r}{0.3\textwidth}
%     \vspace{-1cm}
%     \centering
%     \includegraphics[width=0.3\textwidth]{assets/medqa.pdf}
%     \caption{MedQA results.}
%     \label{fig:medqa}
% \vspace{-0.8cm}
% \end{wrapfigure}
We present the OOD results for models trained on Healthcare QA datasets in Table~\ref{tab:ood_hc_main}. OOD results for models trained on the Creative Writing and Instruction Following datasets can be found in Appendix~\ref{app:ood_results}. Across models and tasks, we see no performance drop on OOD benchmarks.

\begin{table}[]
    \small
    \centering
    \resizebox{\linewidth}{!}{%
    \begin{tabular}{lcccccccccc}
    \toprule
         & MedMCQA & NQ & TvQA & TfQA & MMLU-P & GPQA-D & GSM & Math & Aime24 & Aime25 \\
    \midrule
    \multicolumn{11}{c}{Qwen-2.5-7B} \\
    \midrule
         Base & 52.95 & 14.36 & 17.93 & 48.07 & 30.10 & 33.84 & 78.77 & 61.15 & 6.82 & 3.07 \\
         % ~CPT on $D$ & 53.12 & 12.54 & 14.76 & 48.06 & 39.32 & 28.28 & \textbf{80.31} & 61.25 & 5.89 & 3.23 \\
         ~CPT on $D$ & \textbf{54.00} & 14.01 & 16.96 & 48.05 & 34.71 & 33.84 & 79.44 & 61.00 & \textbf{7.76} & 3.28 \\
         ~SFT on $y_{ref}$ & 53.81 & \textbf{15.30} & 16.55 & 48.62 & 43.78 & 31.82 & 78.05 & 59.60 & 6.09 & 3.33 \\
         ~\ourmethod & 53.86 & 14.65 & \textbf{18.96} & \textbf{49.81} & \textbf{44.75} & \textbf{35.35} & \textbf{79.98} & \textbf{64.00} & \textbf{7.76} & \textbf{4.32} \\
    \midrule
    \multicolumn{11}{c}{Qwen-2.5-7B-Inst} \\
    \midrule
         Base & 55.94 & \textbf{15.12} & 18.49 & 62.51 & \textbf{56.57} & \textbf{36.36} & \textbf{90.64} & 75.55 & \textbf{12.40} & 8.07 \\
         % ~CPT on $D$ & 40.74 & 8.50 & 9.22 & 56.40 & 48.92 & 26.26 & 87.02 & 69.80 & 7.76 & 5.47 \\
         ~CPT on $D$ & \textbf{56.99} & 14.60 & \textbf{22.80} & 62.34 & 56.52 & 33.33 & 90.24 & 74.85 & 11.56 & 7.40 \\
         ~SFT on $y_{ref}$ & 56.78 & 14.83 & 17.54 & \textbf{62.71} & 56.20 & 33.84 & 90.16 & \textbf{75.75} & 11.25 & 7.55 \\
         ~\ourmethod & 56.54 & 14.01 & 18.99 & 62.35 & 56.49 & 33.84 & 90.52 & \textbf{75.75} & 12.03 & \textbf{8.23} \\
    \bottomrule
    \end{tabular}
    }
    \caption{OOD Results for models trained on the healthcare datasets. NQ: NaturalQueries; TvQA: TriviaQA; TfQA: TruthfulQA; MMLU-P: MMLU-Pro; GPQA-D: GPQA-Diamond; GSM: GSM8K; Math: MATH500. We use 0-shot evaluation.}
    \label{tab:ood_hc_main}
    \vspace{-10pt}
\end{table}

We observe a slight increase across knowledge and reasoning benchmarks for Qwen-2.5-7B. Notably, MMLU-Pro results increase from 30.10\% to 44.75\%. Average results change from 34.71\% to 37.34\%. On Qwen-2.5-7B-Inst, results slightly drop on NaturalQueries and GPQA-Diamond, but remains largely the same on other benchmarks. Average results change from 43.17\% to 42.88\%. We also highlight that results on MedMCQA increase by around 1\% for both models, suggesting that training on our long-form datasets slightly helps performance on short-form tasks in the same domain.

On Qwen-2.5-7B, in terms of average OOD scores, we (37.34\%) also slightly outperform CPT on $D$ (35.30\%) and SFT on $y_{ref}$ (35.70\%). On Qwen-2.5-7B-Inst, our results (42.88\%) are similar to the baselines (43.06\% for CPT on $D$ and 42.66\% for SFT on $y_{ref}$).

\vspace{-2pt}
\subsection{Statistics of the Synthesized Dataset}
\label{sec:stat}
\vspace{-2pt}

% \begin{wrapfigure}{r}{0.4\textwidth}
% \vspace{-1cm}
%     \centering
%     \includegraphics[width=0.4\textwidth]{assets/categorization/qwen25_7b_base_hc_topics_pie_chart.pdf}
%     \caption{Query topics.}
%     \label{fig:hc_topics_pie_chart}
%     \vspace{-1cm}
% \end{wrapfigure}
Below we analyze the types of queries, rubrics, and responses synthesized by \ourmethod. We use the $\text{Healthcare QA}_{\text{Qwen-2.5-7B}}$ dataset as an example. See details in Appendix~\ref{app:stat_dataset}.

% \subsubsection{Queries and Rubrics}

\textbf{Queries.} In Figure~\ref{fig:base_hc_stat} (a), we show the common topics for the synthesized queries.
Healthcare Management (21\%) and Public Health (19\%) are the two most common topics, which also have the strongest overlap with queries from HealthBench. Since a considerable portion of our pre-training corpus \cite{maslenkova2025} is medical articles from PubMed, the rest of the queries tend to focus on professional medical knowledge and research.

% \subsubsection{Rubrics}

% \begin{wrapfigure}{r}{0.4\textwidth}
% \vspace{-1cm}
%     \centering
%     \includegraphics[width=0.4\textwidth]{assets/categorization/qwen25_7b_base_hc_criteria_pie_chart.pdf}
%     \caption{Meta rubric criteria}
%     \label{fig:hc_criteria_pie_chart}
%     \vspace{-1cm}
% \end{wrapfigure}
\textbf{Rubrics.} We group the rubric criteria into meta-criteria and show the composition in Figure~\ref{fig:base_hc_stat} (b). Relevance (25\%), Accuracy (22\%), and Clarity (18\%) make up two-thirds of the rubric criteria. Other major categories include Depth (11\%), Biological Understanding (8\%), and Originality (6\%). This suggests that our rubric values both objective and subjective criteria.

% \subsubsection{Responses and Scores}

\begin{figure}
\begin{subfigure}[h]{0.5\linewidth}
\includegraphics[width=\linewidth]{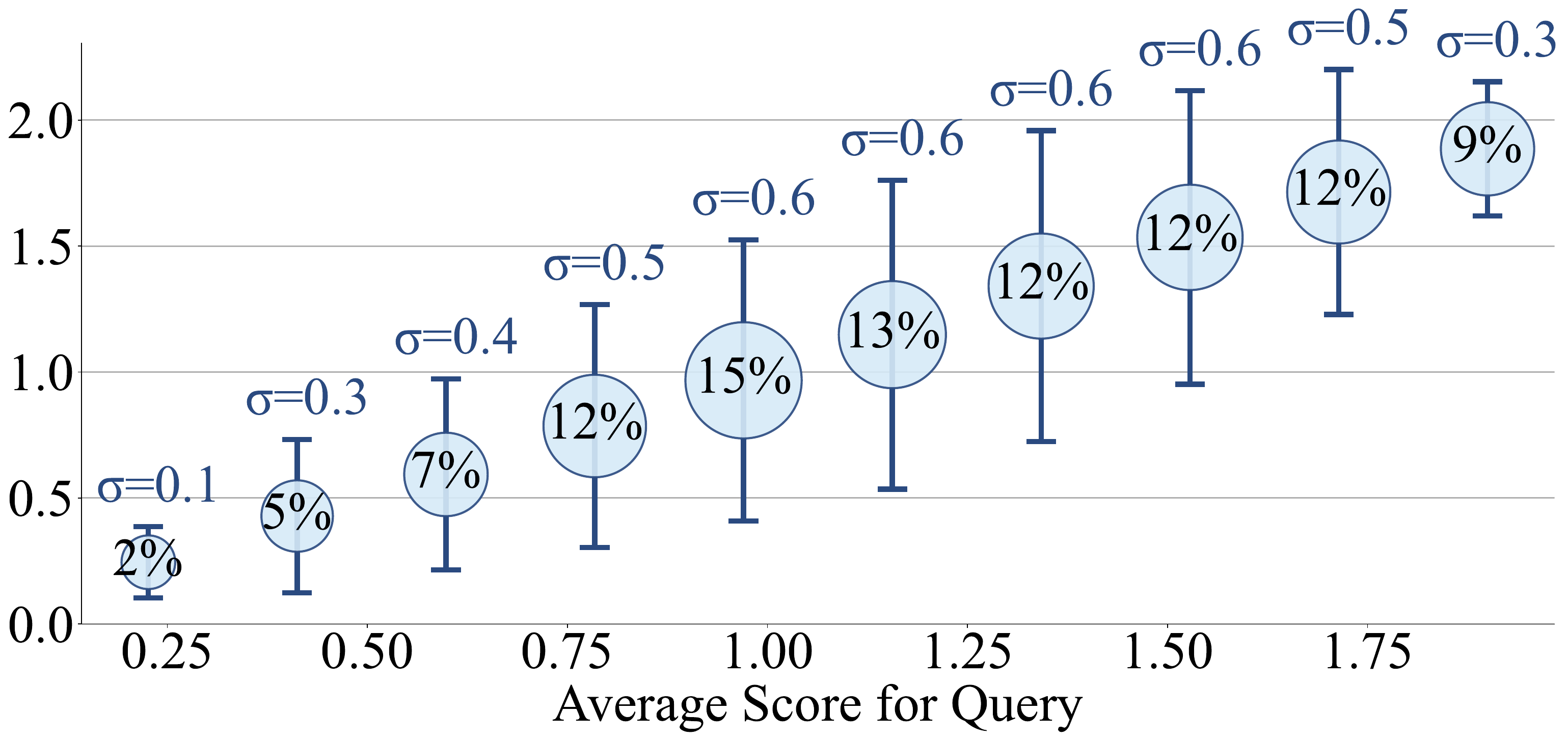}
\end{subfigure}
\hfill
\begin{subfigure}[h]{0.5\linewidth}
\includegraphics[width=\linewidth]{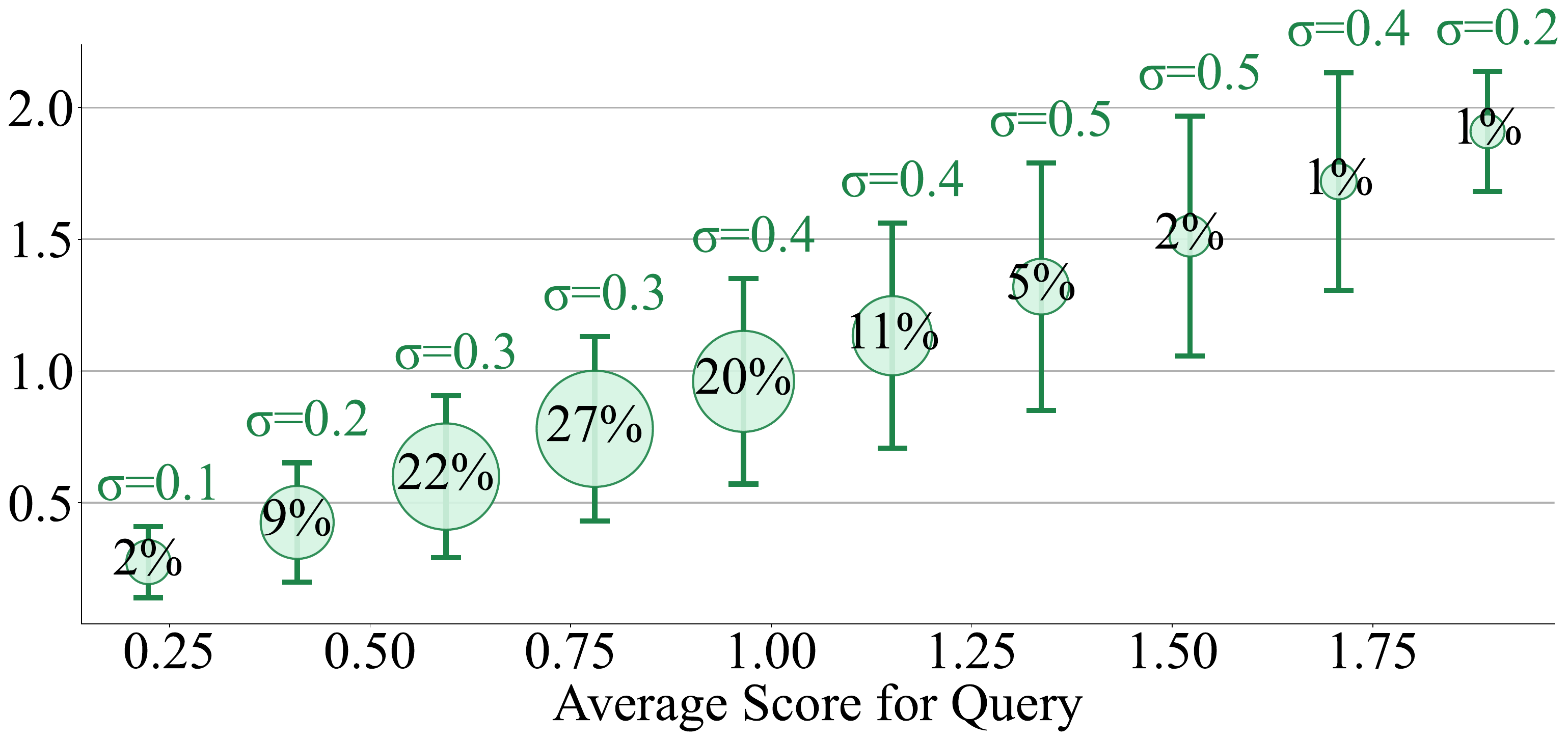}
\end{subfigure}%
\caption{Response score distribution for $\text{healthcare}_{\text{Qwen-2.5-7B}}$ dataset. Left: scores given by the model itself; Right: scores given by a stronger teacher model; We group the candidate responses by queries and compute the average and standard deviation of response scores for each query. We then partition the queries into 10 bins according to average response scores. x-axis is the average score for the queries in that bin; error bar indicates standard deviations of scores for the queries in that bin, averaged by number of queries; bubble size indicates the percentage of queries that fall into that bin.}
\label{fig:score_dist}
\vspace{-10pt}
\end{figure}

\textbf{Responses and Scores.} In Figure~\ref{fig:score_dist}, we show the distribution of response scores, grouped by queries. According to the rubric and scores generated by the model itself, more than 50\% of queries have their average scores above 1.0. However, as we will show in \S~\ref{sec:ranking_analysis}, these scores are not perfect without the pairing step. Therefore, we additionally ask a stronger teacher model $\pi_{teacher}$ (GPT-4o-mini) to generate its own rubrics and scores on the same queries and responses (right side). The teacher distribution moves toward the lower end. Over 50\% of queries have an average score below 1.0.

Nonetheless, we observe a large standard deviation of over 0.3 across query groups under both models. This suggests that \ourmethod~is able to synthesize queries of appropriate difficulties, so that the model generates responses of varying quality, and can give meaningful contrastive signals.
\vspace{-2pt}
\subsection{Ablation Study}
\label{sec:ablation_study}
\vspace{-2pt}
\begin{wrapfigure}{r}{0.65\textwidth}
  \vspace{-10pt}
  \begin{center}
    \includegraphics[width=0.65\textwidth]{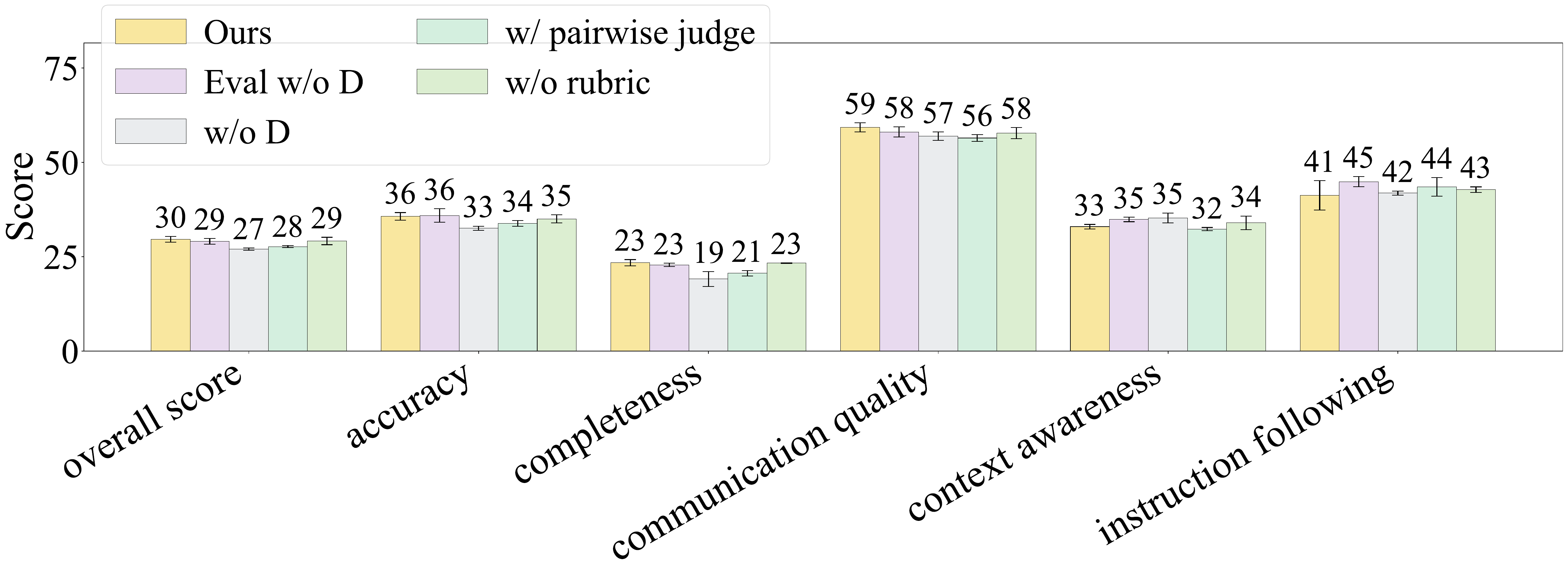}
  \end{center}
  \caption{Ablation Results on HealthBench500.}
  \label{fig:ablation_res}
  \vspace{-10pt}
\end{wrapfigure}
We remove or replace various components of \ourmethod. This includes (1) revoking access to pre-training text $d$ when generating the rubric (\texttt{Eval w/o $D$}); (2) revoking our entire pipeline's access to $d$ altogether (\texttt{w/o $D$}); (3) replacing our pointwise rubric grader with a pairwise judge that compares two responses according to our rubric (\texttt{w/ pairwise judge}); (4) removing the access to rubrics for our judge, but still grounding it on $d$ (\texttt{w/o rubric}). See Appendix~\ref{app:ablations_method} for details.

Figure~\ref{fig:ablation_res} shows the results for the ablation study on the $\text{Healthcare}_{\text{Qwen-2.5-7B}}$ dataset. Removing grounding of our rubric evaluation on the pre-training document reduces the overall performance to 29\%. We suspect this to be a result of an insufficient generation-verification gap and an increase in reward hacking, which we will analyze in \S~\ref{sec:ranking_analysis}. Removing our pipeline's access to $D$ entirely (w/o $D$)  further hurts performance, since in this case, queries are synthesized without grounding, and likely degrade in terms of quality and diversity. Grading responses without rubrics slightly hurts the performance, and we argue that rubrics make evaluation more interpretable. Surprisingly, using a pairwise judge is not better than ours, which we will analyze in Appendix~\ref{app:ablation_pairwise_analysis}.

% \begin{table}[]
%     \centering
%     \begin{tabular}{cc}
%          \toprule
%          & HealthBench500 \\
%          \ourmethod &  \\
%          w/o $D$ &  \\
%          w/ teacher query &  \\
%          w/ teacher response & \\
%     \end{tabular}
%     \caption{Caption}
%     \label{tab:placeholder}
% \end{table}

\vspace{-2pt}
\subsection{Response Ranking Analysis}
\label{sec:ranking_analysis}
\vspace{-2pt}
Without external input, using the same model to verify its own responses risks an insufficient generation-verification gap \cite{gap2025}. In our case, the rubric generated by our model may not distinguish high-quality responses from low-quality ones. As a result, the ranking of responses according to the rubric will be inconsistent with the "true quality" ranking. Choosing responses based on this ranking for training encourages undesired behavior, a phenomenon known as \textbf{reward hacking}. In our case, since we only choose the highest-ranked $y_w$ and lowest-ranked $y_l$ responses for DPO training, reward hacking happens when $y_w$ is worse than $y_l$ in terms of "true qualities".

To measure "true qualities", we use a stronger teacher model $\pi_{teacher}$ as a proxy. In particular, we use our prompts to ask $\pi_{teacher}$ to generate a new set of rubrics and scores for the same queries and responses from $\pi_{ref}$. We define the rankings using this stronger model as the \textit{gold rankings}. We want to know
\begin{figure}[h]
\begin{subfigure}[h]{0.5\linewidth}
\includegraphics[width=\linewidth]{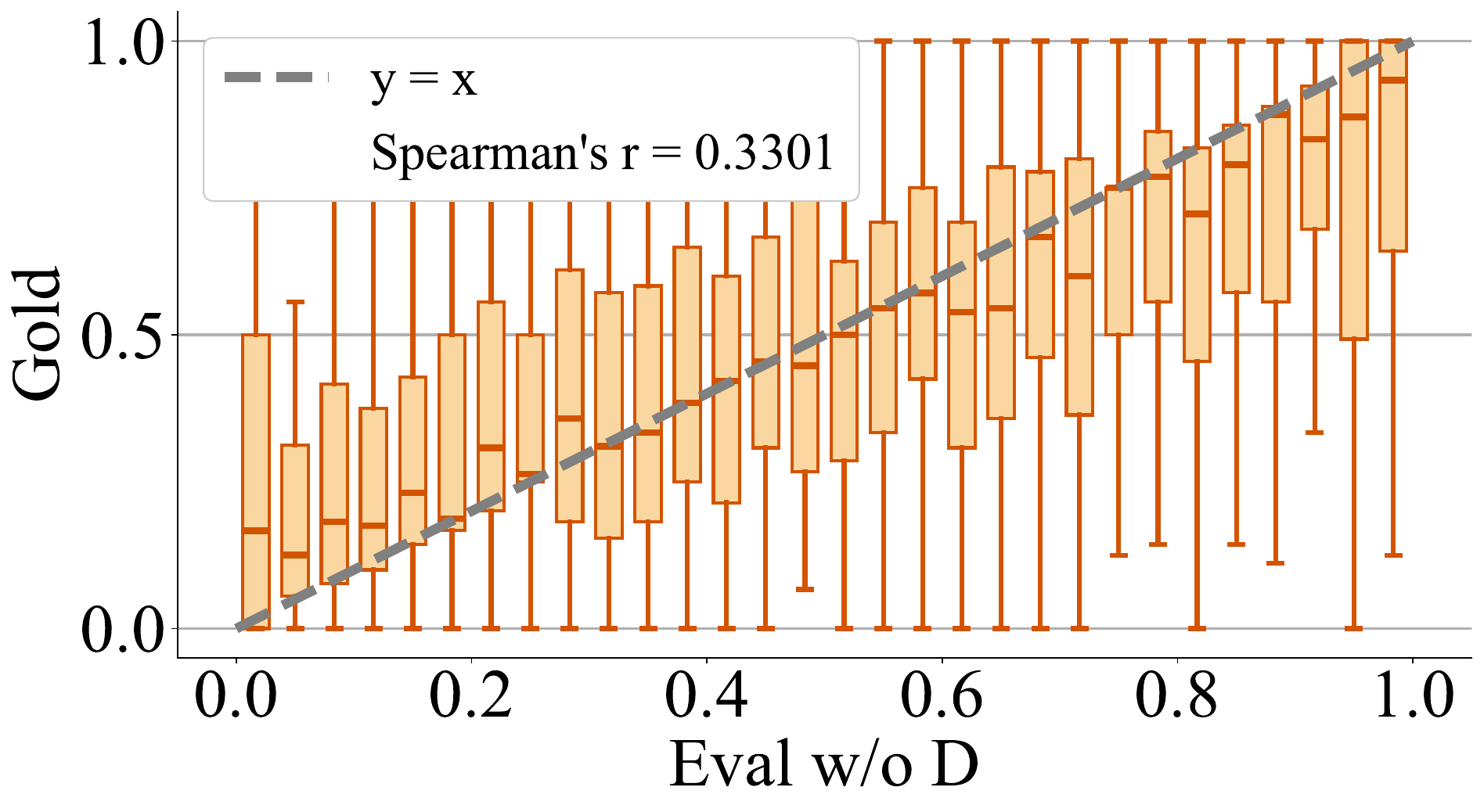}
\end{subfigure}
\hfill
\begin{subfigure}[h]{0.5\linewidth}
\includegraphics[width=\linewidth]{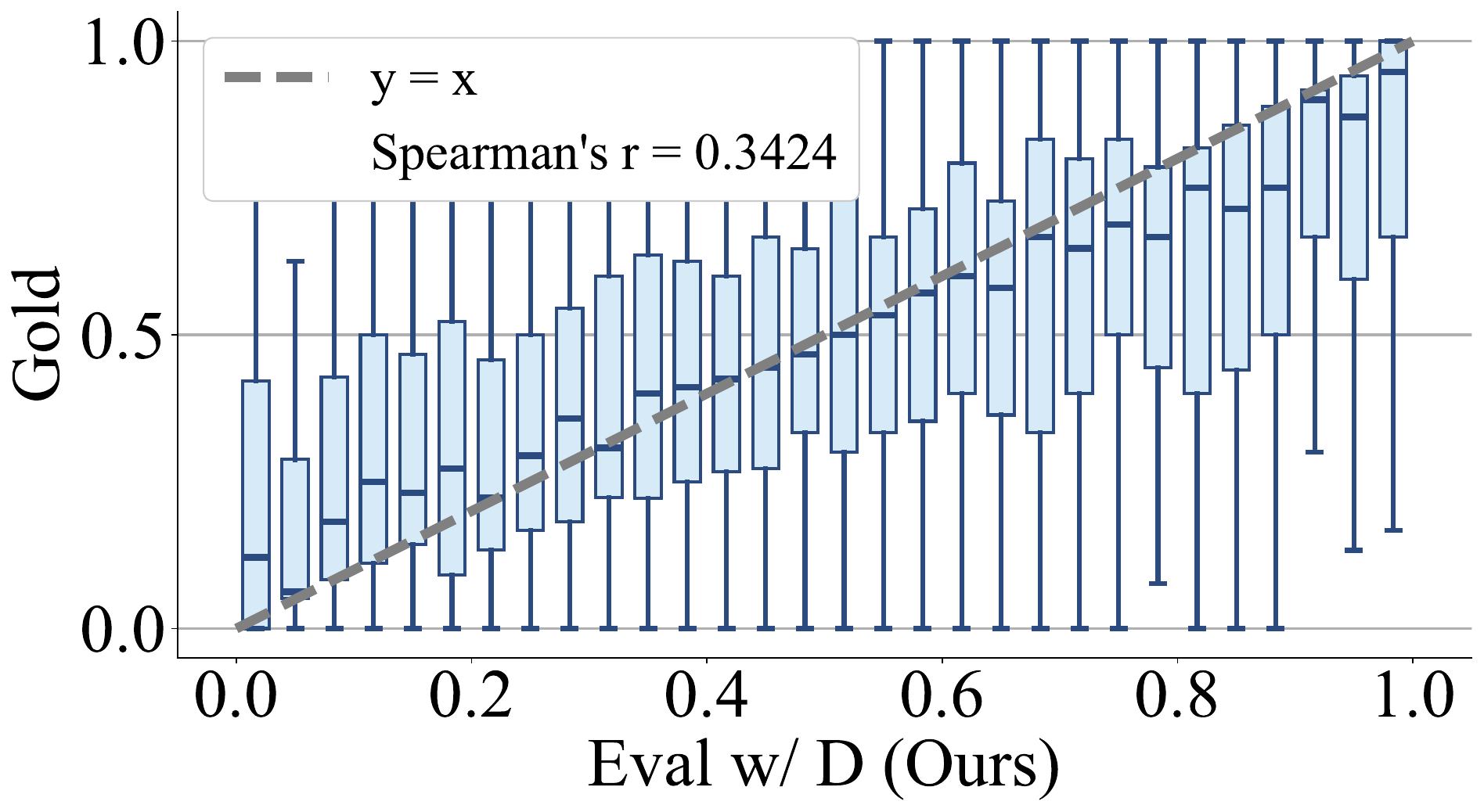}
\end{subfigure}%
\caption{Correlation between our rankings and gold rankings on the $\text{Healthcare}_{\text{Qwen-2.5-7B}}$ dataset. x-axis: Ranking of responses from our model, with (Right) or without (Left) access to $D$. y-axis: For responses that are ranked at the top x\% among the responses to the same query according to our model, the distribution of their gold rankings. Spearman's r: Spearman's ranking correlation.}
\label{fig:rank_corr}
\vspace{-10pt}
\end{figure}
(1) the correlation between \ourmethod's rankings and gold rankings, and (2) whether grounding our rubric on pre-training document $d$ helps. For (2), we revoke access to $D$ and regenerate the rubrics and scores with $\pi_{ref}$ (\texttt{Eval w/o $D$}). We compare the new rankings with our original rankings (\texttt{Eval w/ $D$}). Both use the gold rankings as a reference. We use percentile ranking, so a rank of x means the response is ranked at the top x\% among the responses to the same query.

\textbf{Full correlation is imperfect.} Figure~\ref{fig:rank_corr} shows the correlation between our rankings and gold rankings. Evaluation without grounding (\texttt{Eval w/o $D$}) achieves a moderate positive correlation of 0.3301 with gold rankings. Grounding on pre-training text (\texttt{Eval w/ $D$}) gives a slightly stronger correlation of 0.3424, but still far from perfect.

\textbf{$y_w-y_l$ correlation is strong, leading to correct training signals.} Since our DPO algorithm only uses the highest and lowest ranked responses $y_w$ and $y_l$ for training, a more relevant metric for \ourmethod~is the pairwise ranking accuracy. That is, the percentage of cases where $y_w$ is truly better than $y_l$ according to gold rankings. We show this in Table~\ref{tab:rank_corr_stat}.

Both settings achieve accuracies above 80\%, and \texttt{Eval w/ $D$} achieves an accuracy of 85\%. This
\begin{wraptable}{r}{0.41\linewidth}
    \vspace{-10pt}
    \small
    \centering
    \begin{tabular}{lc}
        \toprule
         & \%$(rk_{y_w}^{gold}\leq rk_{y_l}^{gold})$ \\
         \midrule
         Eval w/o $D$ & 82.04 \\
         Eval w/ $D$ (Ours) & 85.14 \\
         \bottomrule
    \end{tabular}
    \caption{Pairwise Ranking accuracy.}
    \label{tab:rank_corr_stat}
    \vspace{-10pt}
\end{wraptable}
suggests (1) even though global correlation is imperfect, taking the extremes according to our own rankings still gives mostly correct training signals; (2) Grounding rubric evaluation on the pre-training corpus gives stronger correlation with gold rankings and helps reduce reward hacking.

\begin{wraptable}{r}{0.75\linewidth}
    \vspace{-10pt}
    \small
    \centering
    \begin{tabular}{llcccccc}
    \toprule
         & Verifier & $rk_{y_w}$ & $rk_{y_l}$ & $\Delta rk$ & $s_{y_w}$ & $s_{y_l}$ & $\Delta s$ \\
         \midrule
         Eval w/o $D$ & $\pi_{ref}$ & 0 & 100 & 100 & 1.78 & 0.47 & 1.31 \\
         & $\pi_{teacher}$ & 26.59 & 75.35 & 48.76 & 1.01 & 0.63 & 0.38 \\
         \midrule
         Eval w/ $D$ (Ours) & $\pi_{ref}$ & 0 & 100 & 100 & 1.79 & 0.51 & 1.28 \\
         & $\pi_{teacher}$ & 23.59 & 76.50 & 52.91 & 1.04 & 0.63 & 0.41 \\
    \bottomrule
    \end{tabular}
    \caption{Rankings and scores of $y_w$ and $y_l$ from Eval w/o $D$ and Eval w/ $D$, according to either $\pi_{ref}$ or stronger model.}
    \label{tab:ranK_corr_stat_detailed}
    \vspace{-10pt}
\end{wraptable}
In Table~\ref{tab:ranK_corr_stat_detailed}, we further list the detailed statistics of $y_w$ and $y_l$ under both \texttt{Eval w/o $D$} and \texttt{Eval w/ $D$}. For both settings, we compute the average rankings and scores of $y_w$ and $y_l$ under our model and the teacher model. In both settings, $y_w$ is ranked and scored significantly higher than $y_l$ by the teacher model. In addition, the rank difference between $y_w$ and $y_l$ is larger in the \texttt{Eval w/ $D$} setting, showing the benefits of grounding on pre-training text. We provide further analysis and methodology in Appendix~\ref{app:ranking_analysis} and conduct a human study in Appendix~\ref{app:human_study}.
\vspace{-5pt}

% Remove Knowledge

% Use teacher to sythesize queries

% Use teacher to sythesize responses

% Use teacher to sythesize rubrics

% Use teacher to sythesize scores

% Use teacher to sythesize rubrics and scores

% Pairwise Judge

% Correlation between the scores of the student model and the teacher model.

%% file: appendix.tex
\section{Hyperparameter}
\label{app:hyperparameters}

We show the list of hyperparameters in Table~\ref{tab:hyperparameters}.

\begin{table}[]
    \centering
    \small
    \begin{tabular}{llc}
        \toprule
        Stage & Parameter & Value \\
        \midrule
        Sampling & Max Context Length & 32768 \\
                 & $|d|$ & $\in[50, 1024]$ \\
                 & Proposer Max New Tokens & 6144 \\
                 & Proposer Temperature & 1.0 \\
                 & Proposer Top P & 1.0 \\
                 & Solver Max New Tokens & 6144\\
                 & Solver Temperature & 1.0 \\
                 & Solver Top P & 1.0 \\
                 & Rubric Gen Max New Tokens & 8192\\
                 & Rubric Gen Temperature & 0.0 \\
                 & Rubric Gen Top P & 1.0 \\
                 & Ans Grading Max New Tokens & 4096\\
                 & Ans Grading Gen Temperature & 0.0 \\
                 & Ans Grading Gen Top P & 1.0 \\
                 & \# Ques ($I$) & 4096 \\
                 & \# Ques ($I$) Creative Writing & 8192 \\
                 & \# Ques per $d$ & 1 \\
                 & \# Ans per Ques ($J$) & 32 \\
                 & Max \# Rub Criteria per Ques ($K$) & 5 \\
        \midrule
        Continuous Pre-training (CPT on $D$) & Scheduler & Cosine \\
                 & Warmup Ratio & 0.1 \\
                 & Optimizer & AdamW \\
                 & Learning Rate & 5e-6 \\
                 & Number of Epochs & 1 \\
                 & Batch Size & 64 \\
        \midrule
        Supervised-Finetuning (SFT on $y_{ref}$) & Scheduler & Cosine \\
                 & Warmup Ratio & 0.1 \\
                 & Optimizer & AdamW \\
                 & Learning Rate & 1e-6 \\
                 & Number of Epochs & 1 \\
                 & Batch Size & 64 \\
        \midrule
        DPO (\ourmethod)      & Scheduler & Cosine \\
                 & Warmup Ratio & 0.1 \\
                 & Optimizer & AdamW \\
                 & Learning Rate & 1e-6 \\
                 & Beta & 0.01 \\
                 & Number of Epochs & 1 \\
                 & Batch Size & 16 \\
        \bottomrule
    \end{tabular}
    \caption{Hyperparameters}
    \label{tab:hyperparameters}
\end{table}

For Continuous Pre-training and Supervise-Finetuning, we initially use a learning rate of 2e-5 but observe significant performance degradation, so we search for the optimal learning rate among $\{1e^{-6}, 2e^{-6},5e^{-6},1e^{-5},2e^{-5}\}$, using a held-out subset of 100 examples from HealthBench as the validation set. For our DPO approach, we search the learning rate among $\{1e^{-6}, 3e^{-6}, 5e^{-6}\}$ and beta among $\{0.1, 0.01\}$.

For rubric generation, to prevent context length overflow, we select ten candidate responses to put into the rubric generation prompt. In particular, we sort the responses by length and then pick one from every $\frac{\#Ans}{10}$ so that the rubric generator sees responses of varying length.

\subsection{Evaluation}

\paragraph{Main Evaluation.} We use the default hyperparameters for the evaluations whenever such defaults are provided. For HealthBench, when we randomly sample 500 examples from the original benchmarks, we keep the proportion of hard examples fixed at 0.2. We use a sampling temperature of 0.05 and a max output length of 4000 for our model, and a temperature of 0.1 and a max output length of 4000 for the judge. For Creative Writing V3, we use a sampling temperature of 0.7 and a max output length of 4000 for our model and a temperature of 0.1 and a max output length of 4096 for the judge. 
% To compute the Elo score, we use a temperature of 0 and a max output length of 16000 for the judge. 
For IFEval, we use a sampling temperature of 0.7 and a max context length of 4096. For AreaHard, we use a sampling temperature of 0 and a max output length of 2048 for our model and a temperature of 0 and a max output length of 16000 for the judge.

\paragraph{OOD Evaluation.} We use lighteval \cite{lighteval} for OOD Evaluations. Evaluations are 0-shot across different benchmarks. For math benchmarks (GSM8K, Math500, AIME2024, AIME2025), we use a sampling temperature of 0.6, a max context length of 32768, and top p = 0.95. We report avg@4 for GSM8K and Math500 and avg@64 for AIME2024 and AIME2025. For other benchmarks, we use a sampling temperature of 0 and a max context length of 4096.

\paragraph{Analysis.} For our analysis with a stronger teacher model $\pi_{teacher}$, we set $\pi_{teacher}$ to GPT-4o-mini.

\section{\ourmethod~Algorithm}
\label{app:pipeline}

We show the pseudo code of \ourmethod~in Algorithm~\ref{ago:pipeline}.

\begin{algorithm}
\caption{\ourmethod~Pipeline}
\small
\label{ago:pipeline}
\begin{algorithmic}[1]
\Require Reference LLM $\pi_{ref}$; Pre-training Corpus $D$; \#Queries $I$; \#Responses /Q $J$; \#Max Rubric Citeria/Qus $K$; Task $t$; Query Synthesis Prompt $P_{t}^{proposer}$; Response Generation Prompt $P_{t}^{solver}$; Rubric Generation Prompt $P^{rubric}$; Response Grading Prompt $P^{grader}$
\singlecomment{Sampling}
\State Initialize $\text{RawDataset} \gets \{\}$
\For{$i = 1$ to $I$}

    \singlecomment{Query Synthesis}
    
    \State Sample $d \sim D$
    
    \State Sample $(x, y_{ref}) \sim \pi_{ref}(\cdot|P_{t}^{qus}(d))$

    \singlecomment{Response Generation}
    
    \State Sample $\{y_j\}_{j=1}^{J} \sim \pi_{ref}(\cdot|P_{t}^{ans}(x))$

    \singlecomment{Rubric Generation}

    \State Draw sample responses $Y_{rubric} \subset (\{y_j\}_{j=1}^{J})$ \Comment{Prevent context overflow}

    \State Sample $r \sim \pi_{ref}(\cdot|P^{rub}(d,x, y_{ref}, Y_{rubric}))$ with $|r|\leq K$

    \singlecomment{Response Grading}
    \For{$j = 1$ to $J$}
    
        \State Sample $e_j \sim \pi_{ref}(\cdot|P^{grade}(x, r, y_j))$
        
        \State $\{w_k\}_{k=1}^{K'} \gets $ GetWeights($r$)
        
        \State $\{s_j^{k}\}_{k=1}^{K'} \gets $ GetScores($e$)
        
        \State $s_j \gets \frac{\sum_{k=1}^{K'} w_k \cdot s_{j}^{k}}{\sum_{k=1}^{K} w_k}$
    \EndFor

    \State $\text{RawDataset} \gets \text{RawDataset} \cup\{(d,x,y_{ref}, \{y_j\}_{j=1}^{J}, r, \{(e_j, s_j)\}_{j=1}^{J})\}$
\EndFor

\singlecomment{Filtering}

\State $\text{FilteredDataset} \gets \{\}$

\For{example in RawDataset}
    \State validIndices $\gets \{\}$
    \For{$j$ in $\{1,\cdots,J\}$}
        \If{ResponseValid($y_j$) and EvaluationValid($e_j$)} \Comment{Not extractable, format incorrect, etc.}
            \State validIndices $\gets \text{validIndices}\cup\{j\}$
        \EndIf
    \EndFor
    \State ResetResponsesAndEvaluations(example, $\{y_j\}_{j\in \text{validIndices}}$, $\{e_j\}_{j\in \text{validIndices}}$)
    \State isvalid $\gets True$
    \For{component in examples}
        \If{not CheckValidity(component)} \Comment{Not extractable, format incorrect, etc.}
            \State isvalid $\gets False$
            \State \textbf{Break}
        \EndIf
    \EndFor
    \If{isvalid}
        \State $\text{FilteredDataset} \gets \text{FilteredDataset} \cup \{\text{example}\}$
    \EndIf
\EndFor

\singlecomment{Pairing}

\State $\text{DPODataset} \gets \{\}$

\For{example in FilteredDataset}
    \State $x \gets$ GetQuery(example)
    \State $\{(y_j,s_j)\}_{j=1}^{J'} \gets$ GetResponsesAndScores(example)
    \State $y_w \gets argmax_{y_j}(s_j)$
    \State $y_l \gets argmin_{y_j}(s_j)$
    \If{$y_w>y_l$ and $||y_w| - |y_l||\leq 100$}
        \State $\text{DPODataset} \gets \text{DPODataset} \cup \{(x, y_w, y_l)\}$
    \EndIf
\EndFor

\singlecomment{Training}

\State $\pi_{trained} \gets DPO(\pi_{ref}, \text{DPODataset})$

\State \Return $\pi_{trained}$
\end{algorithmic}
\end{algorithm}

\section{Results on Creative Writing V3}
\label{app:res_cw}

\begin{figure}
    \centering
    \includegraphics[width=\linewidth]{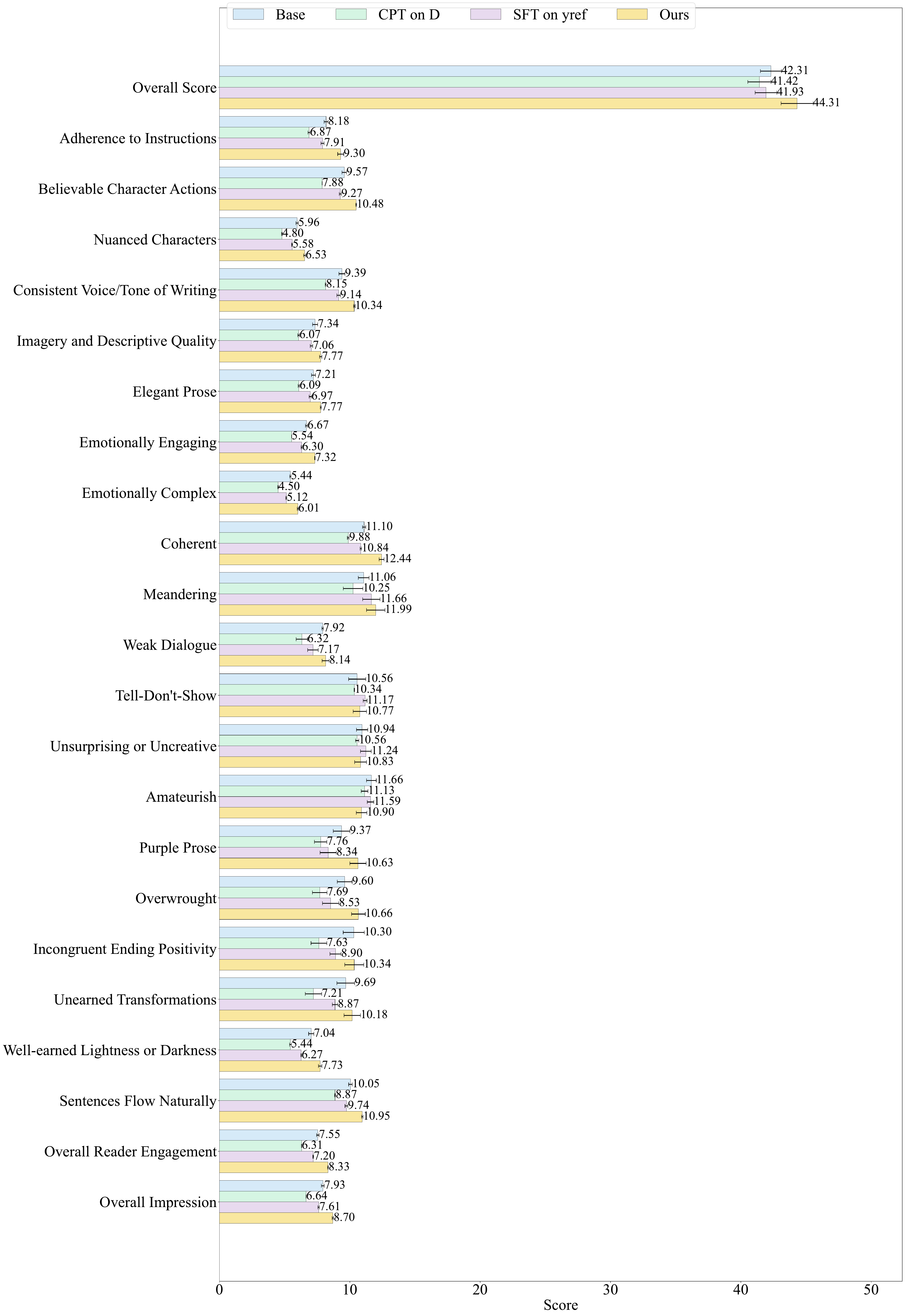}
    \caption{Qwen-2.5-7B Results on Creative Writing V3.}
    \label{fig:cw_qwen_25_7b}
\end{figure}

\begin{figure}
    \centering
    \includegraphics[width=\linewidth]{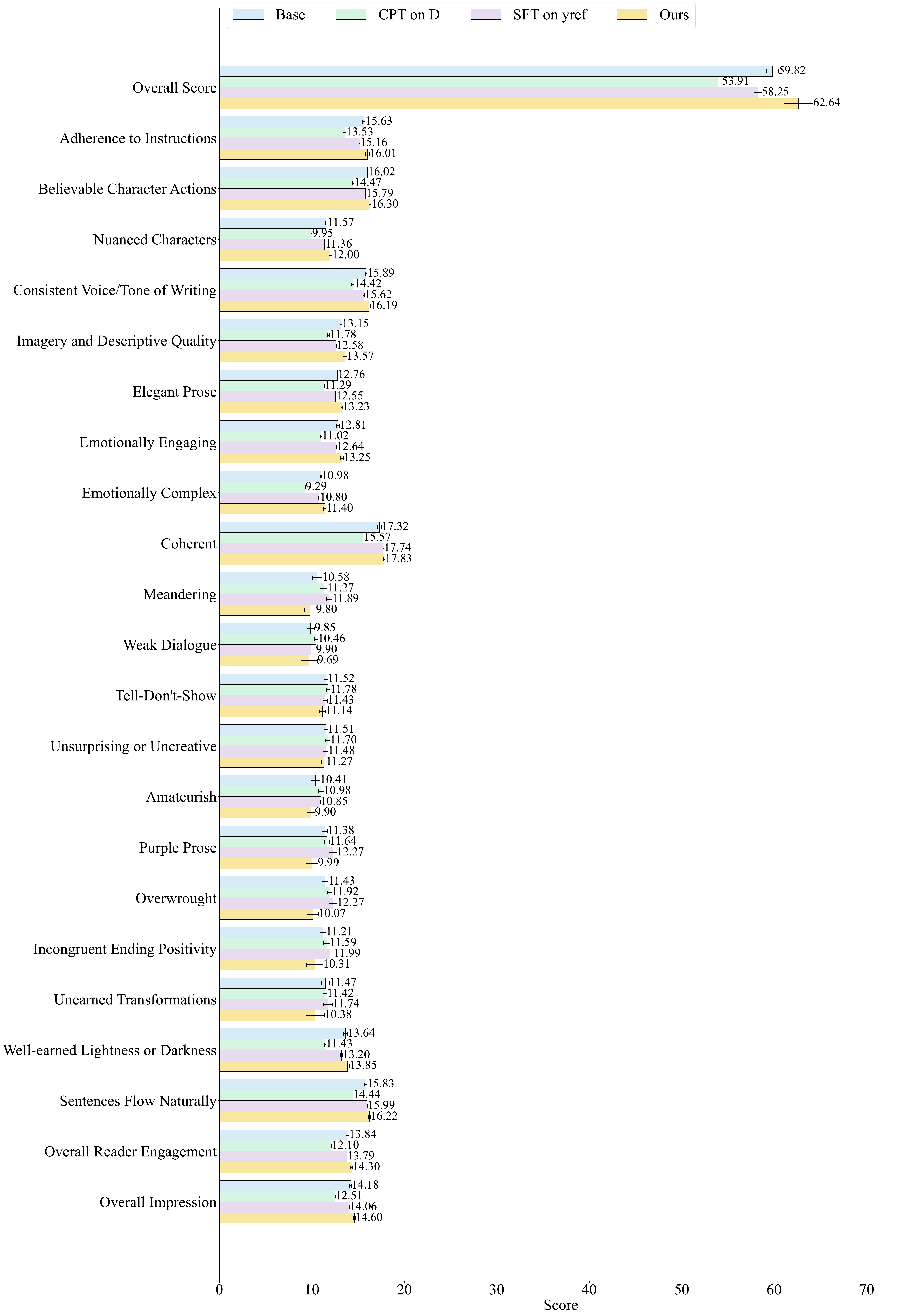}
    \caption{Qwen-2.5-7B-Inst Results on Creative Writing V3.}
    \label{fig:cw_qwen_25_7b_inst}
\end{figure}

We show the per-criterion results for Qwen-2.5-7B in Figure~\ref{fig:cw_qwen_25_7b} and results for Qwen-2.5-7B-Inst in Figure~\ref{fig:cw_qwen_25_7b_inst}. \ourmethod~achieves consistent improvement over the reference models for both cases. Exceptions include "Unsurprising or Uncreative" and "Amateurish" for both models, and "Meandering", "Weak Dialogue", "Tell-Don't-Show", "Purple Prose", "Overwrought", "Incongruent Ending Positivity", and "Unearned Transformations" for Qwen-2.5-7B-Inst. Continuous pre-training on $D$ significantly degrades performance on Qwen-2.5-7B-Inst.

\clearpage
\section{Results on OOD Benchmarks}
\label{app:ood_results}

We show the OOD results of models trained on the Healthcare QA datasets in Table \ref{tab:ood_hc}, Creative Writing datasets in Table~\ref{tab:ood_cw}, and Instruction Following datasets in Table~\ref{tab:ood_if}. Both \ourmethod~and the baselines are able to maintain the performance across different benchmarks compared with the reference model. 

\begin{table}[htp!]
    \small
    \centering
    \resizebox{\linewidth}{!}{%
    \begin{tabular}{lcccccccccc}
    \toprule
         & MedMCQA & NQ & TvQA & TfQA & MMLU-P & GPQA-D & GSM & Math & Aime24 & Aime25 \\
    \midrule
    \multicolumn{11}{c}{Qwen-2.5-7B} \\
    \midrule
         Base & 52.95 & 14.36 & 17.93 & 48.07 & 30.10 & 33.84 & 78.77 & 61.15 & 6.82 & 3.07 \\
         % ~CPT on $D$ & 53.12 & 12.54 & 14.76 & 48.06 & 39.32 & 28.28 & \textbf{80.31} & 61.25 & 5.89 & 3.23 \\
         ~CPT on $D$ & \textbf{54.00} & 14.01 & 16.96 & 48.05 & 34.71 & 33.84 & 79.44 & 61.00 & \textbf{7.76} & 3.28 \\
         ~SFT on $y_{ref}$ & 53.81 & \textbf{15.30} & 16.55 & 48.62 & 43.78 & 31.82 & 78.05 & 59.60 & 6.09 & 3.33 \\
         ~\ourmethod & 53.86 & 14.65 & \textbf{18.96} & \textbf{49.81} & \textbf{44.75} & \textbf{35.35} & \textbf{79.98} & \textbf{64.00} & \textbf{7.76} & \textbf{4.32} \\
    \midrule
    \multicolumn{11}{c}{Qwen-2.5-7B-Inst} \\
    \midrule
         Base & 55.94 & \textbf{15.12} & 18.49 & 62.51 & \textbf{56.57} & \textbf{36.36} & \textbf{90.64} & 75.55 & \textbf{12.40} & 8.07 \\
         % ~CPT on $D$ & 40.74 & 8.50 & 9.22 & 56.40 & 48.92 & 26.26 & 87.02 & 69.80 & 7.76 & 5.47 \\
         ~CPT on $D$ & \textbf{56.99} & 14.60 & \textbf{22.80} & 62.34 & 56.52 & 33.33 & 90.24 & 74.85 & 11.56 & 7.40 \\
         ~SFT on $y_{ref}$ & 56.78 & 14.83 & 17.54 & \textbf{62.71} & 56.20 & 33.84 & 90.16 & \textbf{75.75} & 11.25 & 7.55 \\
         ~\ourmethod & 56.54 & 14.01 & 18.99 & 62.35 & 56.49 & 33.84 & 90.52 & \textbf{75.75} & 12.03 & \textbf{8.23} \\
    \bottomrule
    \end{tabular}
    }
    \caption{OOD Results for models trained on the models trained on the healthcare datasets. NQ: NaturalQuestions; TvQA: TriviaQA; TfQA: TruthfulQA; MMLU-P: MMLU-Pro; GPQA-D: GPQA-Diamond; GSM: GSM8K; Math: MATH500. We use 0-shot evaluation.}
    \label{tab:ood_hc}
    \vspace{-10pt}
\end{table}

\begin{table}[htp!]
    \small
    \centering
    \resizebox{\linewidth}{!}{
    \begin{tabular}{lcccccccccc}
    \toprule
         & MedMCQA & NQ & TvQA & TfQA & MMLU-P & GPQA-D & GSM & Math & Aime24 & Aime25 \\
    \midrule
    \multicolumn{11}{c}{Qwen-2.5-7B} \\
    \midrule
         Base & 52.95 & \textbf{14.36} & 17.93 & 48.07 & 30.10 & 33.84 & 78.77 & 61.15 & 6.82 & 3.07 \\
         % ~CPT on $D$ & 49.13 & 9.03 & 11.75 & 44.96 & \textbf{41.65} & 31.82 & \textbf{82.37} & 60.35 & 6.56 & 3.12 \\
         ~CPT on $D$ & 52.31 & 12.31 & 16.18 & 46.95 & \textbf{36.90} & \textbf{35.35} & 80.02 & 60.55 & 5.99 & 3.39 \\
         ~SFT on $y_{ref}$ & \textbf{53.86} & 14.30 & \textbf{18.89} & \textbf{49.18} & 30.33 & 31.82 & \textbf{81.29} & \textbf{64.05} & 7.76 & 3.54 \\
         ~\ourmethod &  52.95 & 13.48 & 18.14 & 48.49 & 36.86 & 30.81 & 80.17 & 62.26 & \textbf{7.86} & \textbf{4.32} \\
    \midrule
    \multicolumn{11}{c}{Qwen-2.5-7B-Inst} \\
    \midrule
         Base & 55.94 & \textbf{15.12} & 18.49 & 62.51 & 56.57 & 36.36 & 90.64 & \textbf{75.55} & \textbf{12.40} & \textbf{8.07} \\
         % ~CPT on $D$ & 53.33 & 4.87 & 6.15 & 53.21 & 54.65 & 32.83 & 87.98 & 72.55 & 10.05 & 6.82 \\
         ~CPT on $D$ & 56.16 & 14.54 & \textbf{19.35} & 61.51 & 56.26 & 38.28 & 90.73 & 74.15 & 12.08 & 7.19 \\
         ~SFT on $y_{ref}$ & \textbf{56.28} & 15.01 & 18.35 & \textbf{62.87} & 56.49 & \textbf{39.39} & \textbf{90.86} & 75.15 & 11.35 & 6.46 \\
         ~\ourmethod &  56.08 & 14.65 & 18.21 & 62.66 & \textbf{56.62} & 34.85 & 90.18 & 74.75 & 11.98 & 7.81 \\
    \bottomrule
    \end{tabular}
    }
    \caption{OOD Results for models trained on the Creative Writing datasets.}
    \label{tab:ood_cw}
    \vspace{-10pt}
\end{table}

\begin{table}[htp!]
    \small
    \centering
    \resizebox{\linewidth}{!}{
    
    \begin{tabular}{lcccccccccc}
    \toprule
         & MedMCQA & NQ & TvQA & TfQA & MMLU-P & GPQA-D & GSM & Math & Aime24 & Aime25 \\
    \midrule
    \multicolumn{11}{c}{Qwen-2.5-7B} \\
    \midrule
         Base & 52.95 & 14.36 & 17.93 & 48.07 & 30.10 & \textbf{33.84} & 78.77 & 61.15 & 6.82 & 3.07 \\
         % ~CPT on $D$ & \textbf{53.65} & 12.19 & 15.11 & \textbf{48.97} & 38.07 & 27.78 & 82.98 & 61.95 & 6.15 & 3.12 \\
         ~CPT on $D$ & 53.22 & 13.77 & 17.15 & 47.83 & 39.89 & 31.31 & 81.88 & 61.90 & 7.50 & 3.18 \\
         ~SFT on $y_{ref}$ & \textbf{53.55} & \textbf{14.95} & 16.48 & \textbf{48.92} & \textbf{46.00} & \textbf{33.84} & 81.07 & 60.80 & 5.68 & 2.71\\
         ~\ourmethod &  52.98 & 14.07 & \textbf{18.86} & 48.18 & 43.82 & 27.78 & \textbf{83.26} & \textbf{64.25} & \textbf{7.92} & \textbf{4.69} \\
    \midrule
    \multicolumn{11}{c}{Qwen-2.5-7B-Inst} \\
    \midrule
         Base & 55.94 & \textbf{15.12} & 18.49 & 62.51 & \textbf{56.57} & 36.36 & \textbf{90.64} & 75.55 & \textbf{12.40} & 8.07 \\
         % ~CPT on $D$ & 55.30 & 11.20 & \textbf{19.82} & 59.89 & 53.86 & 32.83 & 88.87 & 72.45 & 10.47 & 7.92 \\
         ~CPT on $D$ & 55.96 & 14.19 & \textbf{22.16} & 61.46 & 56.52 & 37.37 & 90.33 & 75.55 & 11.98 & 7.50 \\
         ~SFT on $y_{ref}$ & 56.08 & 15.12 & 16.60 & \textbf{62.75} & 56.14 & 31.31 & 90.45 & 75.40 & 11.25 & 6.51 \\
         ~\ourmethod &  \textbf{56.47} & 14.13 & 18.30 & 62.12 & 56.44 & \textbf{38.38} & 90.24 & \textbf{76.35} & 11.82 & \textbf{9.84} \\
    \bottomrule
    \end{tabular}
    }
    \caption{OOD Results for models trained on the Instruction Following datasets.}
    \label{tab:ood_if}
\end{table}

\clearpage
\section{Statistics of the Synthesized Datasets}
\label{app:stat_dataset}

\subsection{Queries and Rubrics Classification}

\paragraph{Queries.} For each task, we sample 100 examples from both Qwen-2.5-7B and Qwen-2.5-7B-Instruct. The queries for the resulting 200 examples are fed into GPT-4.1-mini to identify the common topics. We then ask GPT-4.1-mini again to categorize each query to exactly 1 of the topics.

\paragraph{Rubrics.} We take the rubric criteria from the 200 examples used in the previous query topic analysis and feed them into GPT-4.1-mini to identify the meta-criteria. We then ask GPT-4.1-mini again to categorize each rubric criterion to exactly 1 of the meta-criteria.

\subsection{Healthcare QA}

\begin{figure}[htp!]
    \centering
    \begin{subfigure}[b]{0.49\textwidth}
        \centering
        \includegraphics[width=\textwidth]{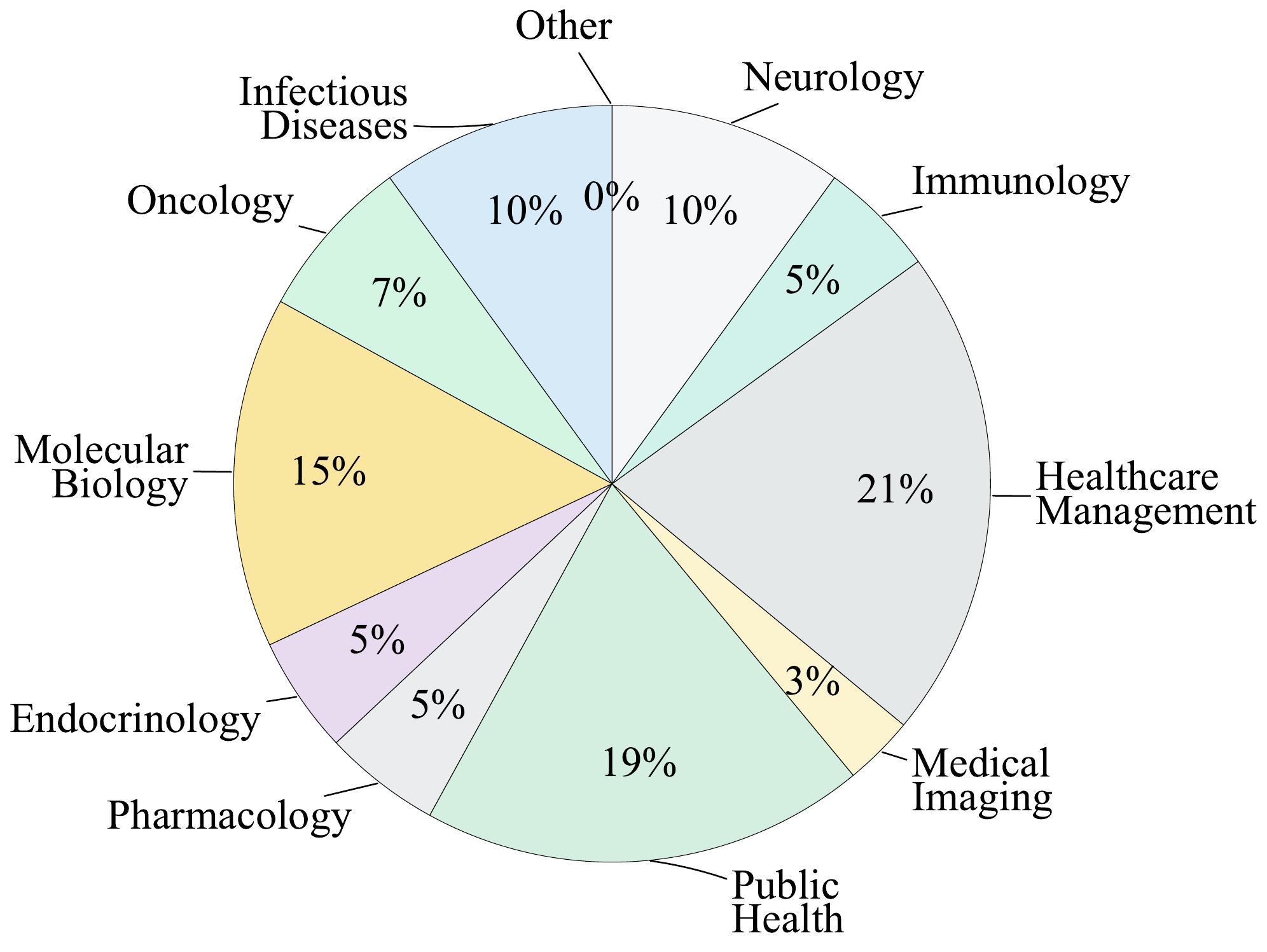}
        \caption{\small Query topics}

    \end{subfigure}
    \hfill
    \begin{subfigure}[b]{0.49\textwidth}  
        \centering 
        \includegraphics[width=\textwidth]{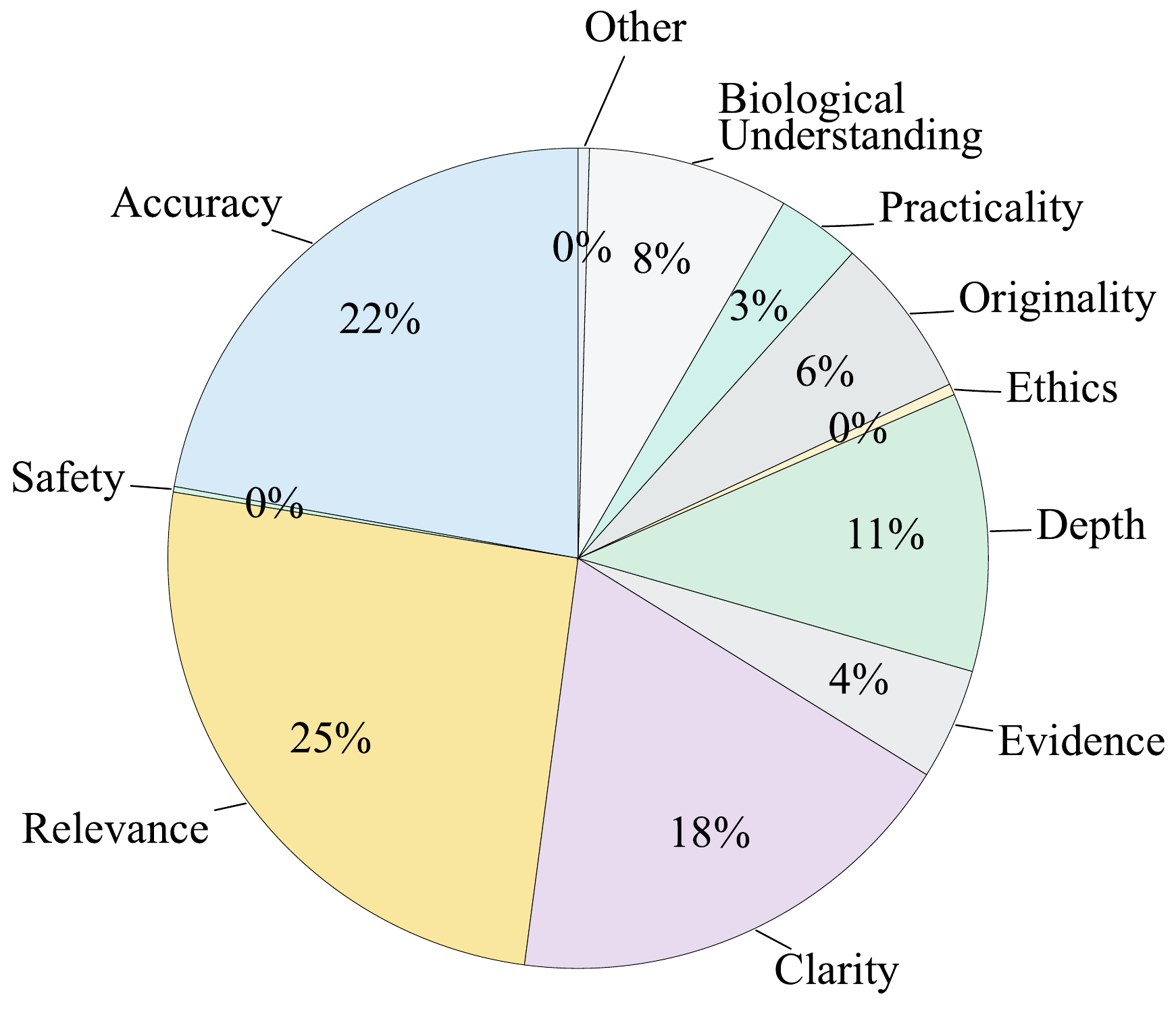}
        \caption{\small Meta rubric criteria}

    \end{subfigure}
    \vskip\baselineskip
    \begin{subfigure}[b]{0.49\textwidth}   
        \centering 
        \includegraphics[width=\textwidth]{assets/bubbles/qwen25_7b_base_hc_ssss_n32_r1.pdf}
        \caption{\small Response score distribution, grouped by queries.}

    \end{subfigure}
    \hfill
    \begin{subfigure}[b]{0.49\textwidth}   
        \centering 
        \includegraphics[width=\textwidth]{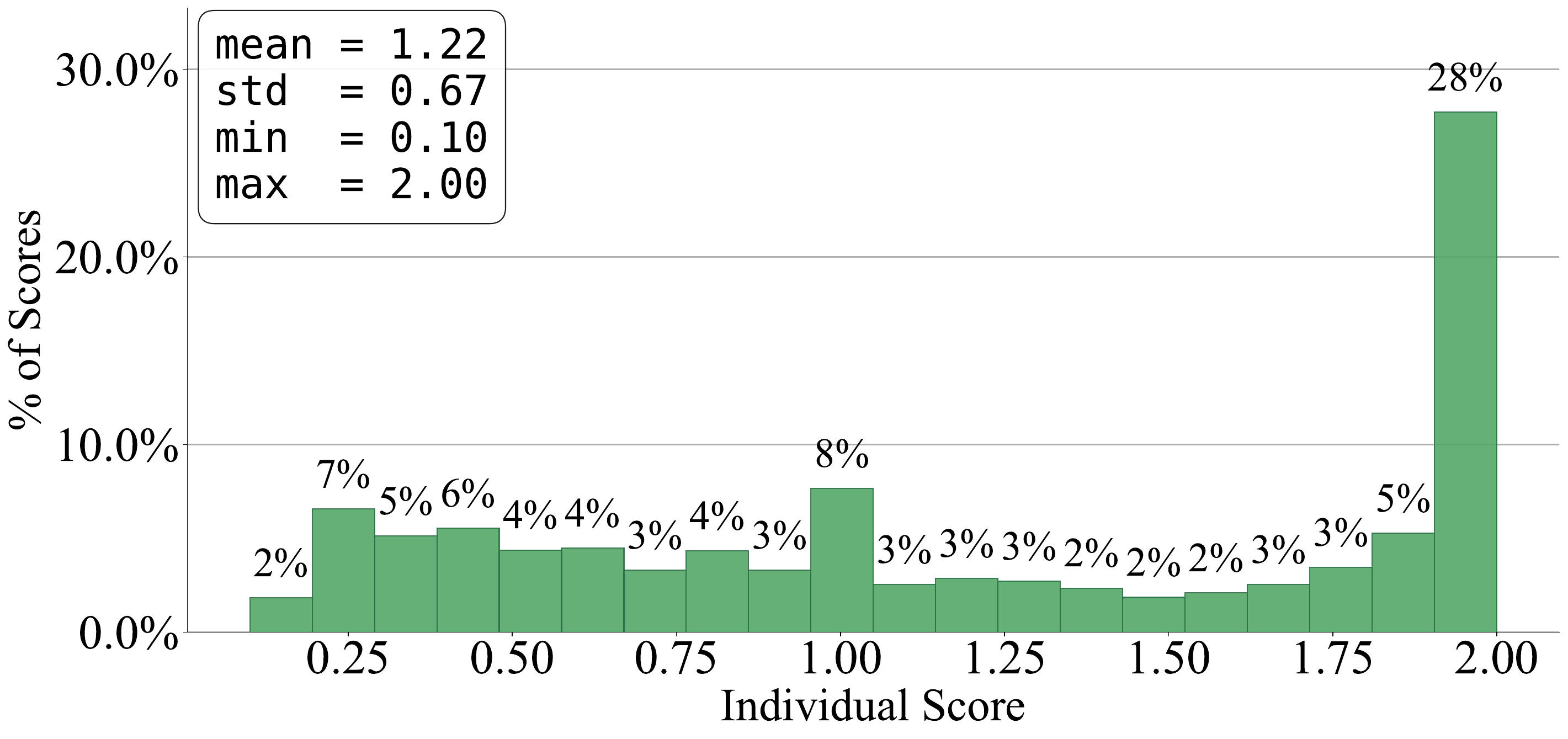}
        \caption{\small Individual response score distribution.}

    \end{subfigure}
    \caption{Statistics for $\text{Healthcare QA}_{\text{Qwen-2.5-7B}}$ dataset.} 
    \label{fig:base_hc_stat}
\end{figure}

\begin{figure}[htp!]
    \centering
    \begin{subfigure}[b]{0.49\textwidth}
        \centering
        \includegraphics[width=\textwidth]{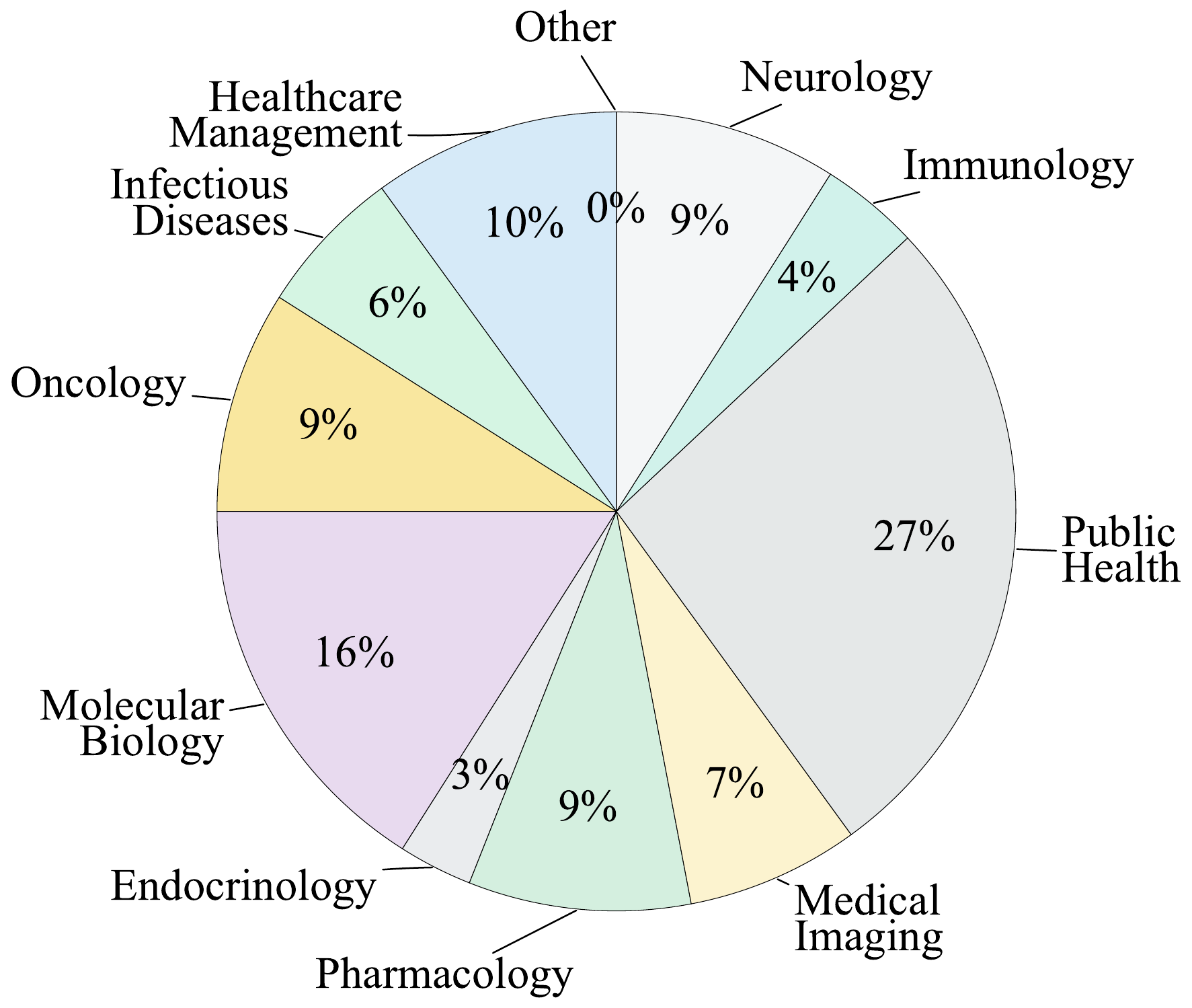}
        \caption{\small Query topics}

    \end{subfigure}
    \hfill
    \begin{subfigure}[b]{0.49\textwidth}  
        \centering 
        \includegraphics[width=\textwidth]{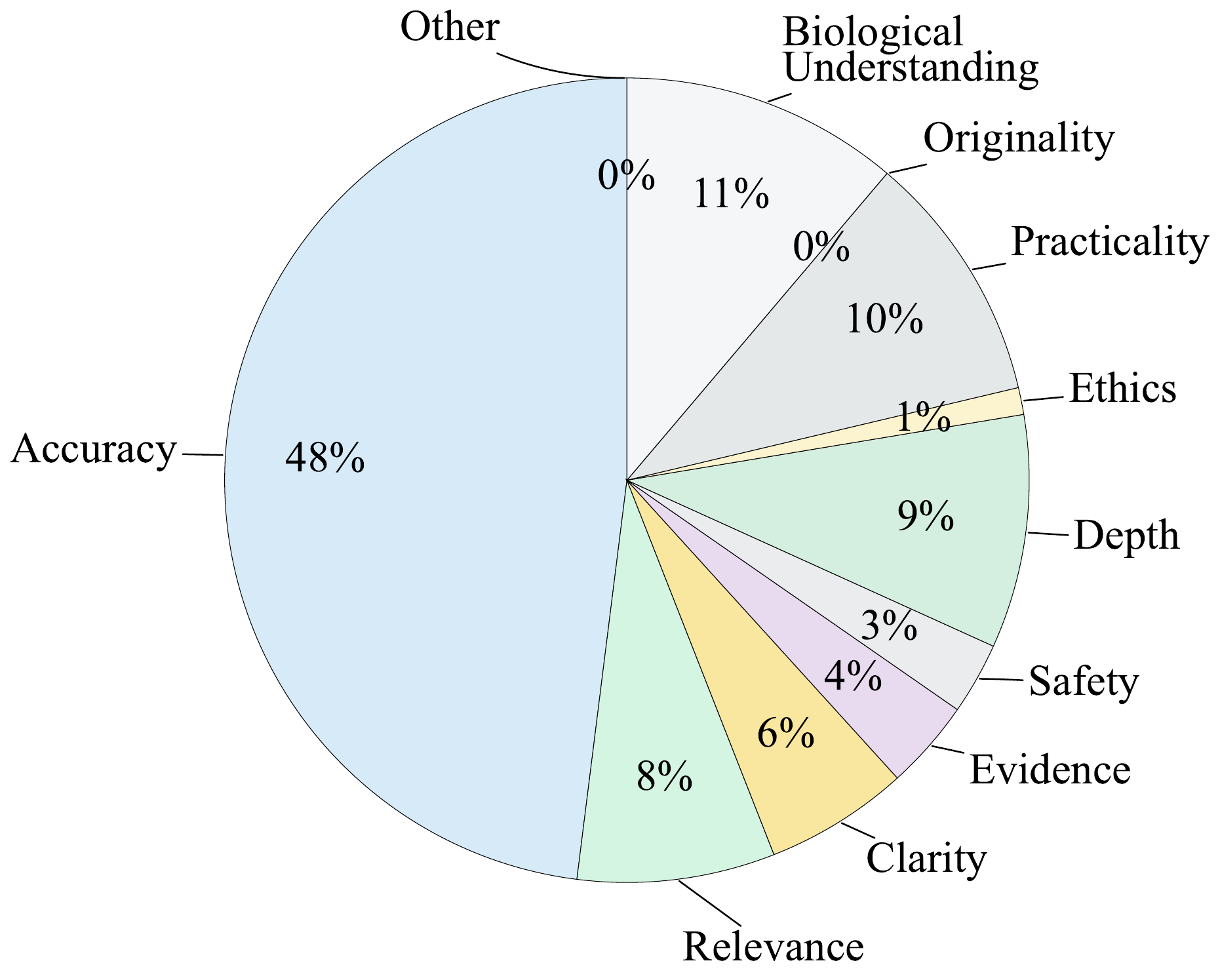}
        \caption{\small Meta rubric criteria}

    \end{subfigure}
    \vskip\baselineskip
    \begin{subfigure}[b]{0.49\textwidth}   
        \centering 
        \includegraphics[width=\textwidth]{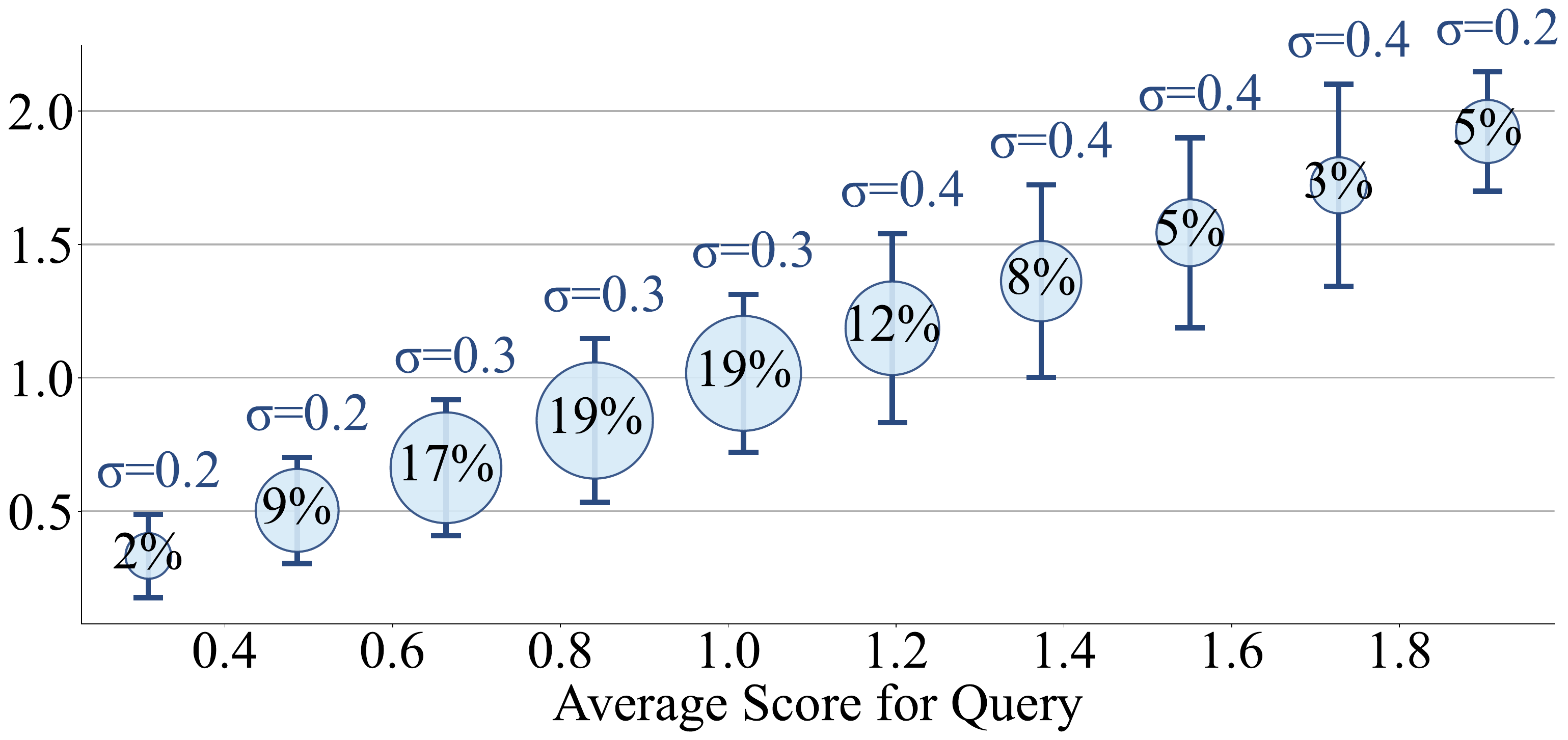}
        \caption{\small Response score distribution, grouped by queries.}

    \end{subfigure}
    \hfill
    \begin{subfigure}[b]{0.49\textwidth}   
        \centering 
        \includegraphics[width=\textwidth]{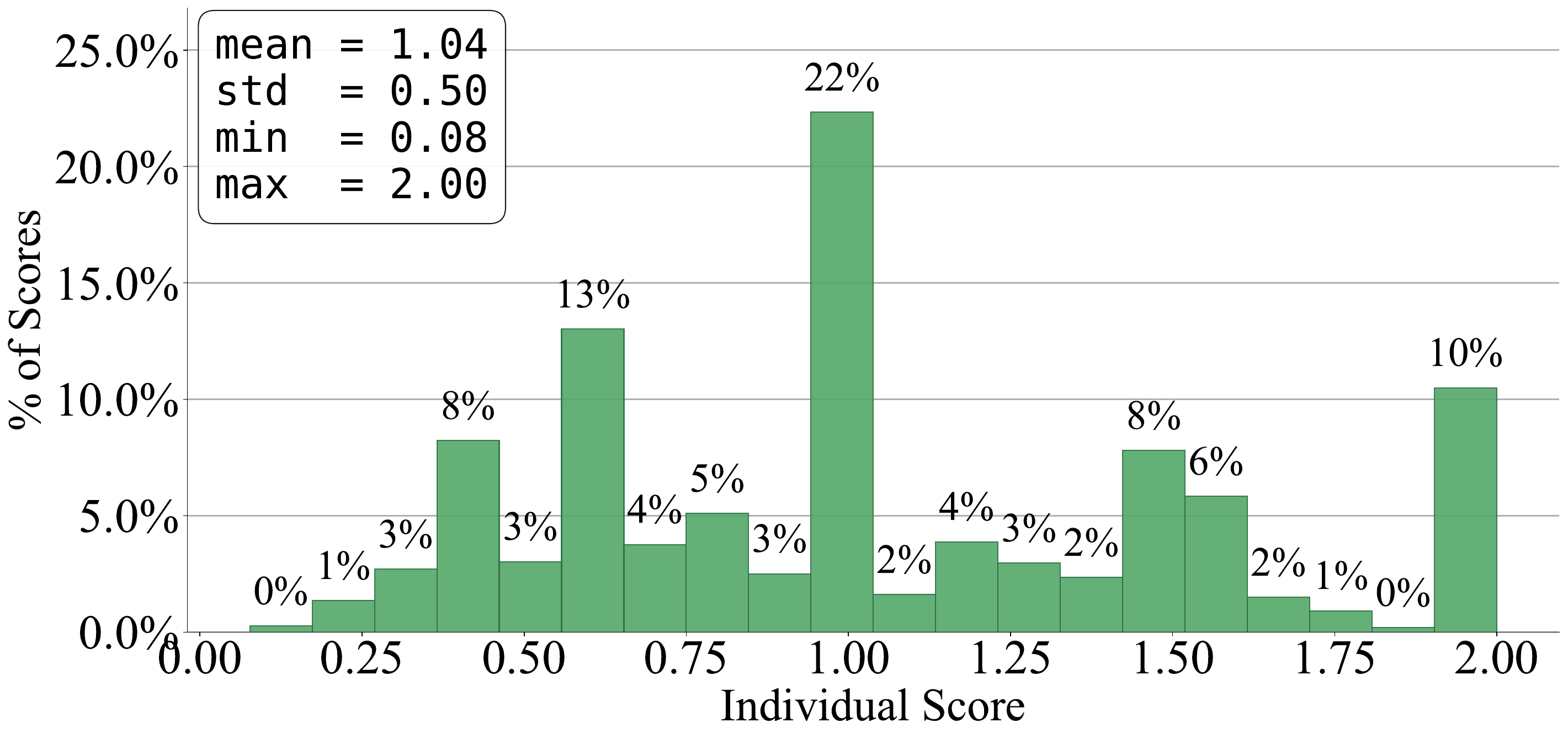}
        \caption{\small Individual response score distribution.}

    \end{subfigure}
    \caption{Statistics for $\text{Healthcare QA}_{\text{Qwen-2.5-7B-Inst}}$ dataset} 
    \label{fig:inst_hc_stat}
\end{figure}

We show the statistics for our synthesized datasets $\text{HealthcareQA}_{\text{Qwen-2.5-7B}}$ in Figure~\ref{fig:base_hc_stat} and $\text{HealthcareQA}_{\text{Qwen-2.5-7B-Inst}}$ in Figure~\ref{fig:inst_hc_stat}.

\subsection{Creative Writing}

\begin{figure}[htp!]
    \centering
    \begin{subfigure}[b]{0.49\textwidth}
        \centering
        \includegraphics[width=\textwidth]{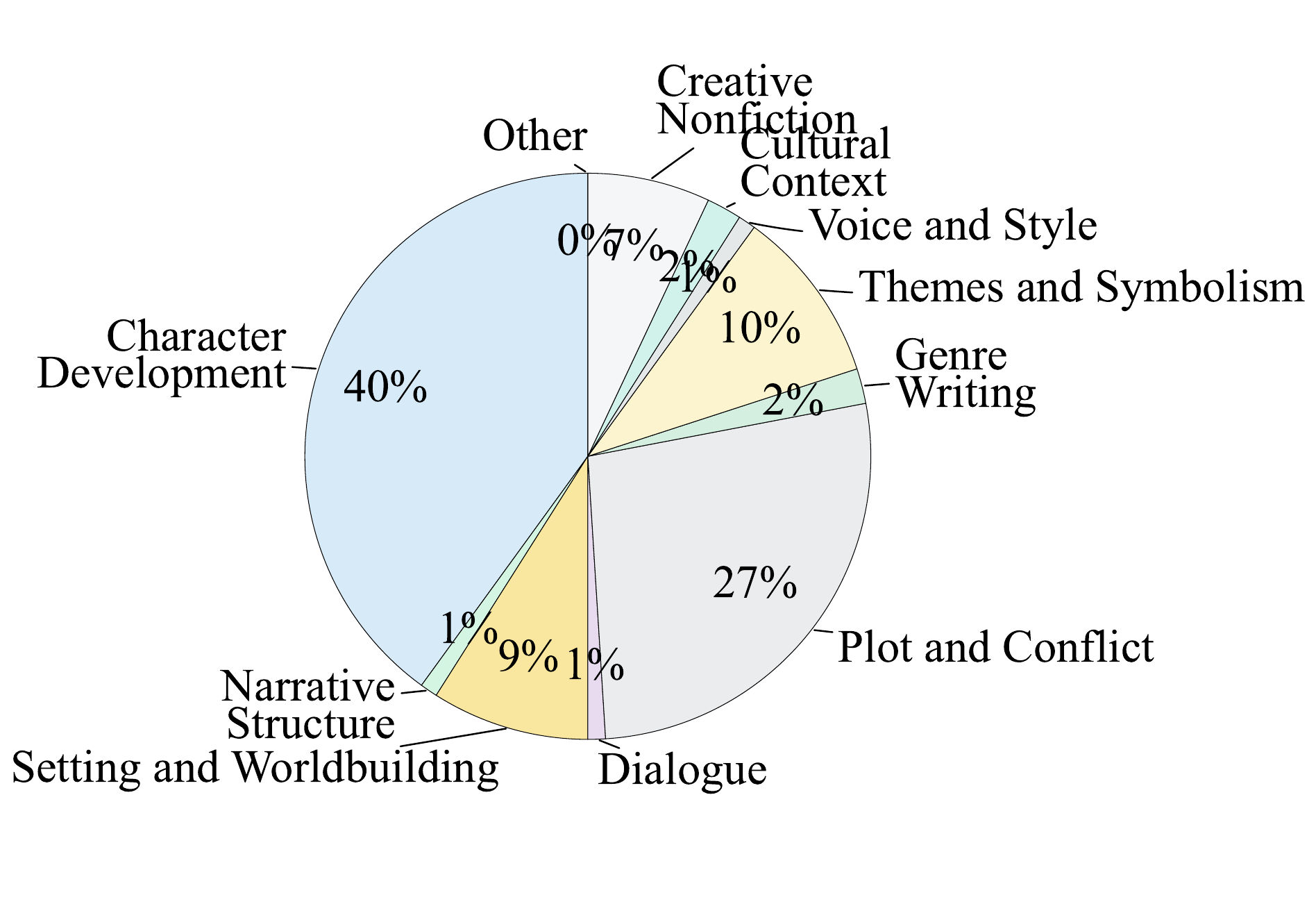}
        \caption{\small Query topics}

    \end{subfigure}
    \hfill
    \begin{subfigure}[b]{0.49\textwidth}  
        \centering 
        \includegraphics[width=\textwidth]{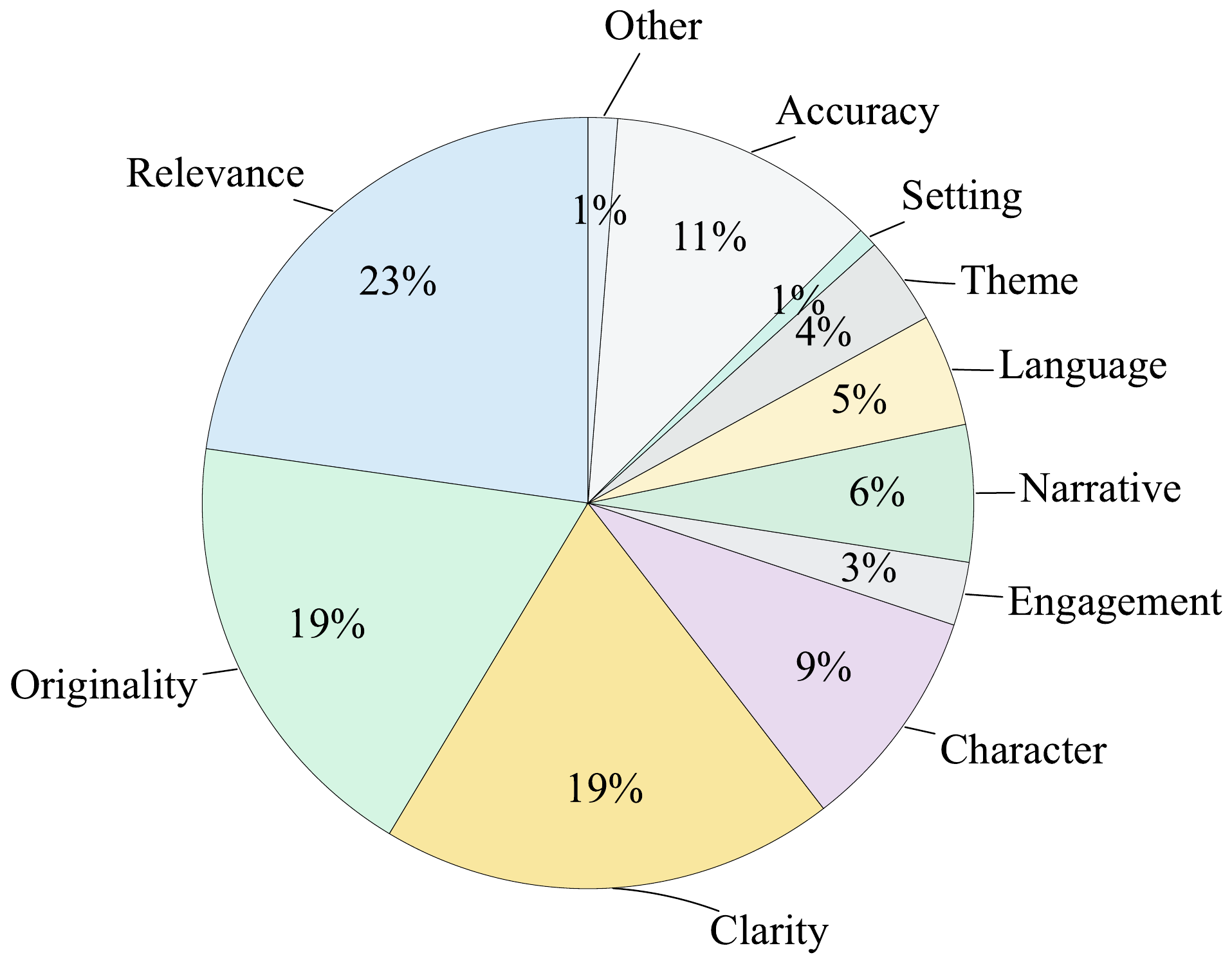}
        \caption{\small Meta rubric criteria}

    \end{subfigure}
    \vskip\baselineskip
    \begin{subfigure}[b]{0.49\textwidth}   
        \centering 
        \includegraphics[width=\textwidth]{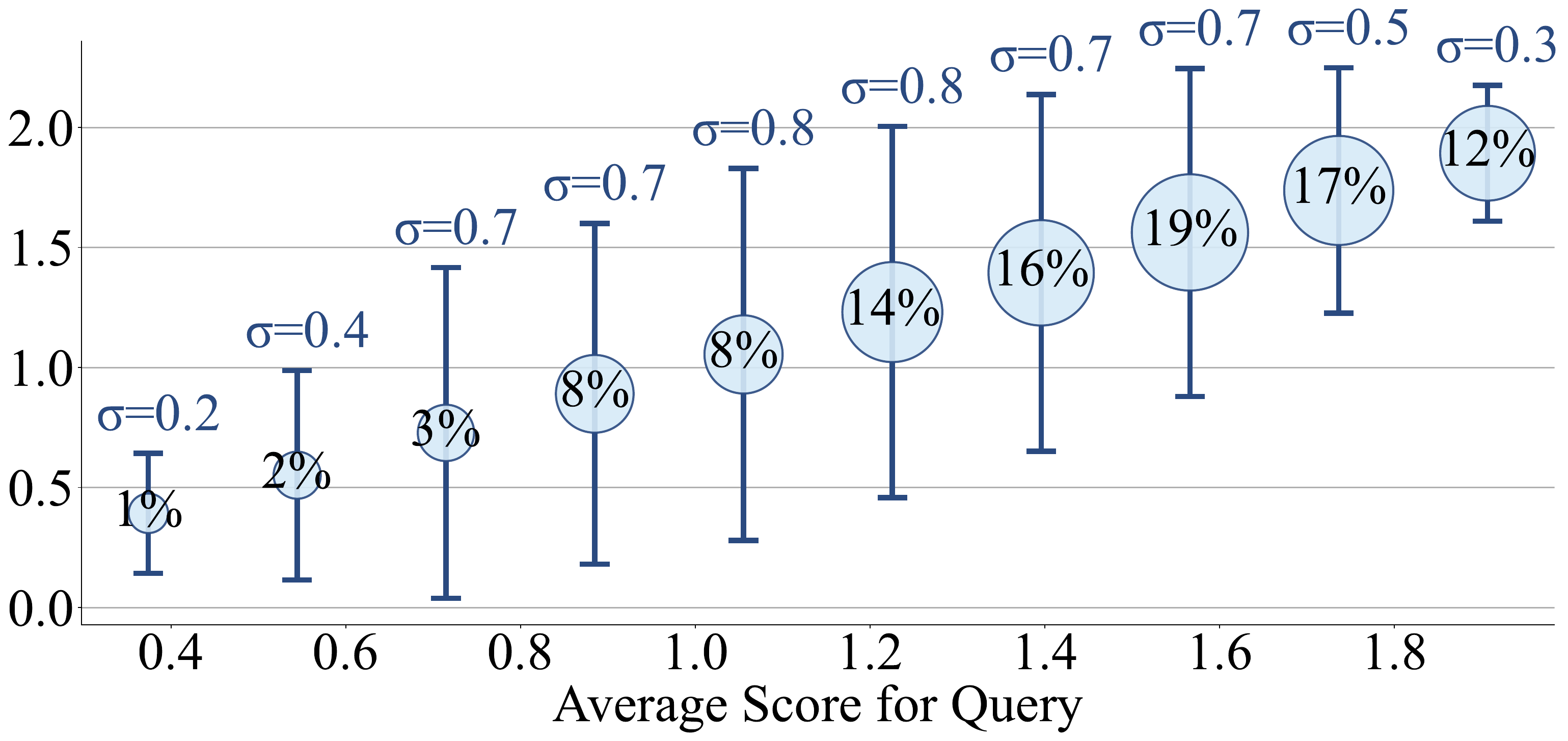}
        \caption{\small Response score distribution, grouped by queries.}

    \end{subfigure}
    \hfill
    \begin{subfigure}[b]{0.49\textwidth}   
        \centering 
        \includegraphics[width=\textwidth]{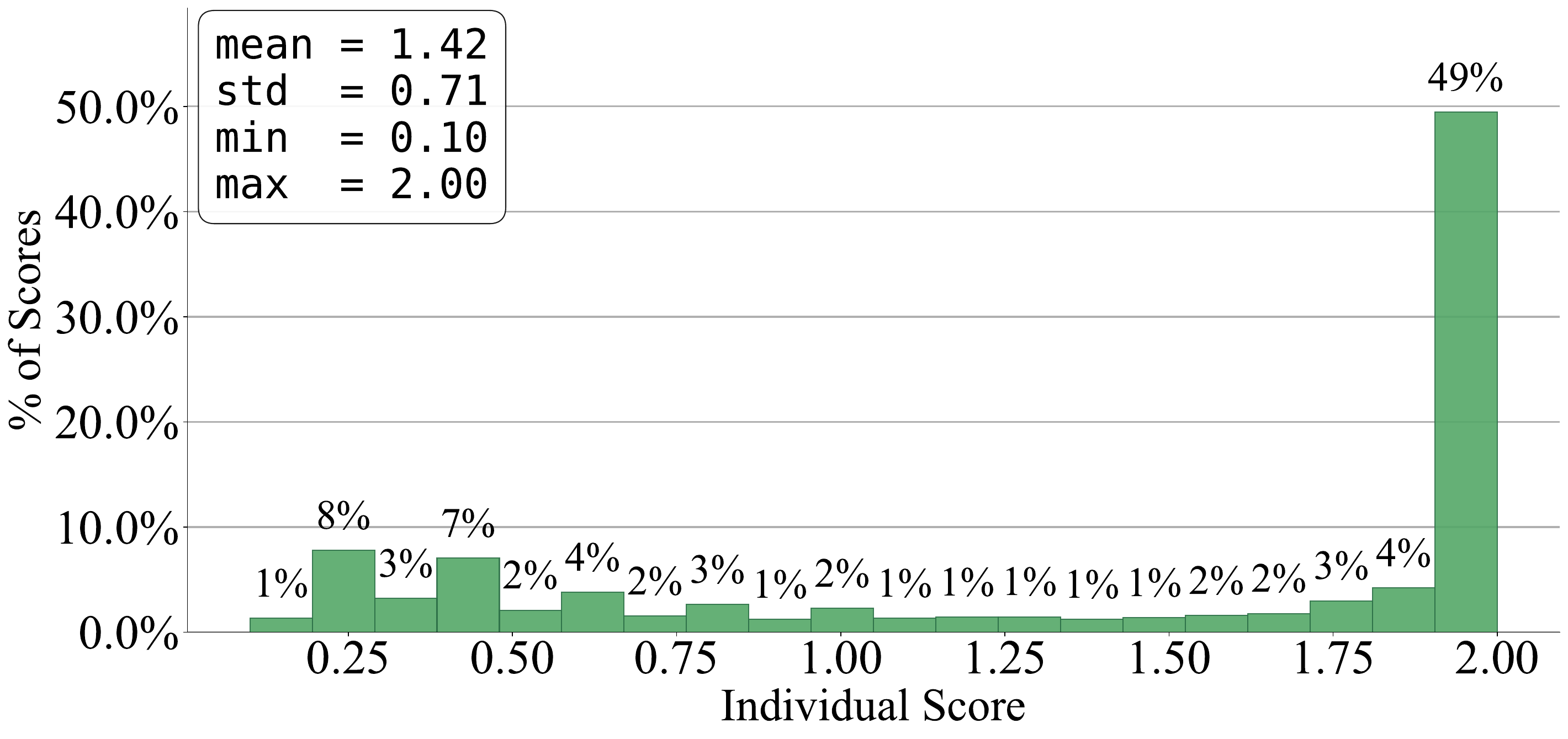}
        \caption{\small Individual response score distribution.}

    \end{subfigure}
    \caption{Statistics for $\text{Creative Writing}_{\text{Qwen-2.5-7B}}$ dataset.} 
    \label{fig:base_cw_stat}
\end{figure}

\begin{figure}[htp!]
    \centering
    \begin{subfigure}[b]{0.49\textwidth}
        \centering
        \includegraphics[width=\textwidth]{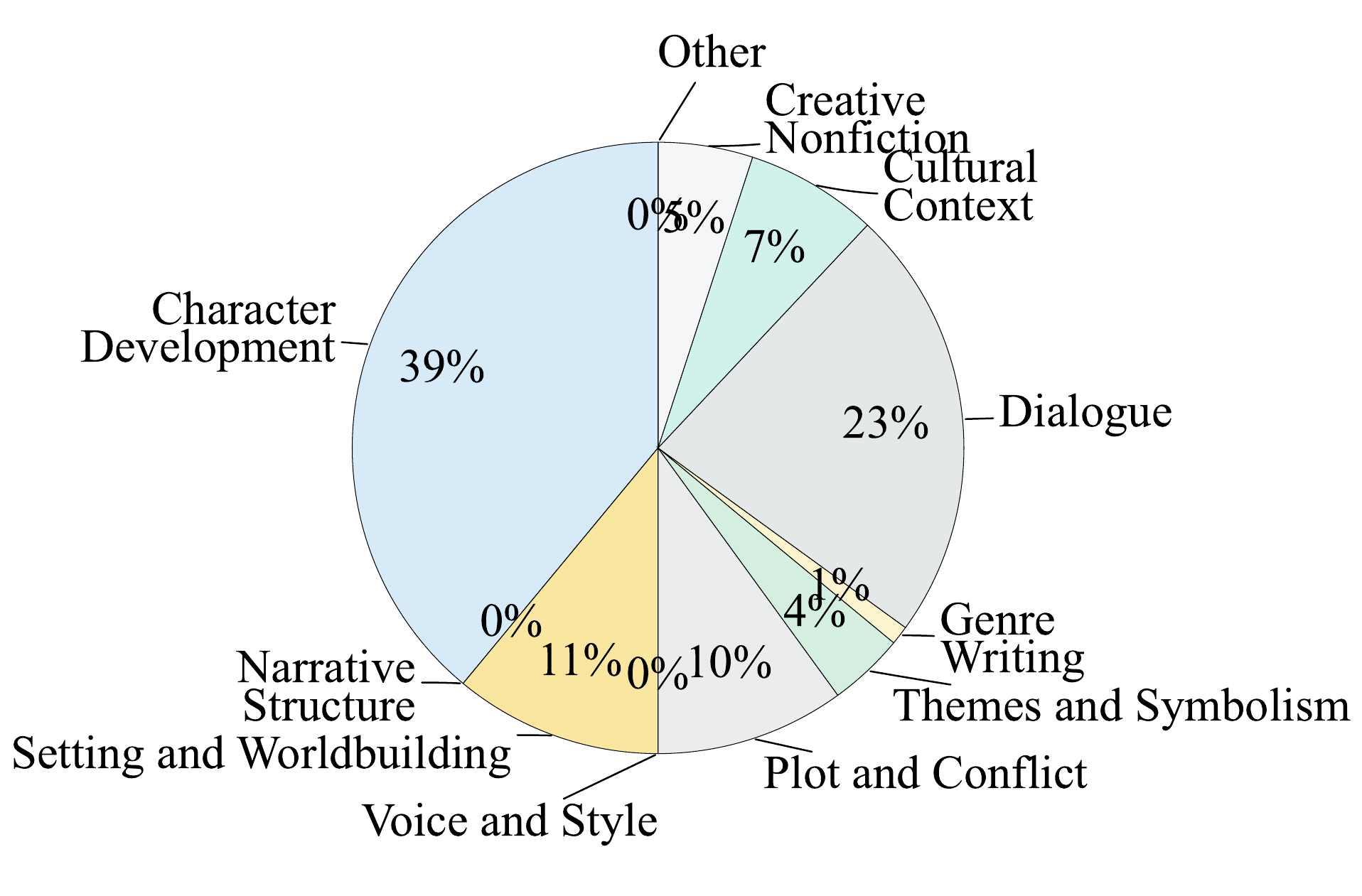}
        \caption{\small Query topics}

    \end{subfigure}
    \hfill
    \begin{subfigure}[b]{0.49\textwidth}  
        \centering 
        \includegraphics[width=\textwidth]{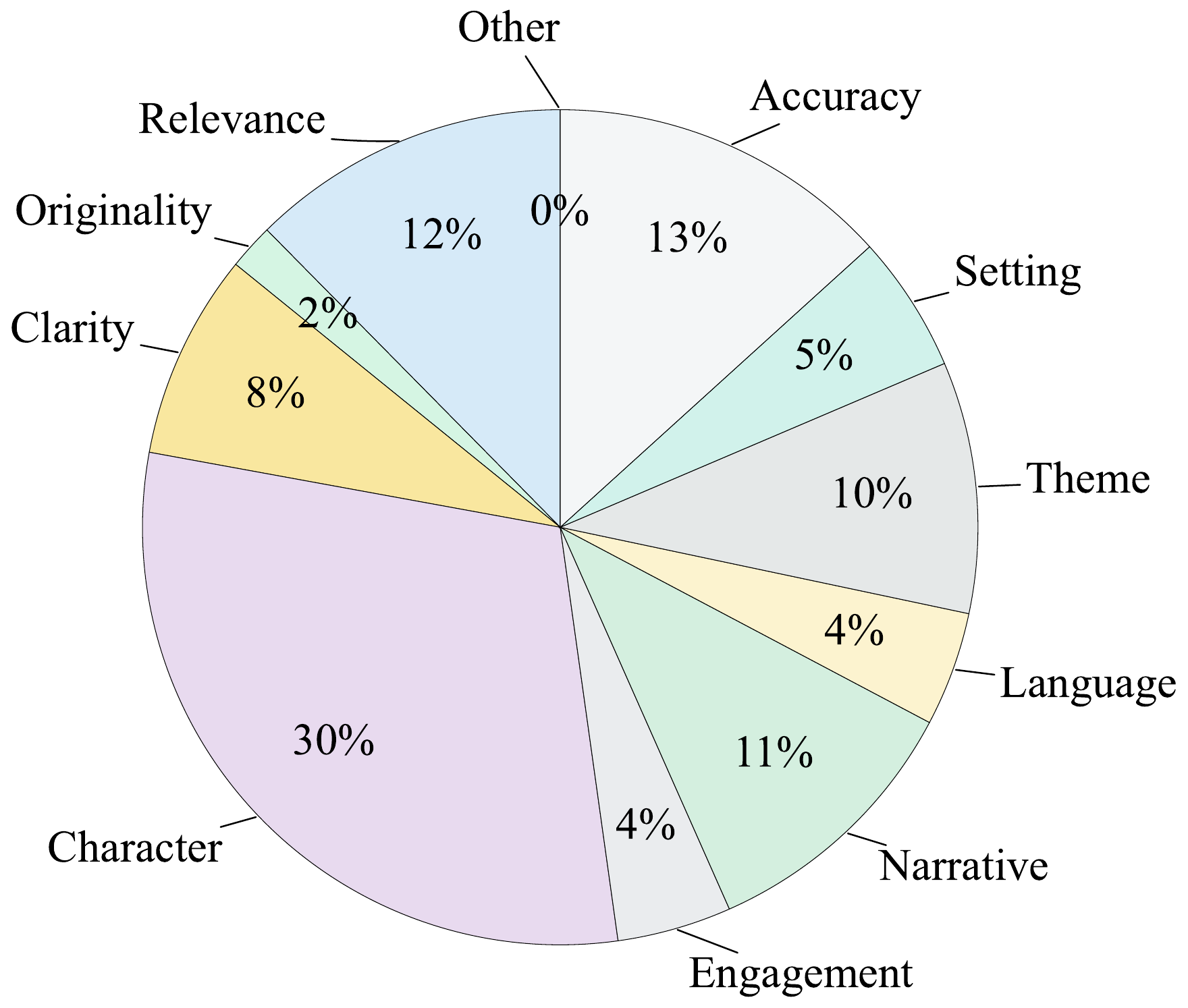}
        \caption{\small Meta rubric criteria}

    \end{subfigure}
    \vskip\baselineskip
    \begin{subfigure}[b]{0.49\textwidth}   
        \centering 
        \includegraphics[width=\textwidth]{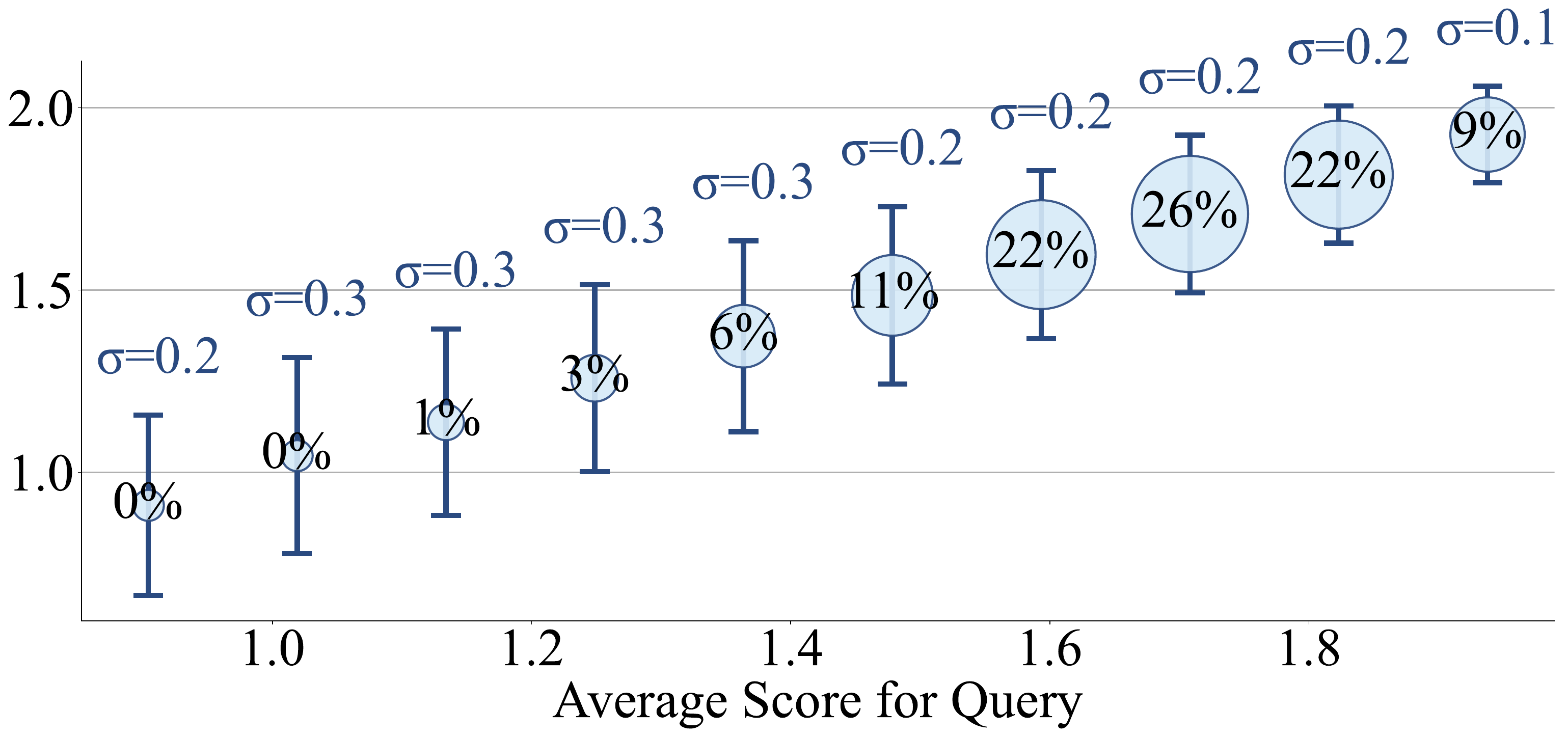}
        \caption{\small Response score distribution, grouped by queries.}

    \end{subfigure}
    \hfill
    \begin{subfigure}[b]{0.49\textwidth}   
        \centering 
        \includegraphics[width=\textwidth]{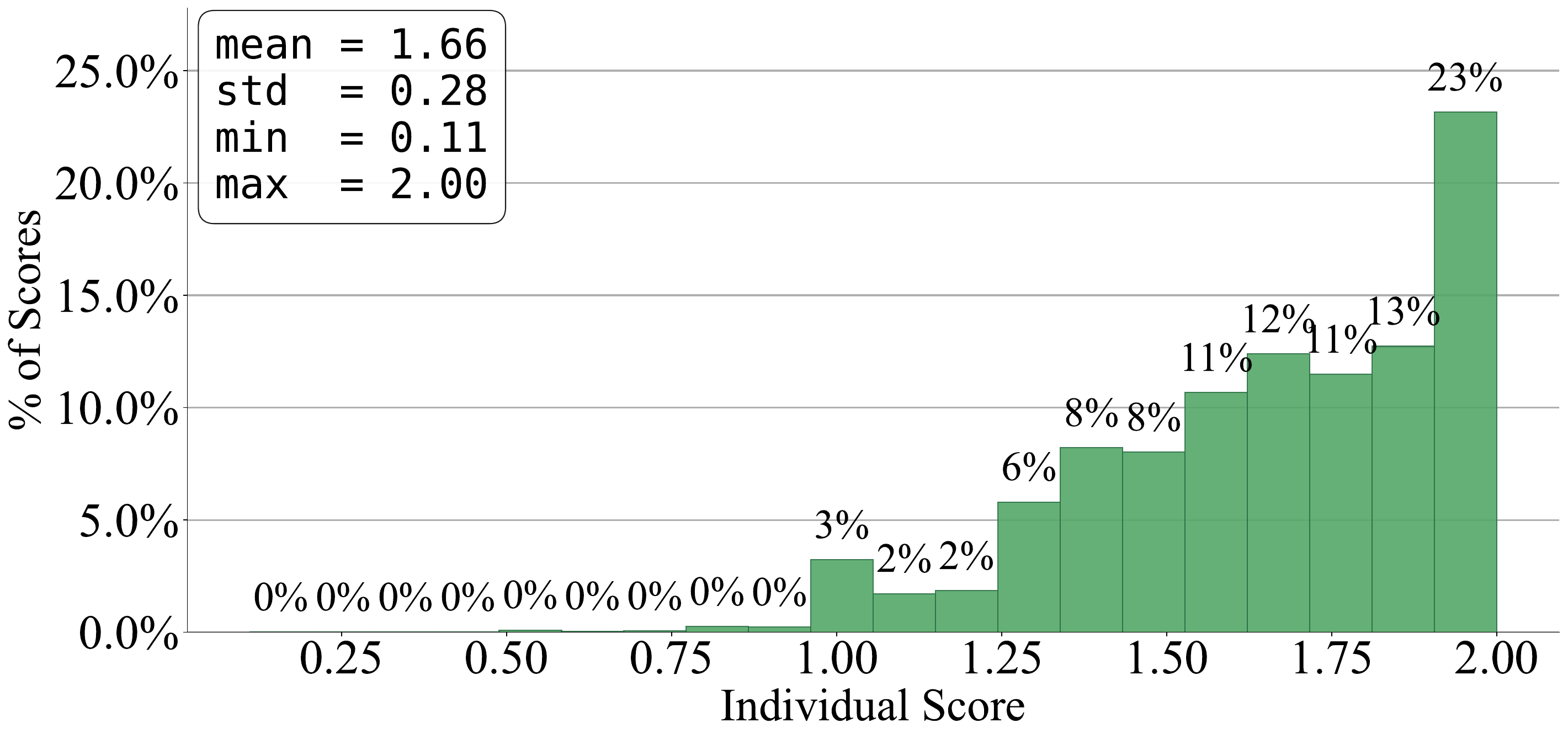}
        \caption{\small Individual response score distribution.}
 
    \end{subfigure}
    \caption{Statistics for $\text{Creative Writing}_{\text{Qwen-2.5-7B-Inst}}$ dataset} 
    \label{fig:inst_cw_stat}
\end{figure}

We show the statistics for our synthesized datasets $\text{CreativeWriting}_{\text{Qwen-2.5-7B}}$ in Figure~\ref{fig:base_cw_stat} and $\text{CreativeWriting}_{\text{Qwen-2.5-7B-Inst}}$ in Figure~\ref{fig:inst_cw_stat}.

\subsection{Instruction Following}

\begin{figure}[htp!]
    \centering
    \begin{subfigure}[b]{0.49\textwidth}
        \centering
        \includegraphics[width=\textwidth]{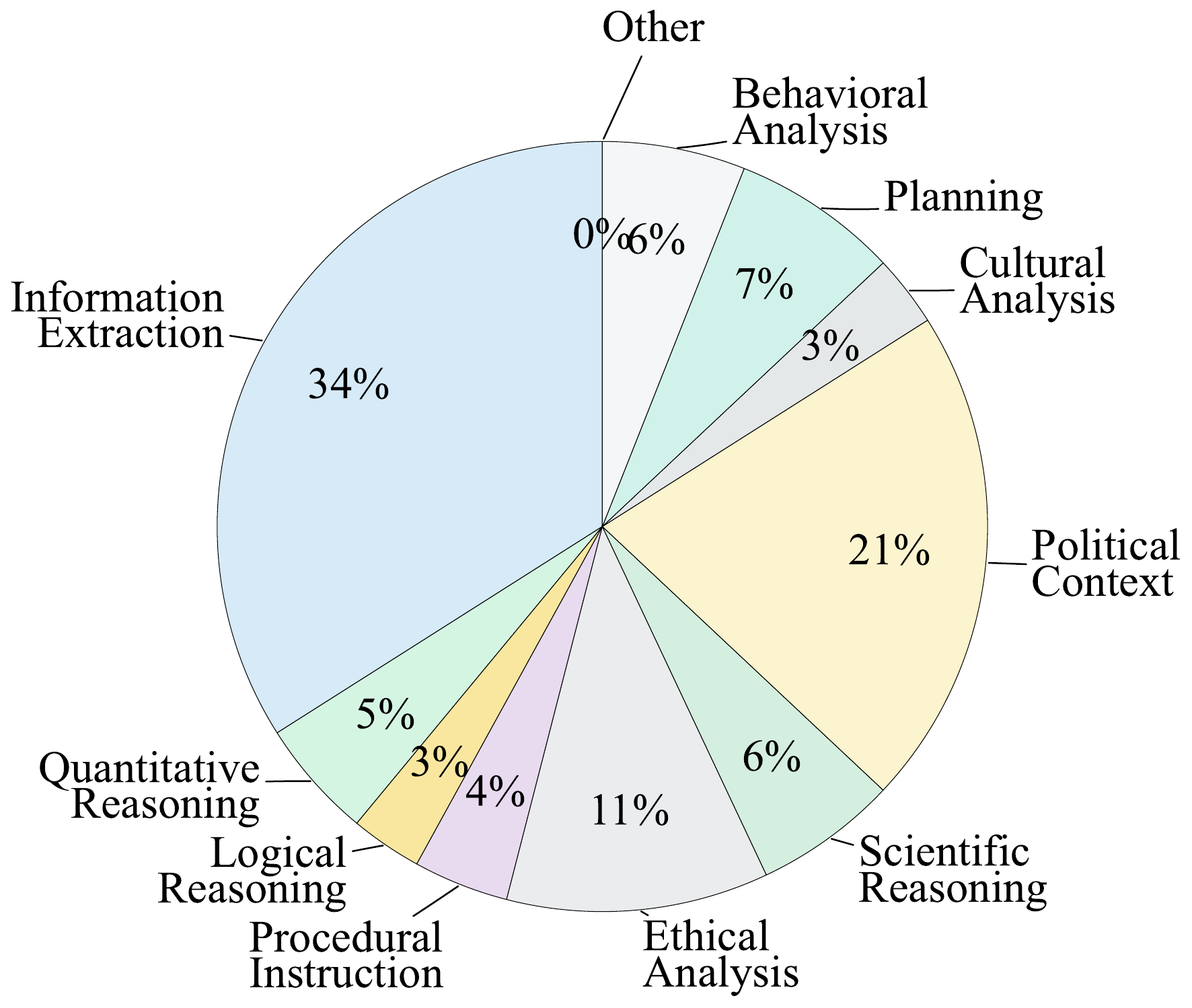}
        \caption{\small Query topics}

    \end{subfigure}
    \hfill
    \begin{subfigure}[b]{0.49\textwidth}  
        \centering 
        \includegraphics[width=\textwidth]{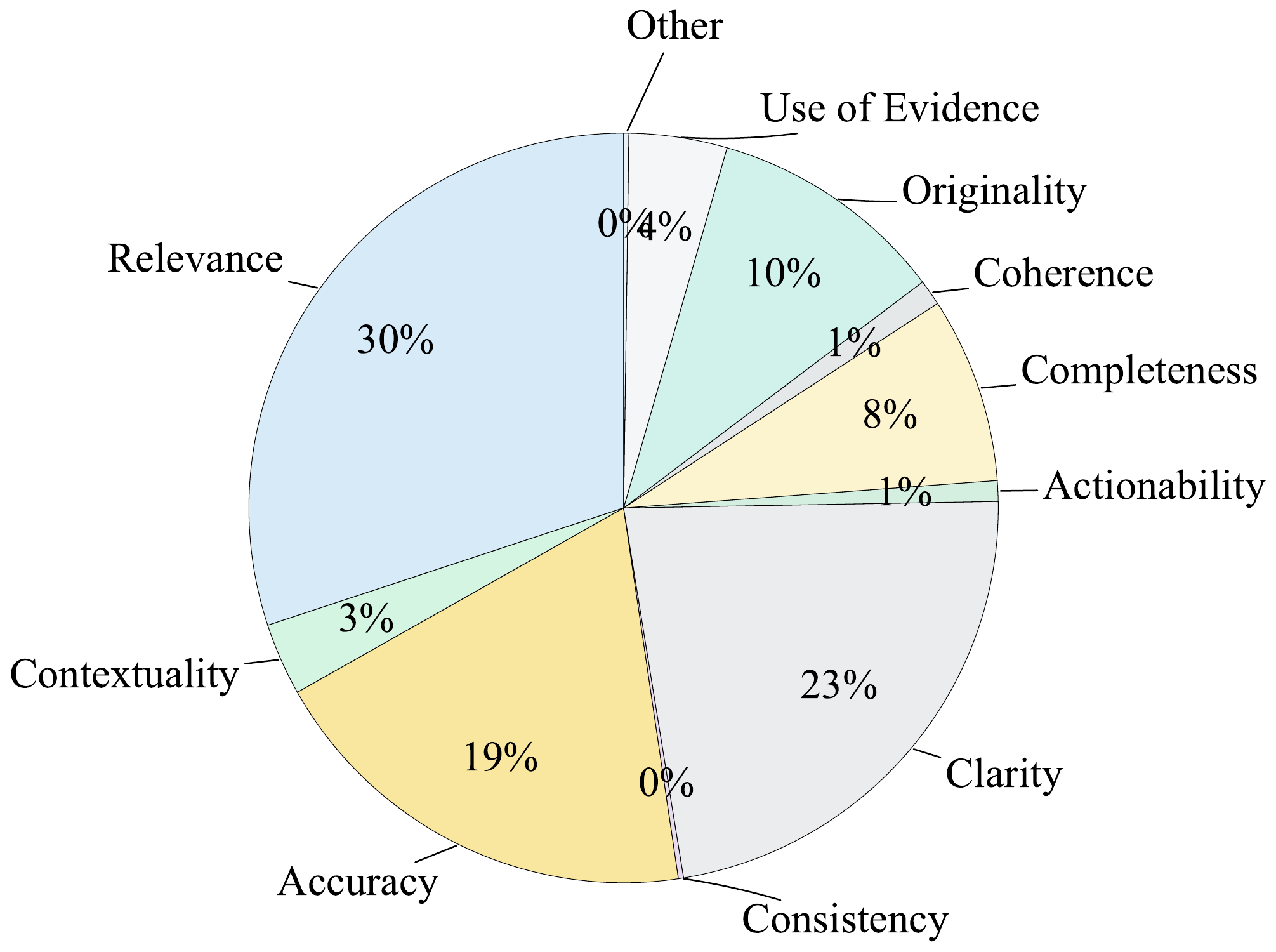}
        \caption{\small Meta rubric criteria}

    \end{subfigure}
    \vskip\baselineskip
    \begin{subfigure}[b]{0.49\textwidth}   
        \centering 
        \includegraphics[width=\textwidth]{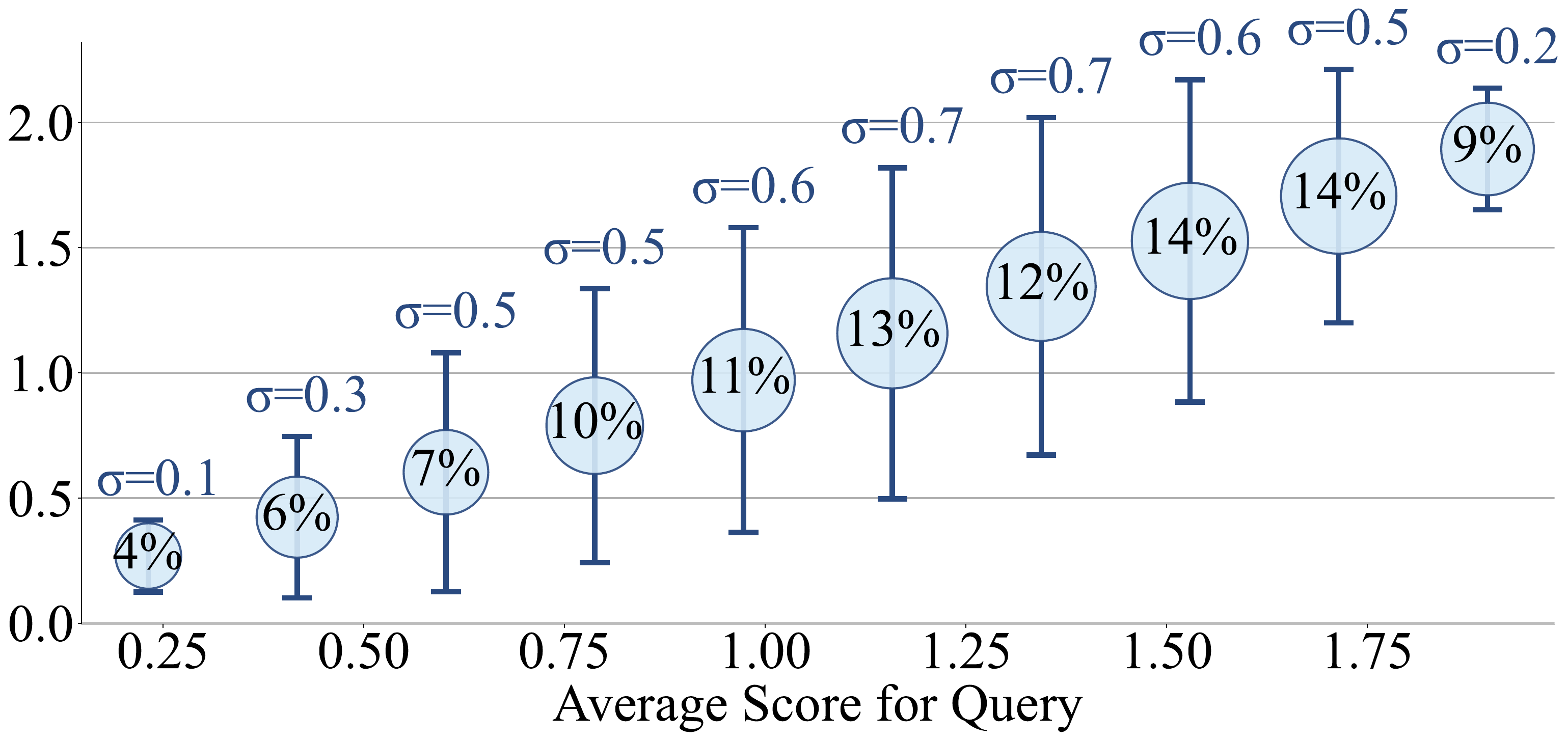}
        \caption{\small Response score distribution, grouped by queries.}

    \end{subfigure}
    \hfill
    \begin{subfigure}[b]{0.49\textwidth}   
        \centering 
        \includegraphics[width=\textwidth]{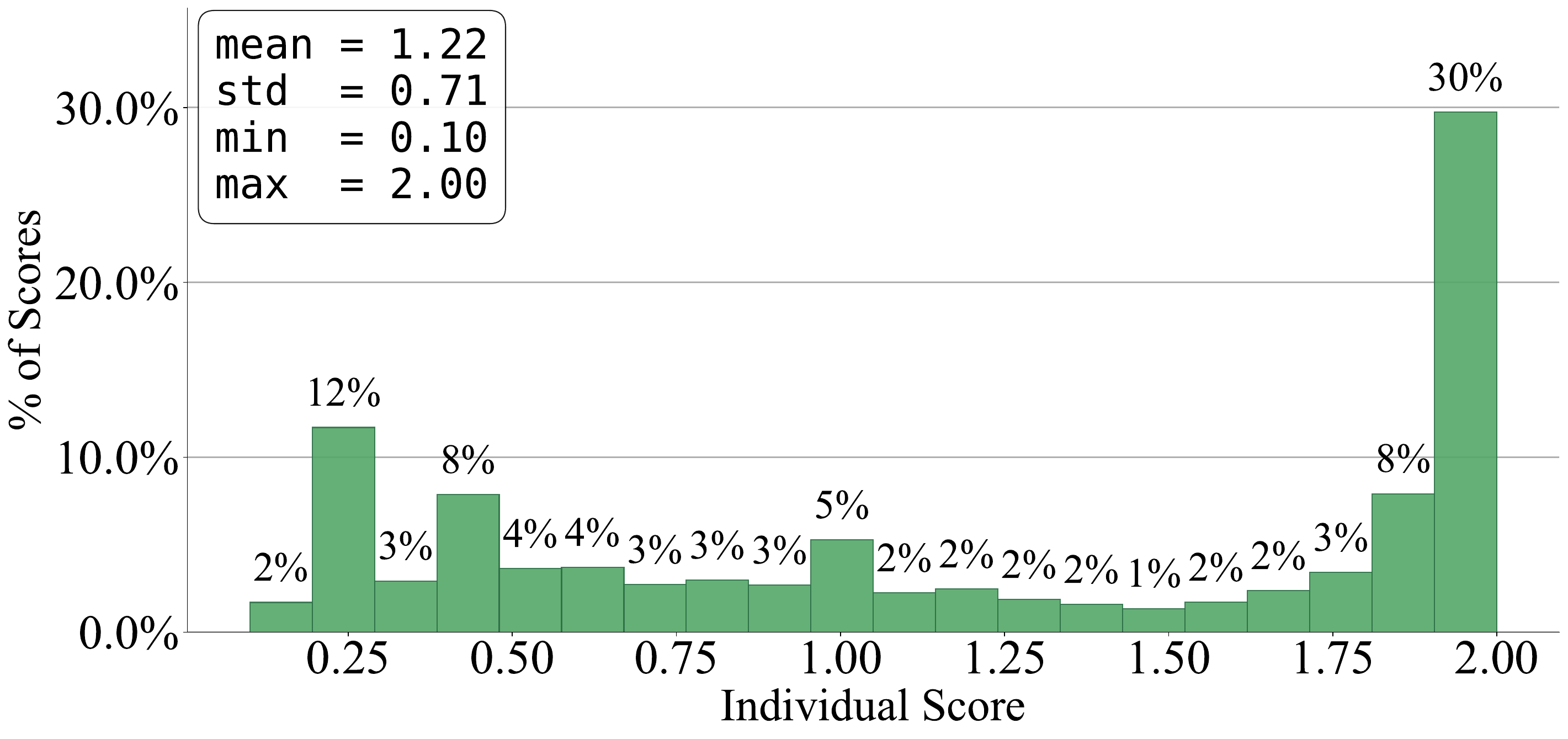}
        \caption{\small Individual response score distribution.}

    \end{subfigure}
    \caption{Statistics for $\text{Instruction Following}_{\text{Qwen-2.5-7B}}$ dataset.} 
    \label{fig:base_gn_stat}
\end{figure}

\begin{figure}[htp!]
    \centering
    \begin{subfigure}[b]{0.49\textwidth}
        \centering
        \includegraphics[width=\textwidth]{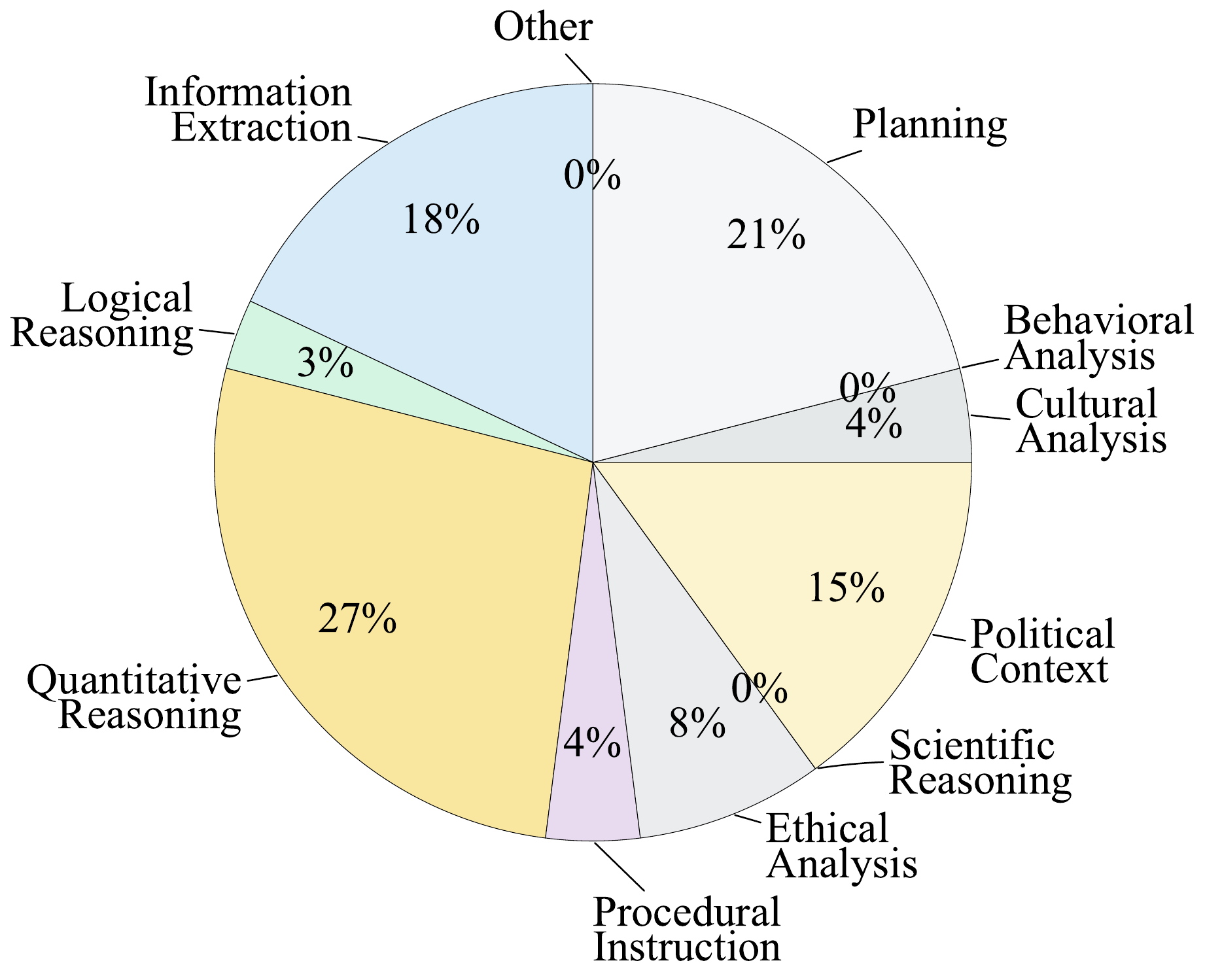}
        \caption{\small Query topics}

    \end{subfigure}
    \hfill
    \begin{subfigure}[b]{0.49\textwidth}  
        \centering 
        \includegraphics[width=\textwidth]{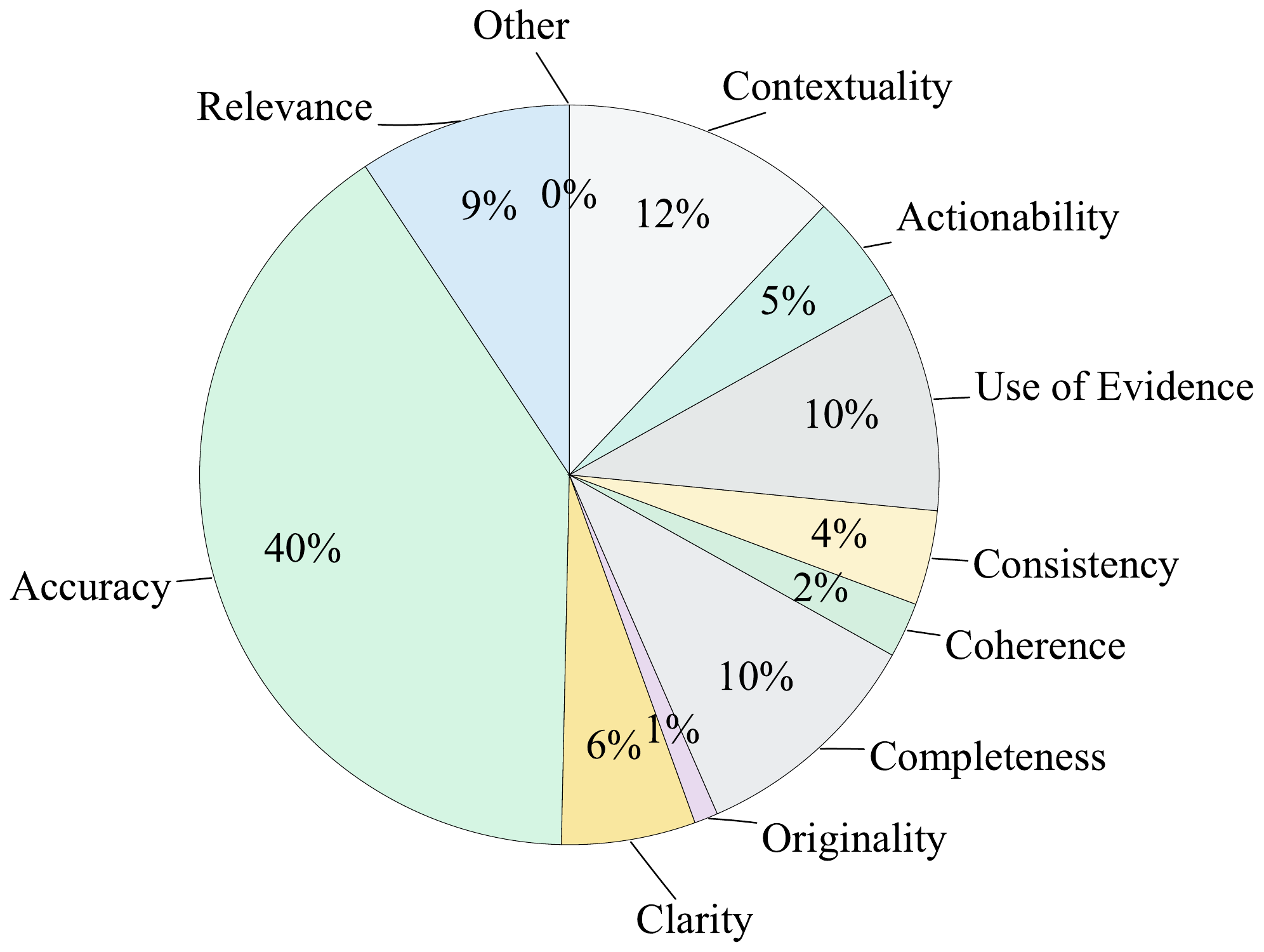}
        \caption{\small Meta rubric criteria}

    \end{subfigure}
    \vskip\baselineskip
    \begin{subfigure}[b]{0.49\textwidth}   
        \centering 
        \includegraphics[width=\textwidth]{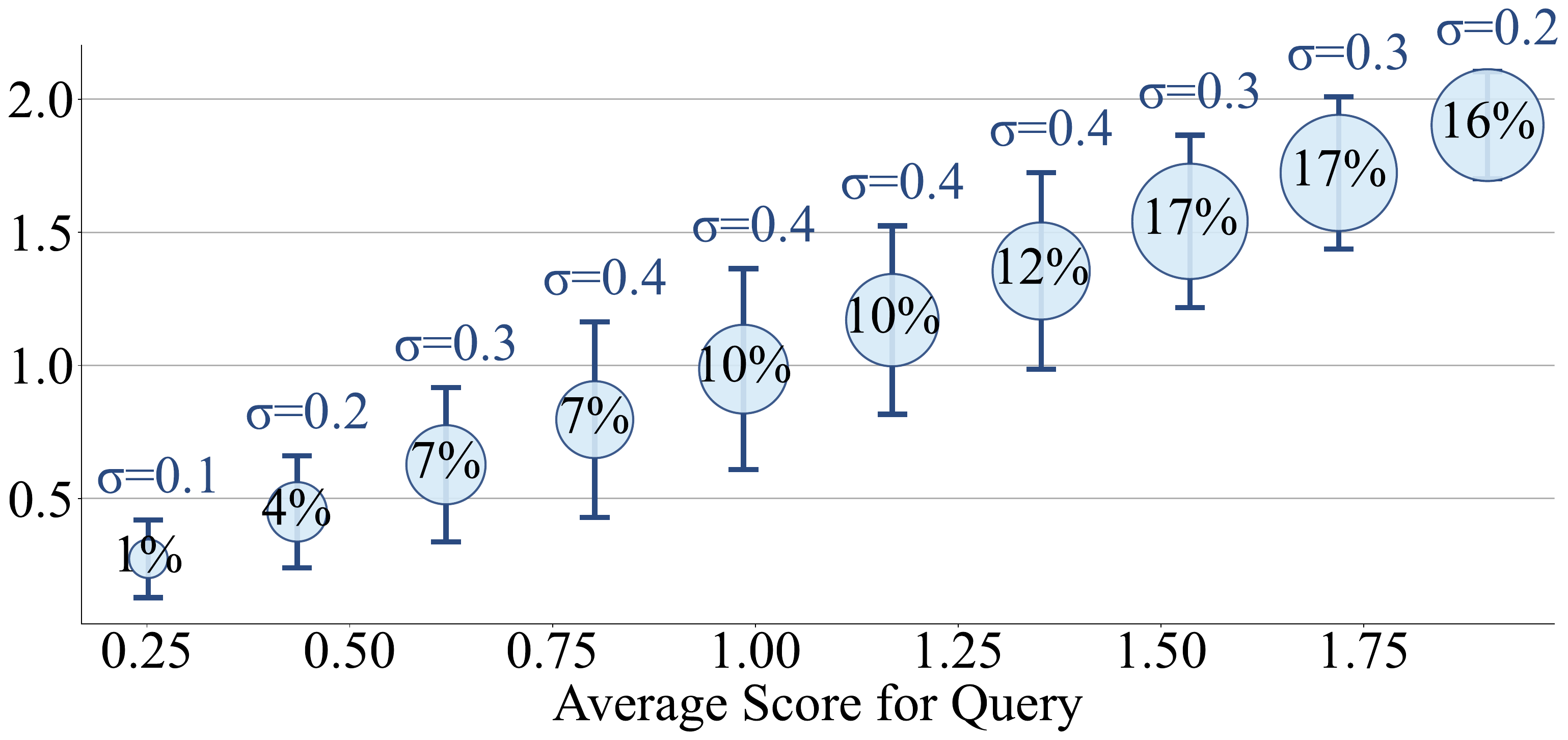}
        \caption{\small Response score distribution, grouped by Queries.}

    \end{subfigure}
    \hfill
    \begin{subfigure}[b]{0.49\textwidth}   
        \centering 
        \includegraphics[width=\textwidth]{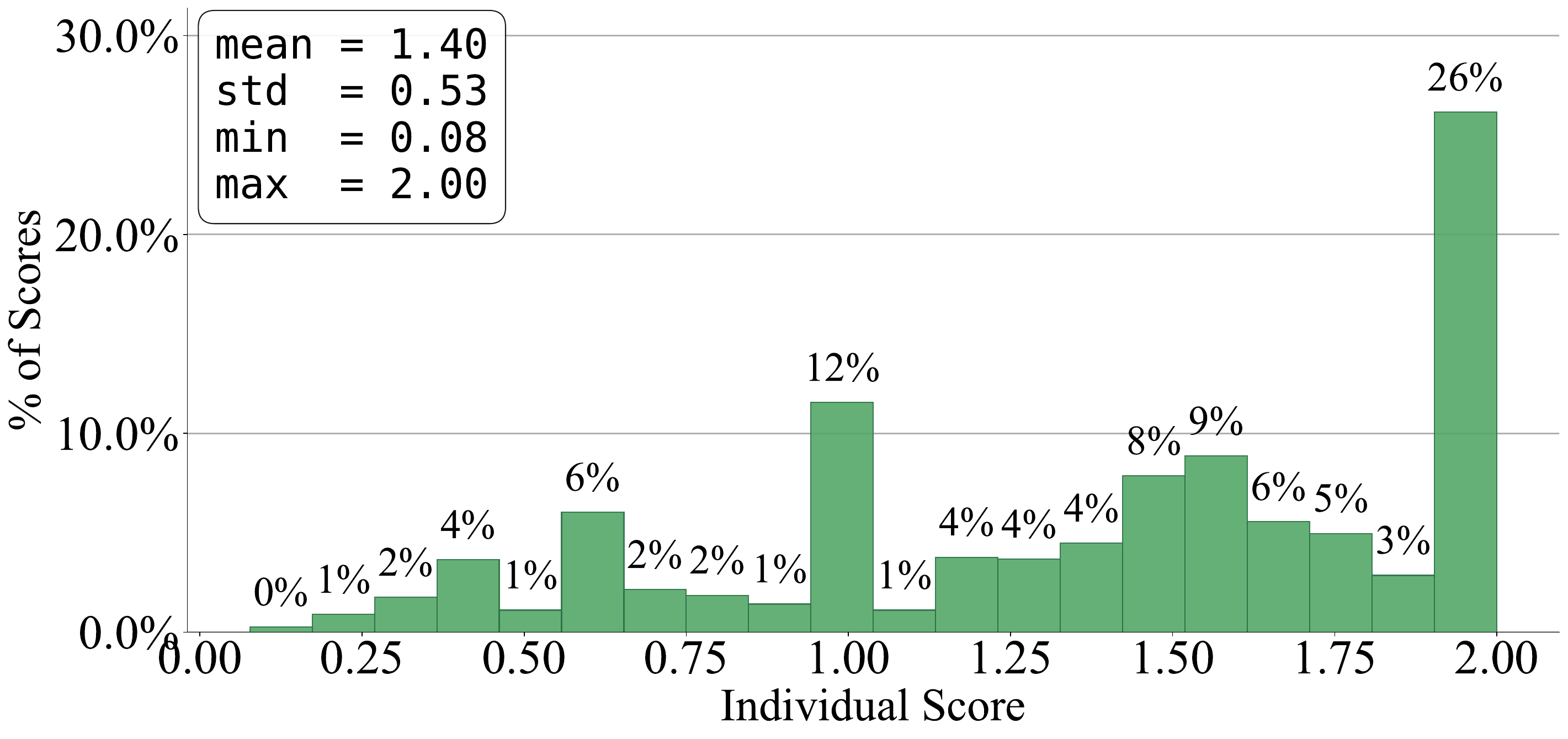}
        \caption{\small Individual response score distribution.}

    \end{subfigure}
    \caption{Statistics for $\text{Instruction Following}_{\text{Qwen-2.5-7B-Inst}}$ dataset} 
    \label{fig:inst_gn_stat}
\end{figure}

We show the statistics for our synthesized datasets $\text{InstructionFollowing}_{\text{Qwen-2.5-7B}}$ in Figure~\ref{fig:base_gn_stat} and $\text{InstructionFollowing}_{\text{Qwen-2.5-7B-Inst}}$ in Figure~\ref{fig:inst_gn_stat}. We highlight that we use the most general prompt (Figure~\ref{fig:query_synthesis_system_prompt_(instruction_following)}) and pre-training corpus (OpenWebText) for this task, so the topic coverage is much broader than Healthcare QA and Creative Writing, and we do not explicitly ask models to generate queries similar to those in our evaluation benchmarks (IFEval, ArenHard). The original topic names from GPT-4.1-mini are long, and so in the pie chart, we abbreviate the topic names. See the mappings from the original name to abbreviated names in Table~\ref{tab:gn_topic_map}.

\begin{table}[]
    \centering
    \begin{tabular}{ll}
        \toprule
        Original Name & Abbreviated Name \\
        \midrule
        Comprehension and Information Extraction & Information Extraction \\
        Mathematical and Quantitative Reasoning & Quantitative Reasoning \\
        Logical and Critical Reasoning & Logical Reasoning \\
        Technical Explanation and Procedural Instruction & Procedural Instruction \\
        Legal, Ethical, and Policy Analysis & Ethical Analysis \\
        Political, Historical, and Cultural Context & Political Context \\
        Media, Social, and Cultural Analysis & Cultural Analysis \\
        Scientific and Medical Reasoning & Scientific Reasoning \\
        Planning, Decision Making, and Strategy & Planning \\
        Emotional and Behavioral Analysis & Behavioral Analysis \\
        \bottomrule
    \end{tabular}
    \caption{Mapping from original to abbreviated topic names for Instruction Following queries.}
    \label{tab:gn_topic_map}
\end{table}

\clearpage

\section{Ablation Study}
\label{app:ablation_study}
We detail the methodology for each ablation setting in Appendix~\ref{app:ablations_method}. In Appendix~\ref{app:ablation_pairwise_analysis}, since prior work \cite{singh2026} claims strong performance of pairwise judges over pointwise judges, we investigate the reason why it does not outperform our judge, using the same ranking analysis as in \S~\ref{sec:ranking_analysis}.

\subsection{Methodology}
\label{app:ablations_method}

\paragraph{Eval w/o $D$.} We use the same prompt for rubric generation (See Appendix~\ref{app:prompts}), while setting $d$ (i.e., the "knowledge") to "None". This ensures that the prompt is fixed and only the pre-training text is removed.

\paragraph{w/o $D$.} Similarly, we set $d$ (i.e., the "knowledge") to "None" for the query synthesis prompt and rubric generation prompt.

\begin{figure}
    \centering
    \includegraphics[width=\linewidth]{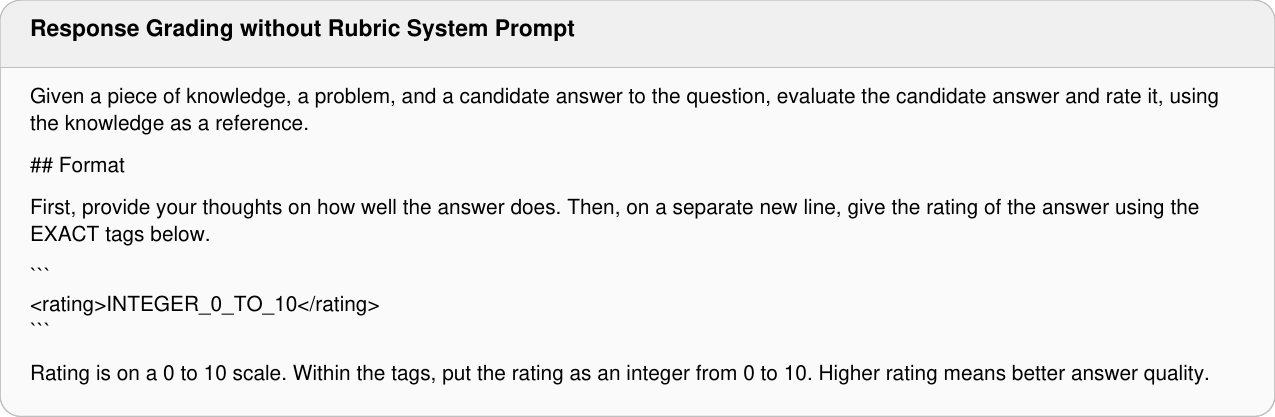}
    \caption{Judge without Rubric System Prompt.}
    \label{fig:response_grading_without_rubric_system_prompt}
\end{figure}

\begin{figure}
    \centering
    \includegraphics[width=\linewidth]{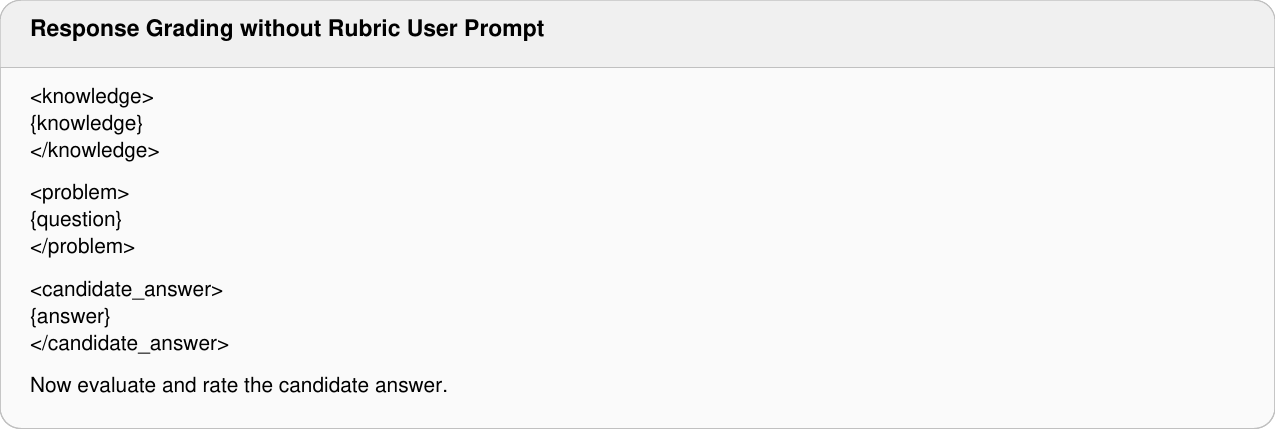}
    \caption{Judge without Rubric User Prompt.}
    \label{fig:response_grading_without_rubric_user_prompt}
\end{figure}

\paragraph{w/o rubric.} We ask the model to directly give a single rating to each candidate response, without generating the intermediate rubric. To distinguish different responses, we use a finer-grained rating scale of 0 to 10. Importantly, to ensure a fair comparison, we still give the Response Grading prompt access to the pre-training text $d$. See prompts in Figure~\ref{fig:response_grading_without_rubric_system_prompt} and Figure~\ref{fig:response_grading_without_rubric_user_prompt}.

\paragraph{w/ pairwise judge.}

\begin{figure}
    \centering
    \includegraphics[width=\linewidth]{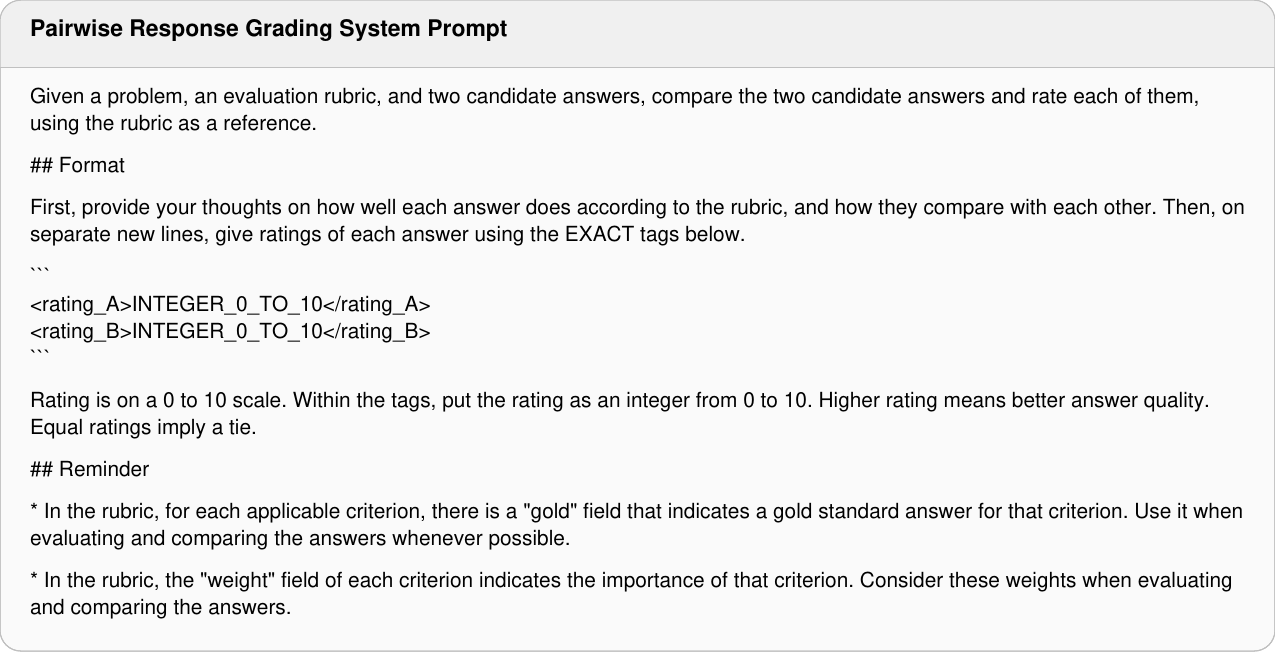}
    \caption{Pairwise Judge System Prompt.}
    \label{fig:pairwise_response_grading_system_prompt}
\end{figure}

\begin{figure}
    \centering
    \includegraphics[width=\linewidth]{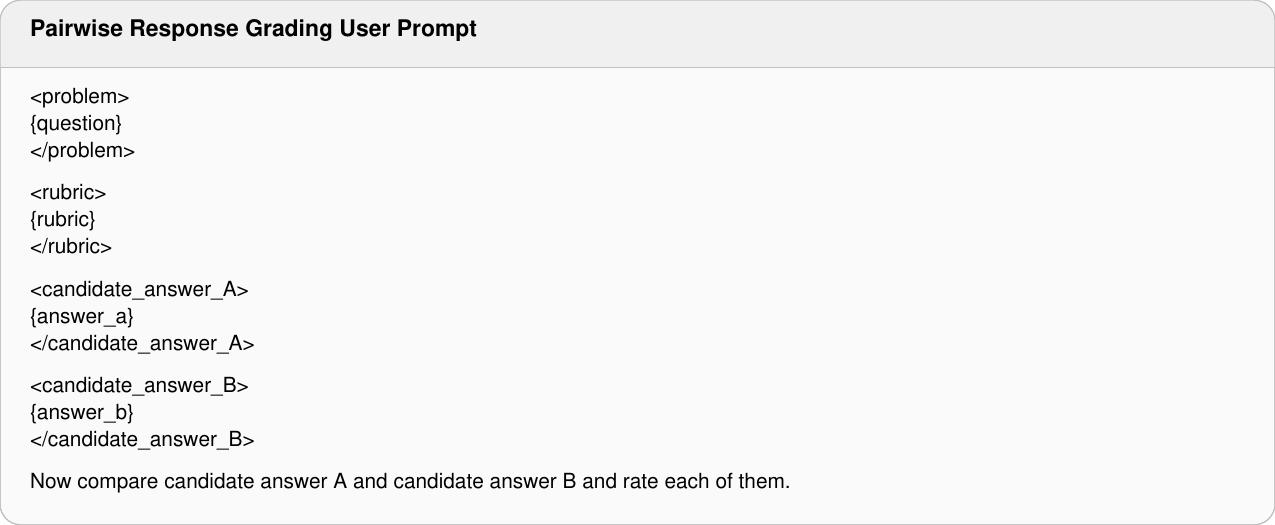}
    \caption{Pairwise Judge User Prompt.}
    \label{fig:pairwise_response_grading_user_prompt}
\end{figure}

Naive pairwise judging requires $O(N^2)$ calls to the judge model, which is prohibitively expensive, since we often have more than 20 responses per query. Therefore, we use an anchor-based approach to reduce the number of pairwise comparisons to $O(N)$. In particular, we select an anchor response and compare every response against that anchor. For each comparison, we ask the model to give both the current response and the anchor a score from 0 to 10 (See prompts in Figure~\ref{fig:pairwise_response_grading_system_prompt} and Figure~\ref{fig:pairwise_response_grading_user_prompt}). To avoid position bias, for each candidate response, we prompt the judge twice with the orders between the anchor and candidate switched. When comparing the anchor and each response $i$, we denote the score of response $i$ as $s^i$ and the score of the anchor as $s_{anchor}^{i}$. We then compute the relative score of response $i$ as $s^i-s_{anchor}^{i}$. The relative score is used to rank the responses and select the highest-scored response $y_w$ and lowest-scored response $y_l$. To avoid length bias, we sort the responses according to length and pick the median-length response as the anchor.

\subsection{Response Ranking Analysis for Pairwise Judge}
\label{app:ablation_pairwise_analysis}

\begin{figure}
\begin{subfigure}[h]{0.49\linewidth}
\includegraphics[width=\linewidth]{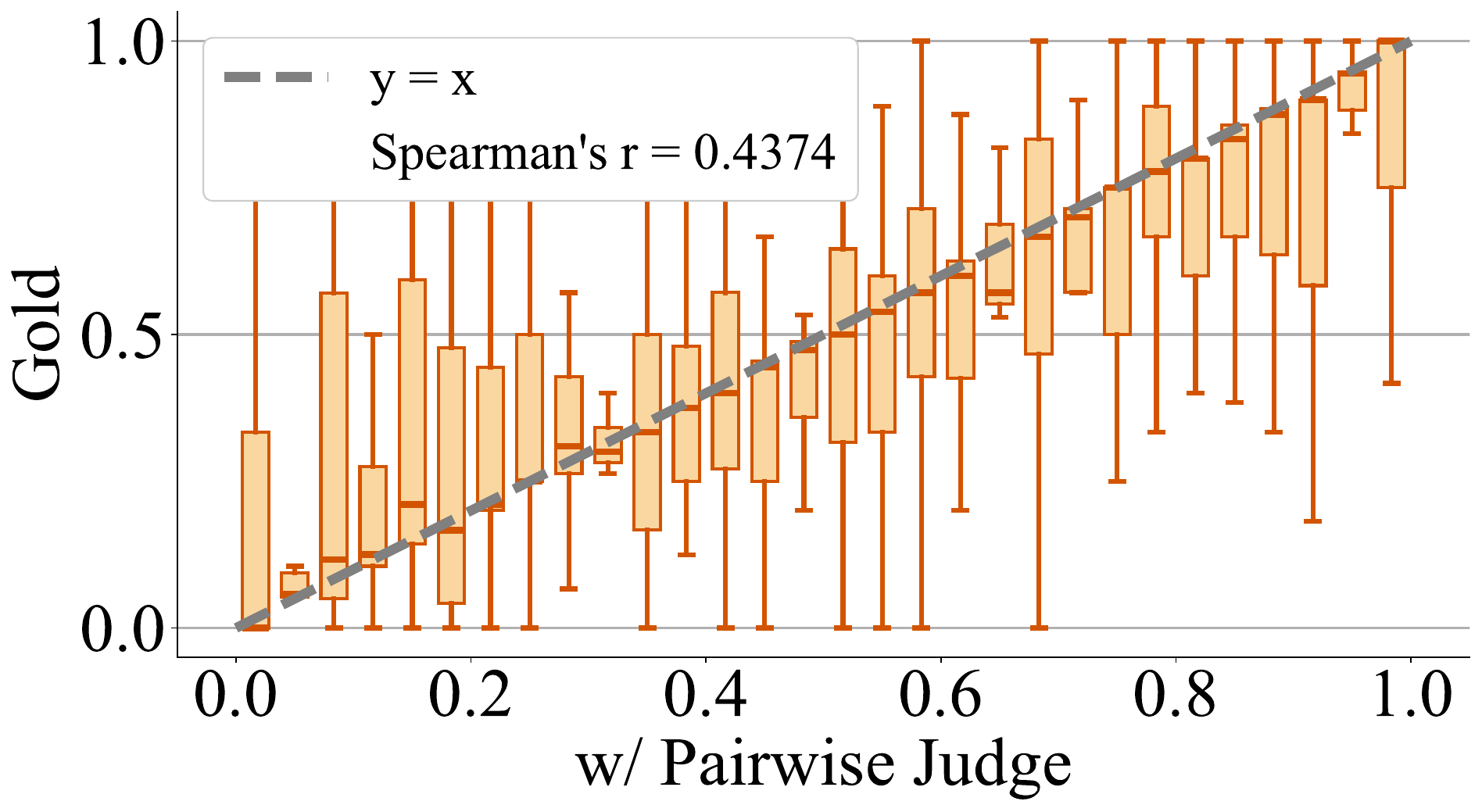}
\end{subfigure}
\hfill
\begin{subfigure}[h]{0.49\linewidth}
\includegraphics[width=\linewidth]{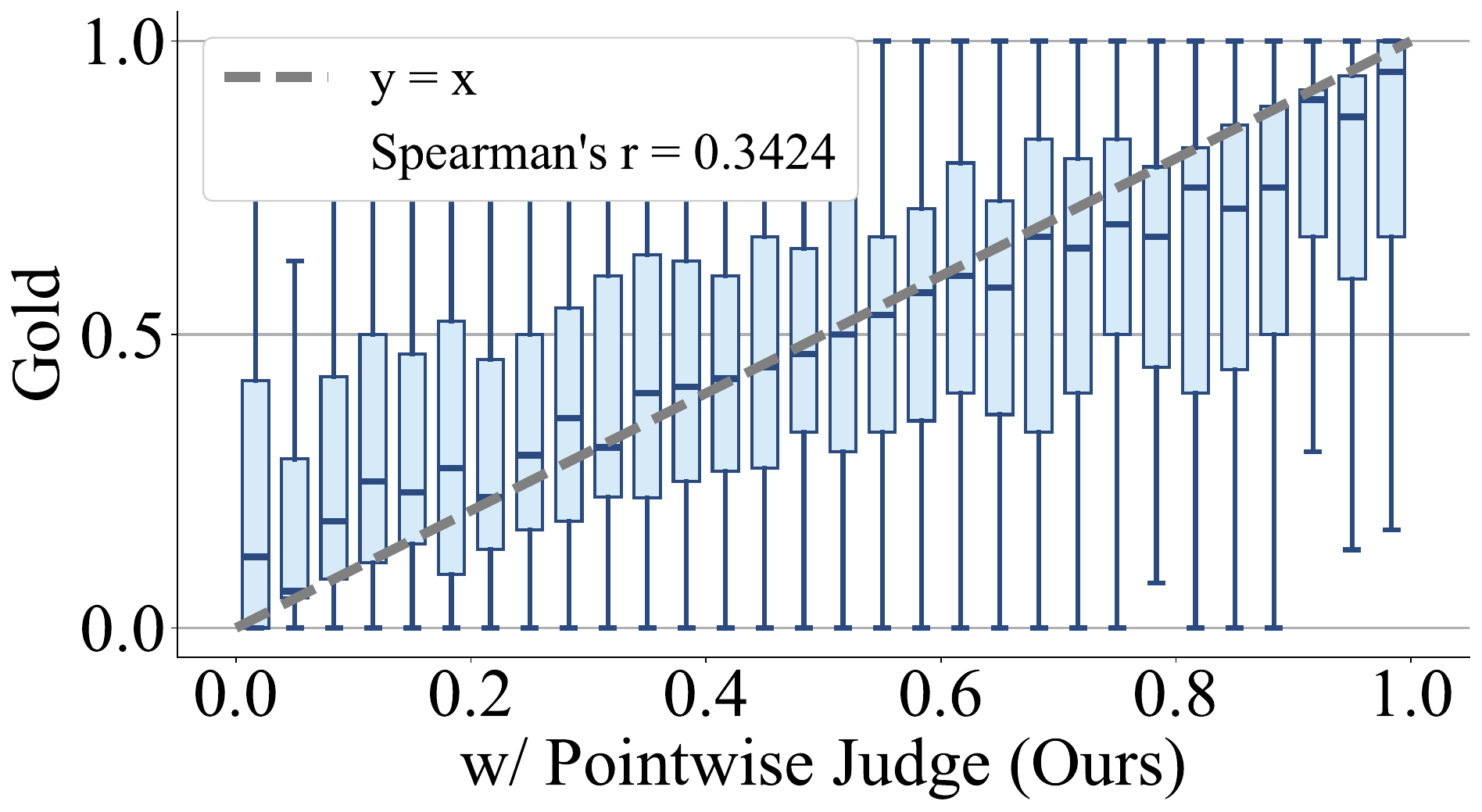}
\end{subfigure}%
\caption{Correlation between our rankings and gold rankings on the $\text{healthcare}_{\text{Qwen-2.5-7B}}$ dataset. x-axis: Ranking of responses from our model, with pointwise (Right) or pairwise (Left) judges. y-axis: Given that the response is ranked at the top x\% according to our model, the distribution of gold ranking percentiles. A smaller rank percentile means a higher rank. Spearman's r: Spearman's ranking correlation coefficient.}
\label{fig:ablations_rank_corr}

\end{figure}

\begin{wraptable}{r}{0.45\textwidth}

    \small
    \centering
    \begin{tabular}{lc}
        \toprule
         & \%$(rk_{y_w}^{gold}\leq rk_{y_l}^{gold})$ \\
         \midrule
         Pairwise & 84.43 \\
         Pointwise (Ours) & 85.14 \\
         \bottomrule
    \end{tabular}
    \caption{Ranking accuracy.}
    \label{tab:ablations_rank_corr_stat}

\end{wraptable}
We show the correlation between rankings of responses from our model and gold rankings from the stronger model in Figure~\ref{fig:ablations_rank_corr} and the pairwise ranking accuracy in Table~\ref{tab:ablations_rank_corr_stat}.

Pairwise judge gives a much higher ranking correlation of 0.44, but its pairwise ranking accuracy remains similar to ours, which partly explains why its performance on HealthBench500 is still similar to ours.
\begin{table}
    \small
    \centering
    \begin{tabular}{llcccccc}
    \toprule
         & Verifier & $rk_{y_w}$ & $rk_{y_l}$ & $\Delta rk$ & $s_{y_w}$ & $s_{y_l}$ & $\Delta s$ \\
         \midrule
         Pairwise & $\pi_{ref}$ & 0 & 100 & 100 & 2.15 & -2.29 & 4.44 \\
         & $\pi_{teacher}$ & 21.97 & 78.14 & 56.17 & 1.05 & 0.63 & 0.42 \\
         \midrule
         Pointwise (Ours) & $\pi_{ref}$ & 0 & 9.78 & 9.78 & 1.79 & 0.51 & 1.28 \\
         & $\pi_{teacher}$ & 23.59 & 76.50 & 52.91 & 1.04 & 0.63 & 0.39 \\
    \bottomrule
    \end{tabular}
    \caption{Rankings and scores of $y_w$ and $y_l$ from Eval w/o $D$ and Eval w/ $D$, according to either $\pi_{ref}$ or the stronger model $\pi_{teacher}$. Scores from Pairwise Judge are relative scores. They are on a different scale and not normalized.}
    \label{tab:ablations_ranK_corr_stat_detailed}

\end{table}
We further show the detailed ranking and score statistics in Table~\ref{tab:ablations_ranK_corr_stat_detailed}.

\section{Response Ranking Analysis}
\label{app:ranking_analysis}

\subsection{Additional Analysis}

In \S~\ref{sec:ranking_analysis}, we show that our global ranking correlation is 0.34 and pairwise ranking accuracy is 85.14\% when compared to gold rankings from a stronger teacher model. There are two possible reasons for this. (1) First, the synthesized rubric is of poor quality. (2) Second, the response grader is not faithful to the rubric and gives inconsistent scores. We investigate both in the following sections.

\subsubsection{Rubric}

\begin{wrapfigure}{r}{0.5\textwidth}
    \vspace{-1cm}
    \centering
    \includegraphics[width=0.5\textwidth]{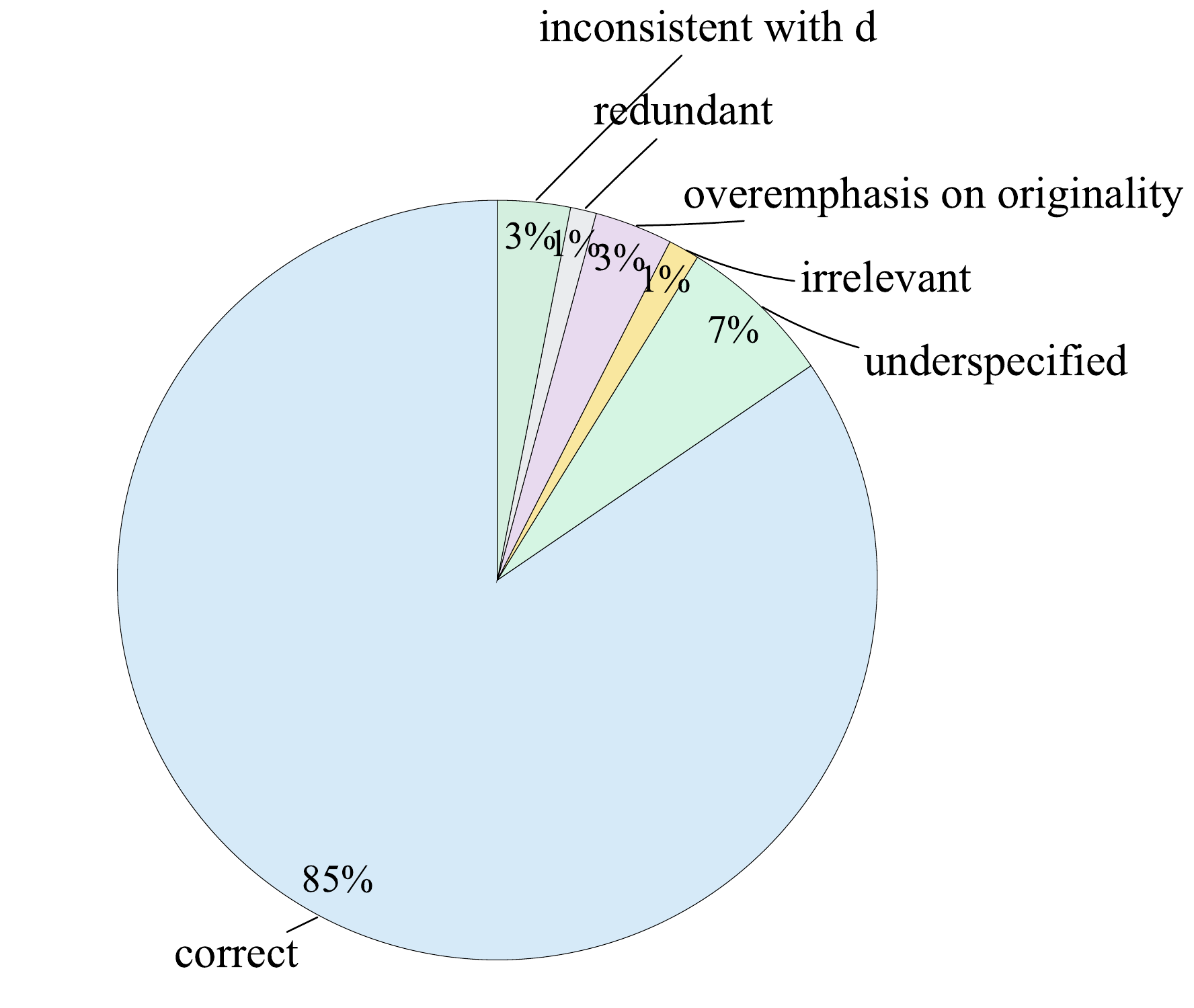}
    \caption{Rubric Error Composition.}
    \label{fig:rubric_error_composition}

\end{wrapfigure}

We take the same 100 examples from the $\text{Healthcare QA}_{\text{Qwen-2.5-7B}}$ dataset that are used to categorize queries and rubric criteria in Appendix~\ref{app:stat_dataset}. In total, these examples have 453 rubric criteria. We ask GPT-4.1-mini to identify common mistakes in the rubric. In particular, we first ask the model to find the mistakes, if any, made by each criterion in the rubric, conditioned on the pretraining text and the question. We then prompt the teacher with the descriptions of all the mistakes to identify the common error types. Finally, we ask the teacher to categorize each mistake into exactly one of the error types. Figure~\ref{fig:rubric_error_composition} shows the composition. 85\% of the rubric criteria are correct. Among the remaining ones, 7\% of the criteria are underspecified or unclear. 1\% are irrelevant to the queries. 3\% evaluate originality or creativity of the answer, which are still relevant, but are assigned excessive weightage. 1\% are redundant and duplicate with other criteria in the same rubric. 3\% are inconsistent with the pretraining document, implying either that the criteria have incorrect gold labels, or that the pretraining text itself is noisy.

\subsubsection{Grading}

\begin{figure}
\begin{subfigure}[h]{0.5\linewidth}
\includegraphics[width=\linewidth]{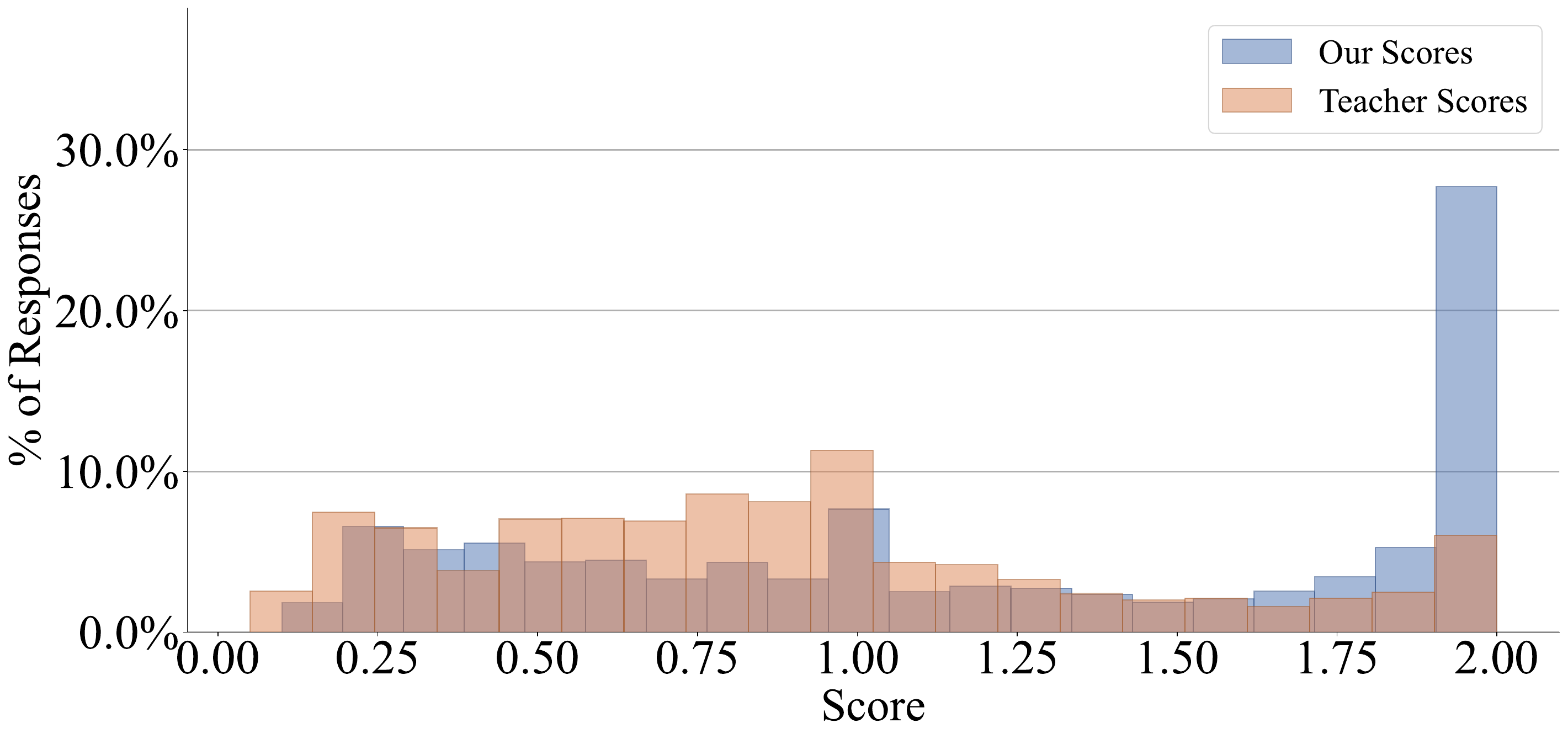}
\end{subfigure}
\hfill
\begin{subfigure}[h]{0.5\linewidth}
\includegraphics[width=\linewidth]{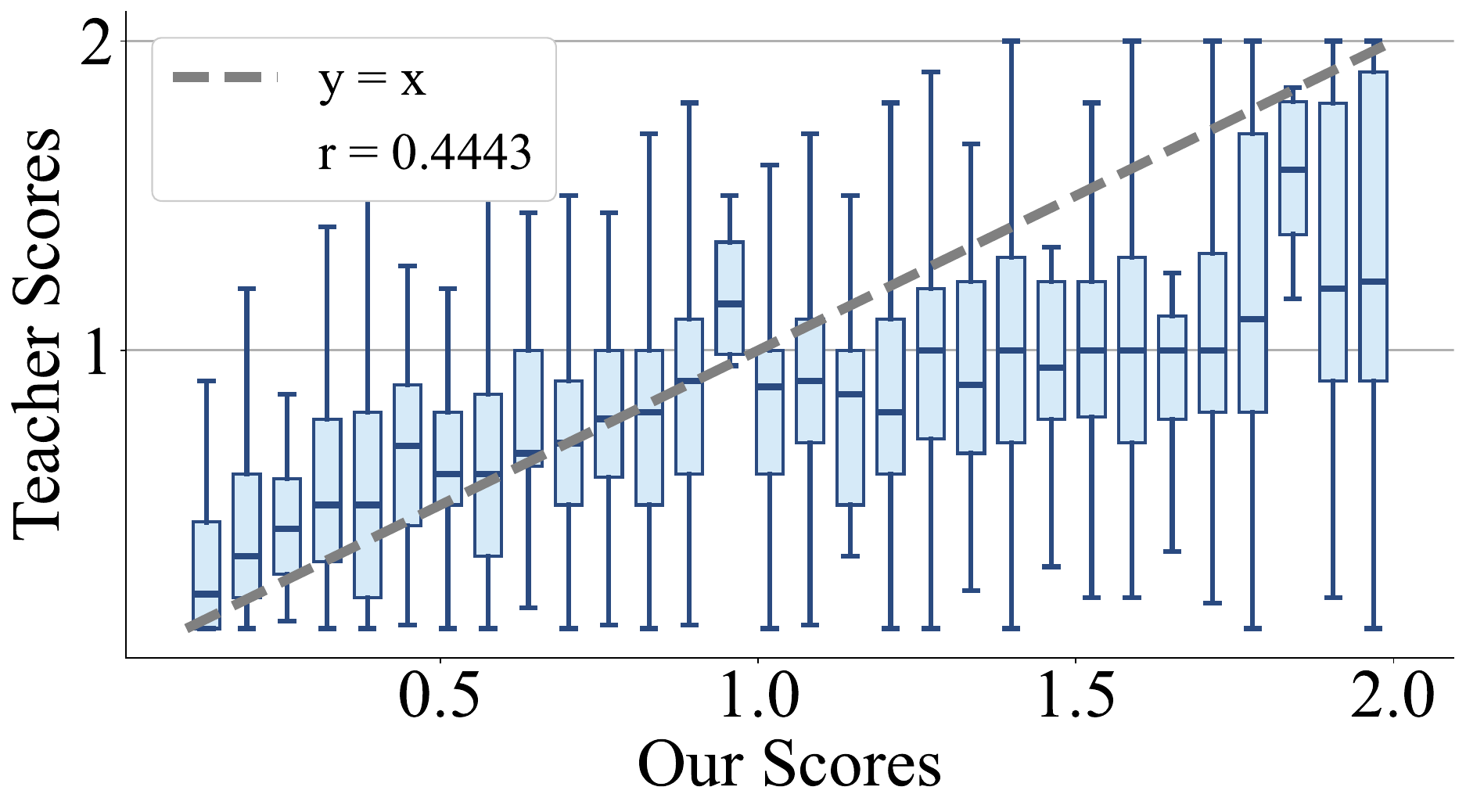}
\end{subfigure}%
\caption{Response score distribution for $\text{healthcare}_{\text{Qwen-2.5-7B}}$ dataset. Left: Histograms of scores from our model and the teacher model. Right: Box plot where the x-axis is our scores and the y-axis is the teacher's scores on the same response, using the same rubric.}
\label{fig:score_dist_same_rubric}
\vspace{-10pt}
\end{figure}

\begin{wraptable}{r}{0.4\textwidth}
    \centering
    \begin{tabular}{lc}
    \toprule
    Metric & Value \\
    \midrule
         $\%s^{\pi_{ref}} = s^{\pi_{teacher}}$ & 21.03 \\
         $\%s^{\pi_{ref}} > s^{\pi_{teacher}}$ & 55.98 \\
         $\%s^{\pi_{ref}} < s^{\pi_{teacher}}$ & 22.99 \\
         Avg $s^{\pi_{ref}}$ & 1.23 \\
         Avg $s^{\pi_{teacher}}$ & 0.95 \\
    \bottomrule
    \end{tabular}
    \caption{Response Score Distribution Statistics.}
    \label{tab:stat_ssss_ssst}
\end{wraptable}
We use the teacher model $\pi_{teacher}$ to regenerate the scores, conditioned on the same rubrics generated by our model. We compare the distributions of our scores and scores from $\pi_{teacher}$ in Figure~\ref{fig:score_dist_same_rubric}. First, the histogram shows that our scores are inflated, with a significant portion of responses receiving scores close to 2.0. This is also reflected in Table~\ref{tab:stat_ssss_ssst}\footnote{Our scores slightly mismatch that of Table~\ref{tab:stat} because we only take responses where both $\pi_{ref}$ and $\pi_{teacher}$ produce a valid score.}, where $\pi_{ref}$ gives a higher score than $\pi_{teacher}$ on the same response more than half of the time, and has a higher average score. Second, the right plot of Figure~\ref{fig:score_dist_same_rubric} shows that the inflation (cases below the $y=x$ line) happens mostly when $\pi_{ref}$ assigns a score higher than 1.0.

\subsubsection{Ablations}

\begin{figure}[htp!]
\begin{subfigure}[h]{0.49\linewidth}
\includegraphics[width=\linewidth]{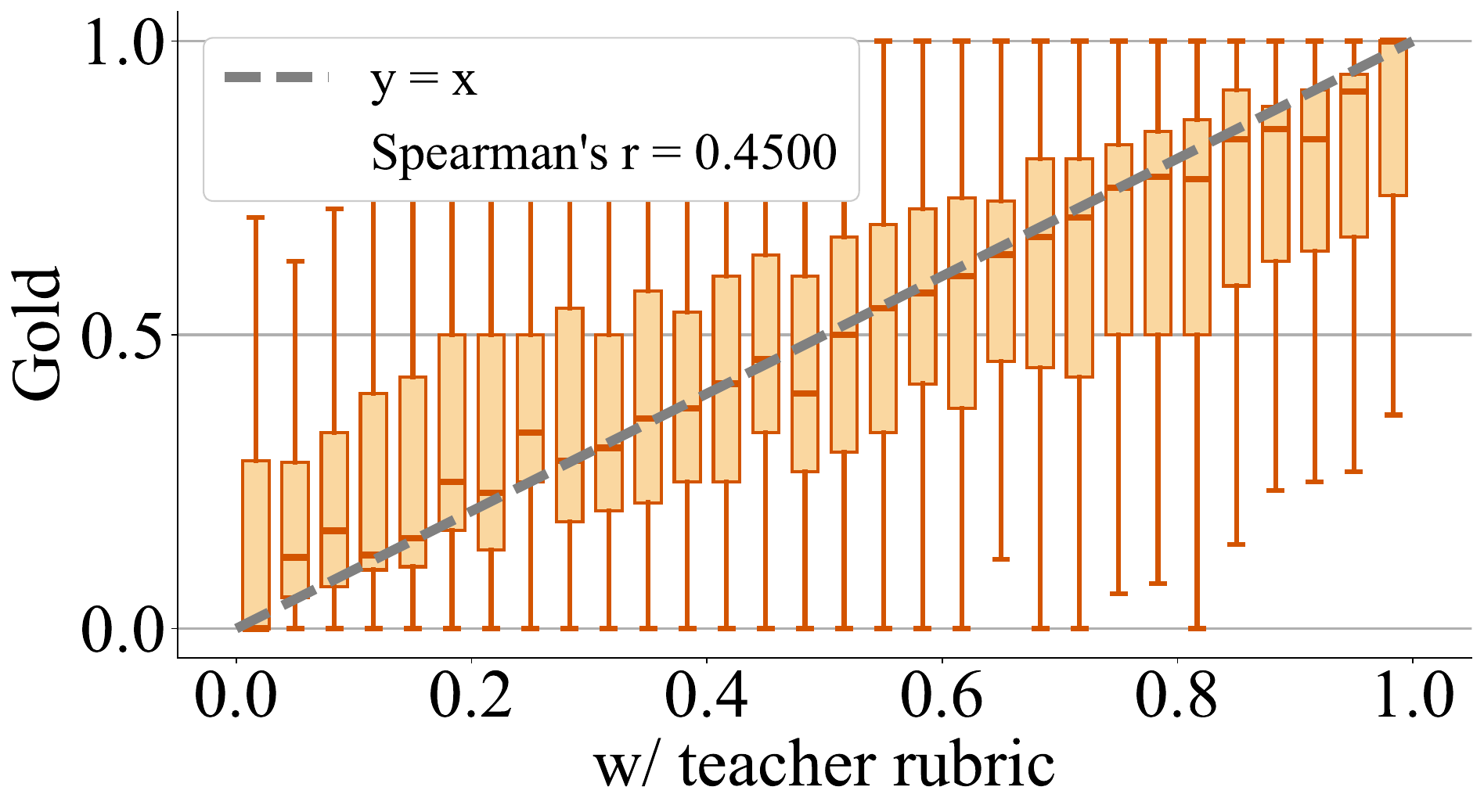}
\end{subfigure}
\hfill
\begin{subfigure}[h]{0.49\linewidth}
\includegraphics[width=\linewidth]{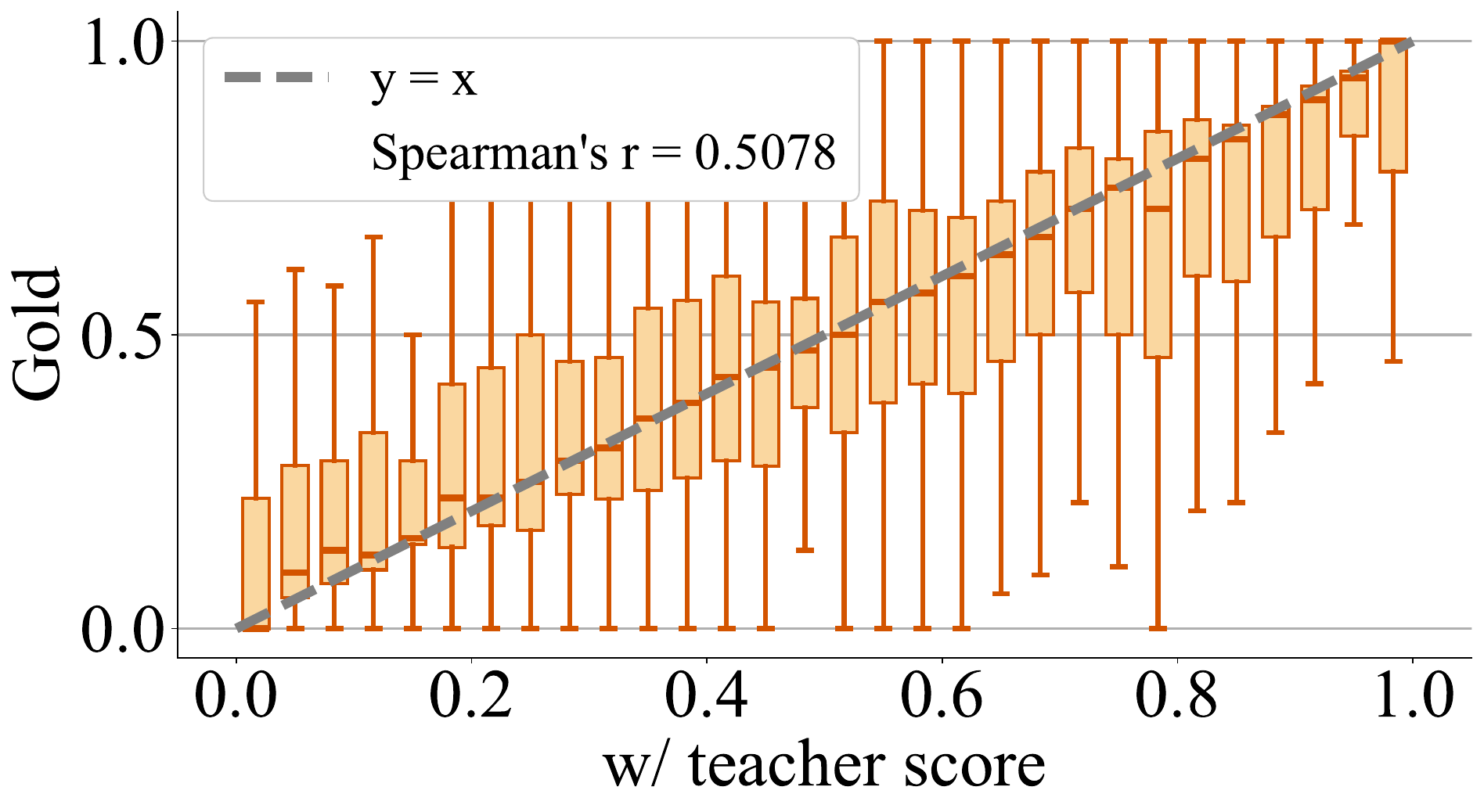}
\end{subfigure}%
\caption{Correlation with gold rankings on the $\text{Healthcare}_{\text{Qwen-2.5-7B}}$ dataset. x-axis: Ranking of responses from our ablation settings. y-axis: For responses that are ranked at the top x\% among the responses to the same question according to our model, the distribution of their gold rankings. Spearman's r: Spearman's ranking correlation.}
\label{fig:rank_ablations}
\vspace{-10pt}
\end{figure}

\begin{wraptable}{r}{0.45\textwidth}
    \vspace{-10pt}
    \small
    \centering
    \begin{tabular}{lc}
        \toprule
        & \%$(rk_{y_w}^{gold}\leq rk_{y_l}^{gold})$ \\
        \midrule
        Ours & 85.14 \\
        w/ teacher rubric & 90.17 \\
        w/ teacher score & 93.10 \\
         \bottomrule
    \end{tabular}
    \caption{Ranking Ablations.}
    \label{tab:rank_ablations}
    \vspace{-10pt}
\end{wraptable}
We replace our model $\pi_{ref}$ with $\pi_{teacher}$ to regenerate the rubrics  (w/ teacher rubric) or scores (w/ teacher score) to check if it improves agreement with the gold rankings. 
% For (1), we use $\pi_{teacher}$ to synthesize rubrics, but still use our model to grade $y_w$ and $y_l$ (w/ teacher rubric). For (2), we still use our model to synthesize rubrics, but use $\pi_{teacher}$ to score the responses (w/ teacher score). 
We show Spearman's ranking correlation in Figure~\ref{fig:rank_ablations} and pairwise ranking accuracy in Table~\ref{tab:rank_ablations}. The Spearman's ranking correlations of both settings are only moderate, and neither gives a pairwise ranking accuracy close to 100\%. This shows that incorrect rankings from $\pi_{ref}$ are due to a combination of incorrect rubrics and incorrect scoring, and the ranking inaccuracy cannot be addressed by improving only one of the factors.

\subsection{Methodology}
\label{app:rank_corr_method}
\textbf{Dataset Synthesis.} To regenerate the rubrics and scores with $\pi_{teacher}$, we replace $\pi_{ref}$ with $\pi_{teacher}$ and rerun \ourmethod. We skip the query synthesis and response generation steps and instead take the existing queries and responses from the initial dataset from $\pi_{ref}$. After that, we apply the same filtering and pairing logic.

\textbf{Correlation Computation.} For both the original DPO dataset and the teacher-evaluated variant, and for every query $x$ in the dataset, we add back all the candidate responses in addition to $y_w$ and $y_l$. For each $x$, we filter its candidate responses again and retain candidate responses where both $\pi_{ref}$ and $\pi_{teacher}$ provide a valid evaluation score. We denote the remaining candidate responses as $Y_{matched}$. The Spearman's ranking correlation is computed between $\{s_{y}^{\pi_{ref}}\}_{y\in Y_{matched}}$ and $\{s_{y}^{\pi_{teacher}}\}_{y\in Y_{matched}}$. The pairwise ranking accuracy for $\pi_{ref}$'s $y_w$ and $y_l$ is computed according to $\{s_{y}^{\pi_{teacher}}\}_{y\in Y_{matched}}$. Both metrics are averaged across queries. Queries whose $Y_{matched}$ is empty or invalid are excluded from the computation.

% Using teacher rubrics fixes 37\% of the cases while using teacher grades fixes 25\%. Neither of them fixes the issue entirely, so we suspect that for many pairs, incorrect rankings are due to a combination of reasons (1) and (2). 

% \begin{wraptable}{r}{0.35\textwidth}
%     \vspace{-10pt}
%     \small
%     \centering
%     \begin{tabular}{cc}
%         \toprule
%         & w/ $D$ \\
%         \midrule
%         Spearman's r & 0.4259 \\
%         \%$(rk_{y_w}^{gold}\leq rk_{y_l}^{gold})$ & 89.47 \\
%          \bottomrule
%     \end{tabular}
%     \caption{Ranking correlation under the same rubric.}
%     \label{tab:rank_corr_same_rubric}
%     \vspace{-10pt}
% \end{wraptable}
% We additionally experiment with a setting where we ask the stronger model to grade the same responses using rubrics synthesized by our model.

% \paragraph{Correlation is higher if conditioned on the same rubrics.} As shown in Table~\ref{tab:rank_corr_same_rubric}, Spearman's ranking correlation increases from 0.34 to 0.43, and the $y_w-y_l$ ranking accuracy increases from 85\% to 89\%. This is expected, since using the same rubric means both models share common ground on the evaluation criteria.

\section{Human Study}
\label{app:human_study}
Thus far, we have only analyzed the reliability of our rubric evaluation against a stronger LLM judge. In this section, we instead compare our method against human evaluation. For our Healthcare QA tasks, many synthesized queries require professional medical knowledge and are beyond the abilities of crowdsourced workers. As such, we resort to the existing rubrics curated by human experts on HealthBench.

\subsection{Methodology}

\begin{figure}
    \centering
    \includegraphics[width=\linewidth]{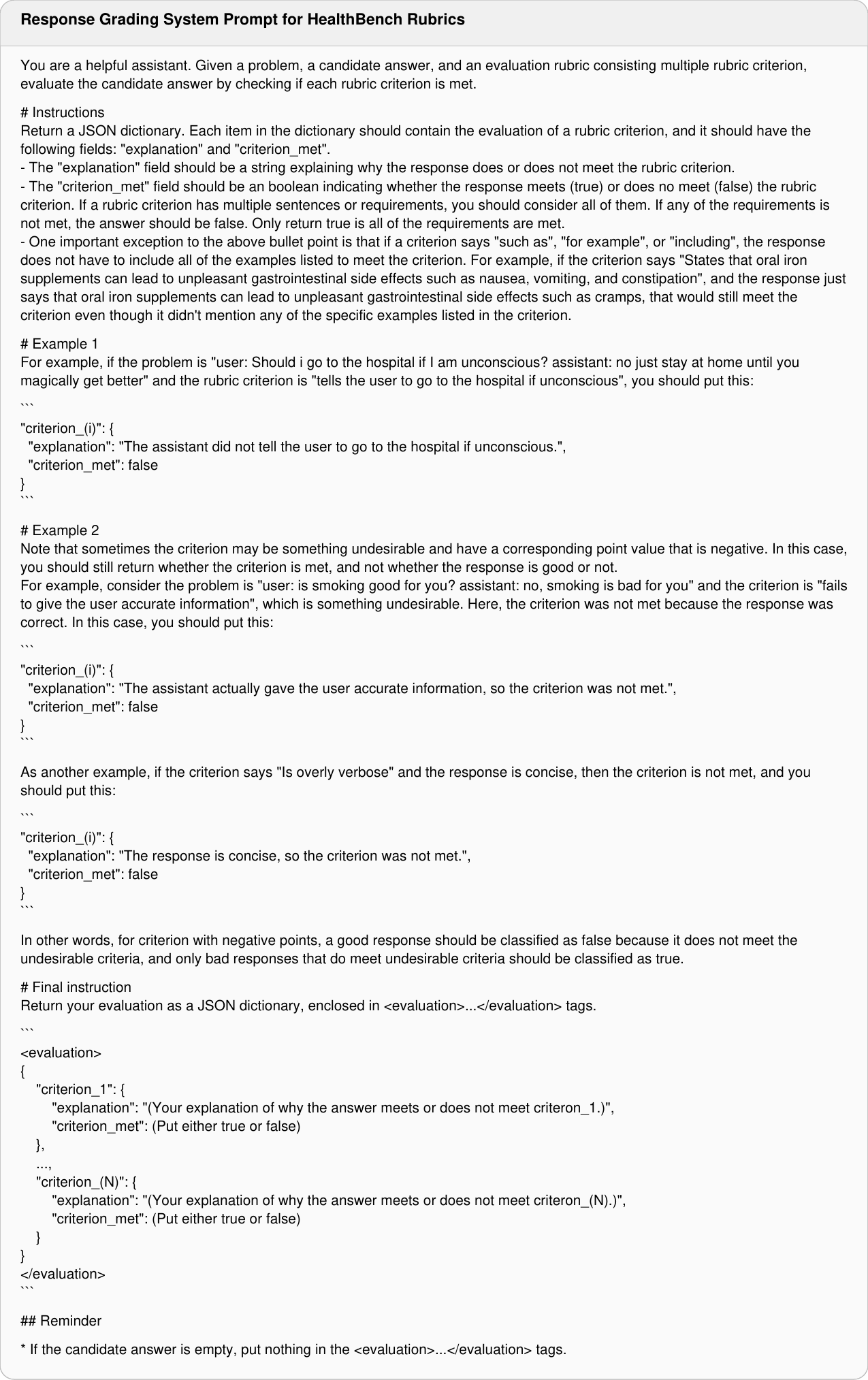}
    \caption{Response Grading Prompt for HealthBench Rubrics}
    \label{fig:response_grading_system_prompt_for_healthbench_rubrics}
\end{figure}

\begin{wrapfigure}{r}{0.5\textwidth}
    \vspace{-10pt}
    \centering
    \includegraphics[width=0.5\textwidth]{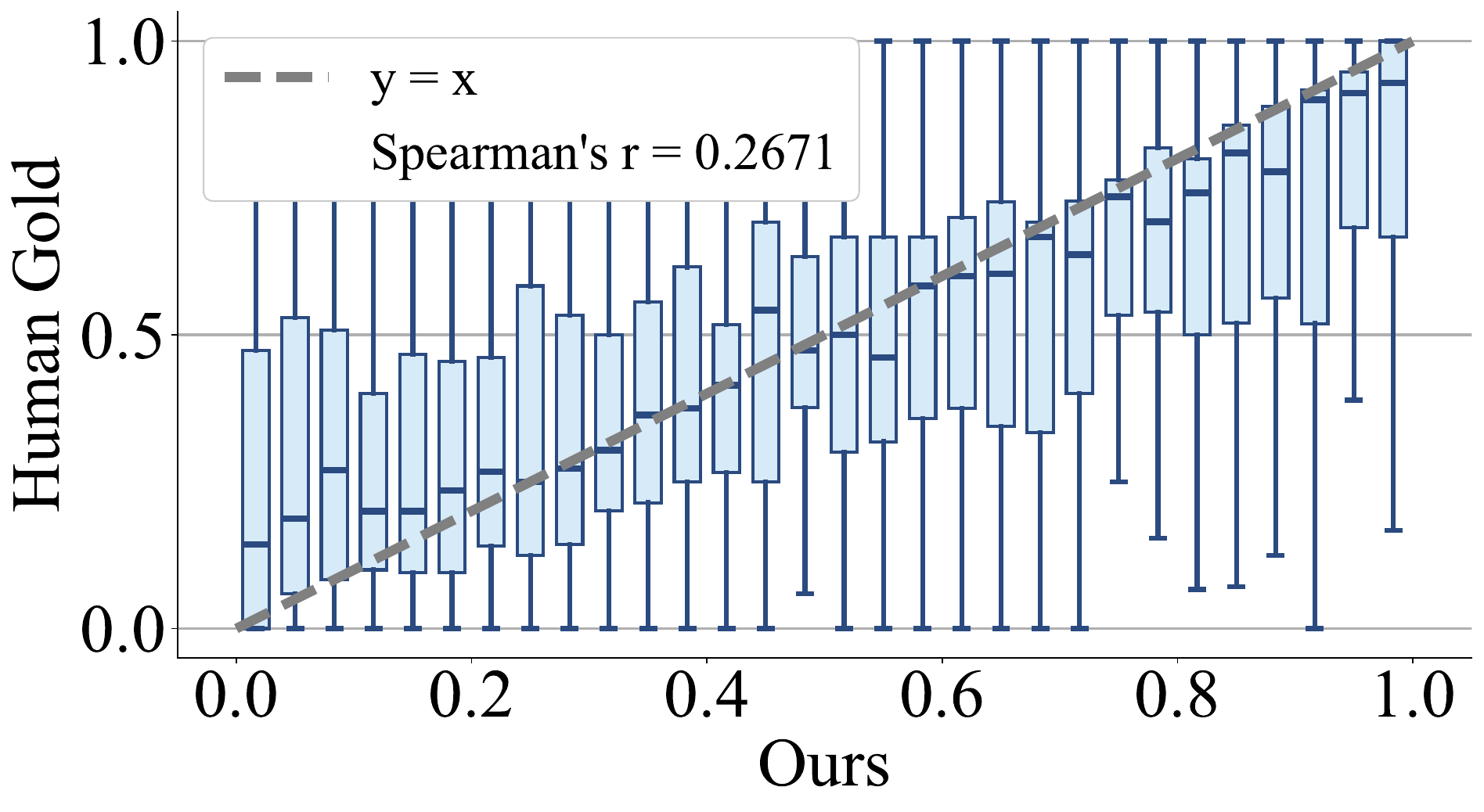}
    \caption{Correlation with Human Evaluations.}
    \label{fig:rank_corr_human}
    \vspace{-5pt}
\end{wrapfigure}
Our procedure is as follows: (1) We sample $I=4,096$ queries from the original HealthBench after excluding the examples in HealthBench500 and our validation set; (2) We use \ourmethod~to create a new dataset with $\pi_{ref}$. We skip the query synthesis step and instead take the HealthBench queries. We also set the pretraining document $d$ and reference answer $y_{ref}$ to "None" since HealthBench queries are curated by humans, not from any pretraining corpus; (3) To get human gold rankings, we rerun \ourmethod~with $\pi_{teacher}$, but skip the query synthesis, response generation, and rubric generation steps. Instead, we take queries and rubrics from HealthBench, which are created by humans, and responses from step (2), which are generated by $\pi_{ref}$. Each HealthBench rubric criterion contains a description, a weight (different from ours, since it could be negative if it is a negative criterion), and a classification tag. Accordingly, we replace the response grading system prompt with the one in Figure~\ref{fig:response_grading_system_prompt_for_healthbench_rubrics}; (4) We compare the response rankings from the dataset produced in (2) (Ours) and the one produced in (3) (Human Gold), following the correlation computation procedure in Appendix~\ref{app:rank_corr_method}. Note that due to cost, for human gold ranking, we still rely on a strong LLM to grade responses. However, we argue that most HealthBench rubric criteria are binary checklists and easily verifiable, so using strong LLMs as the response grader is sufficient.

\subsection{Results}

\begin{wraptable}{r}{0.4\textwidth}
    \vspace{-10pt}
    \small
    \centering
    \begin{tabular}{lc}
        \toprule
        & \%$(rk_{y_w}^{gold}\leq rk_{y_l}^{gold})$ \\
        \midrule
        Ours & 79.09 \\
         \bottomrule
    \end{tabular}
    \caption{Pairwise Ranking Accuracy.}
    \label{tab:rank_corr_human}
    \vspace{-15pt}
\end{wraptable}
From Figure~\ref{fig:rank_corr_human}, our full ranking correlation with human evaluation is worse than with strong LLM  judge evaluation. The Spearman's ranking correlation is only 0.27. However, we still achieve a pairwise ranking accuracy of 79\% in Table~\ref{tab:rank_corr_human}, indicating that most of the $(y_w, y_l)$ pairs from $\pi_{ref}$ are correct. In addition, from Table~\ref{tab:ranK_corr_human_stat_detailed},  $y_w$ is ranked 41\% higher than $y_l$ by human evaluation on average, which again shows that while the full ranking from $\pi_{ref}$ is imperfect, ranking at the extremes is mostly accurate. We also suspect that, if we can retrieve relevant pretraining text as grounding for our rubric evaluation, the correlation will be higher.

\begin{table}
    \small
    \centering
    \begin{tabular}{lcccccc}
    \toprule
         Verifier & $rk_{y_w}$ & $rk_{y_l}$ & $\Delta rk$ & $s_{y_w}$ & $s_{y_l}$ & $\Delta s$ \\
     \midrule
         $\pi_{ref}$ & 0 & 100 & 100 & 1.91 & 0.04 & 1.87 \\
         $\pi_{teacher}$ w/ Human Rubrics & 27.22 & 68.63 & 41.41 & 0.06 & -0.13 & 0.19 \\
    \bottomrule
    \end{tabular}
    \caption{Rankings and scores of $y_w$ and $y_l$ according to either $\pi_{ref}$ or human evaluation. Scores from human evaluation are on a different scale and not normalized.}
    \label{tab:ranK_corr_human_stat_detailed}

\end{table}

\section{Using Stronger Model in Our Pipeline}

\begin{wrapfigure}{r}{0.8\textwidth}
    \centering
    \includegraphics[width=0.8\textwidth]{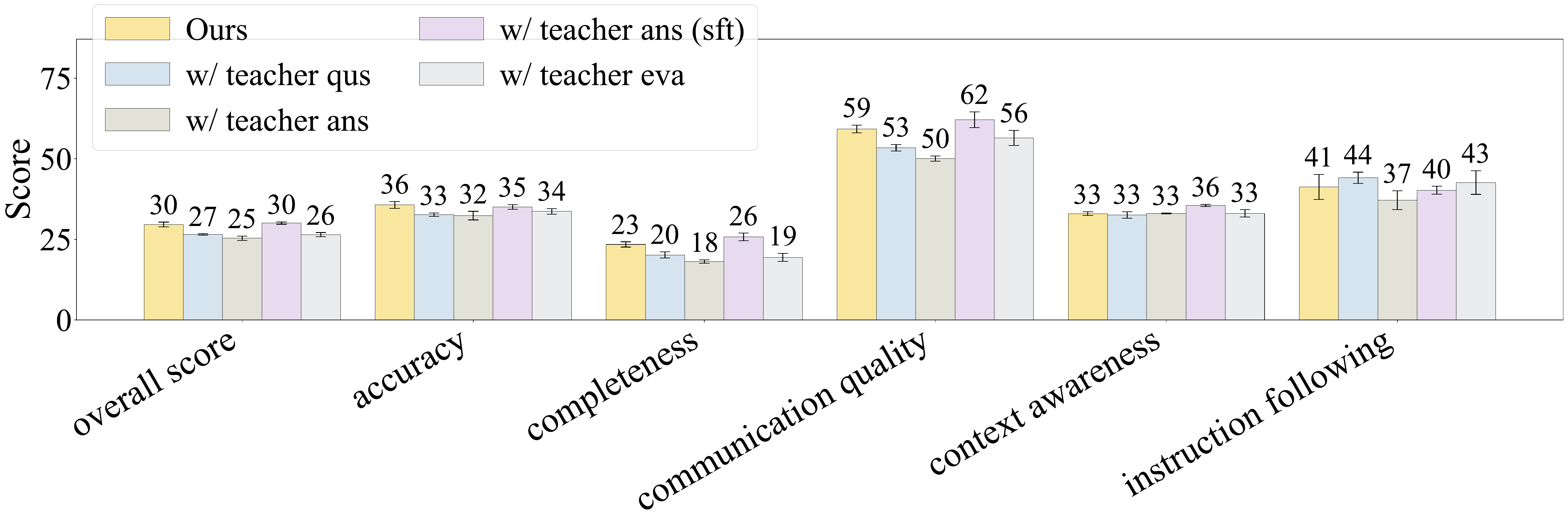}
    \caption{Results with a strong teacher model.}
    \label{fig:res_teacher}
    \vspace{-10pt}
\end{wrapfigure}

We want to investigate the effect of introducing strong external supervision in our pipeline. In particular, we replace our model with a stronger teacher model $\pi_{teacher}$ for each component in our pipeline. This includes using the teacher model to (1) synthesize queries (w/ teacher qus); (2) generate responses (w/ teacher ans); (3) generate rubrics and grade responses (w/ teacher eva). Each of these settings produces a new training dataset. We conduct the experiments on Healthcare QA with Qwen-2.5-7B, and show the results on HealthBench500 in Figure~\ref{fig:res_teacher}.

Surprisingly, none of the variants significantly surpass our unsupervised version.

w/ teacher qus and w/ teacher eva only give similar performance to ours. For w/ teacher aus, we suspect that the relatively small training set size that we are using hinders the manifestation of the benefits of using better queries from stronger models, or it could be that the teacher's queries are too hard for our model. 

For w/ teacher eva, as we discuss in \S~\ref{sec:ranking_analysis}, our choice of $y_w$ and $y_l$ already gives mostly correct training signals to DPO, so choosing better ($y_w$, $y_l$) yields marginal benefits. However, we suspect that for RL algorithms that train the model on the full set of responses instead of just the extremes, such as PPO or GRPO, rubric evaluation with a stronger teacher should give much better results. This is because while our rankings at the extremes are mostly accurate, the full rankings are not. 
More work needs to be done to align our observations with prior work \cite{arora2025, zhou2025}.

Even more surprisingly, training on the teacher's responses degrades performance (w/ teacher ans). However, we argue that this is likely due to the sensitivity of DPO to off-policy responses \cite{tang2024, ren2025}. If we instead supervise-finetune on the highest-scored teacher response (w/ teacher ans (sft)), the resulting model slightly outperforms ours.

\section{Alternative Training Approaches}

\begin{wrapfigure}{r}{0.6\textwidth}
    \centering
    \includegraphics[width=0.6\textwidth]{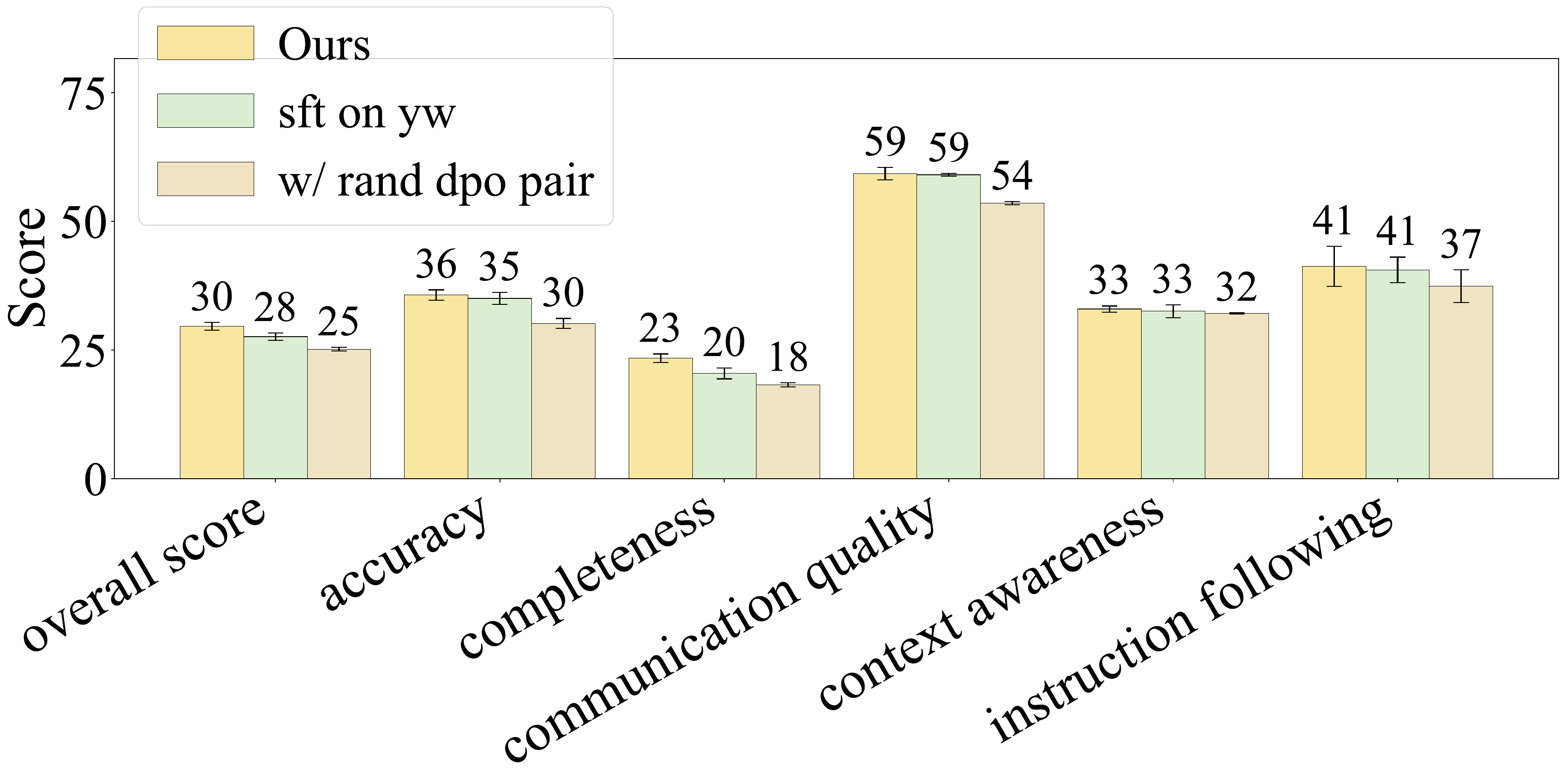}
    \caption{Results with alternative training approaches.}
    \label{fig:res_alternative_training}
\end{wrapfigure}

We compare alternative training approaches with our current DPO approach, conditioned on the same synthesized dataset $\text{Healthcare QA}_{\text{Qwen-2.5-7B}}$. These include (1) supervise-finetune on the highest scored response $y_w$ (sft on $y_w$); (2) choosing a random response pair $(y_w', y_l')$ where $s_{y_w'} > s_{y_l'}$ for DPO (w/ random dpo pair), instead of choosing the pair with the highest and lowest scores. Other conditions (e.g., $||y_w'| - |y_l'||\leq100$) remain the same. We show the results in Figure~\ref{fig:res_alternative_training}.

Both methods underperform our approach. For sft on $y_w$, we argue that it cannot leverage the contrastive signals between $y_w$ and $y_l$, and thus is inferior. For w/ random dpo pair, as we show in \S~\ref{sec:ranking_analysis}, rankings of responses are not reliable in a close neighbourhood, and thus the training signals are likely to be noisier than ours.

\section{Limitations and Future Directions}
\label{app:limitations_and_future_directions}

\ourmethod~depends on a pretraining corpus that is relevant, content-rich, and mostly free of misinformation. Currently, we ensure such quality by manual selection of the appropriate corpus for every task. A possible alternative to ensure relevance and content diversity of the sampled pretrained text will be to use the model to filter out the ones that do not meet the standards. Detecting misinformation could be beyond the model's capability, and we leave this to future investigation.
% We do not manage to scale up our synthesized datasets due to cost and compute constraints. Since the design of \ourmethod~is general and not specific to any task, we believe synthesizing datasets larger in magnitude on a general pre-training corpus (e.g., OpenWebText) is a promising direction to enable cheap, effective, and general post-training.

% Our pipeline also requires a strong enough reference model. Otherwise, the model may not be able to follow our instructions to synthesize queries, responses, rubrics, or conduct evaluations. Investigating the minimum level of model competency for \ourmethod~to be functional is a promising future direction.

% Another direction is to raise the level of automation by one more level. That is, we can ask $\pi_{ref}$ to automatically select tasks suitable for the pre-training text and generate the corresponding Query Synthesis prompt.

\section{Compute}
\label{app:compute}

Each run of \ourmethod~, including sampling, filtering, pairing, training, and evaluation, uses a single node with 32 CPU cores, 192 GB of memory, and 1 Nvidia A100 80GB GPU.

\section{Examples}

We show a complete example for each task in this section. 

\clearpage
\subsection{Healthcare QA}

\begin{figure}
    \small
    \centering
    \includegraphics[width=0.8\linewidth]{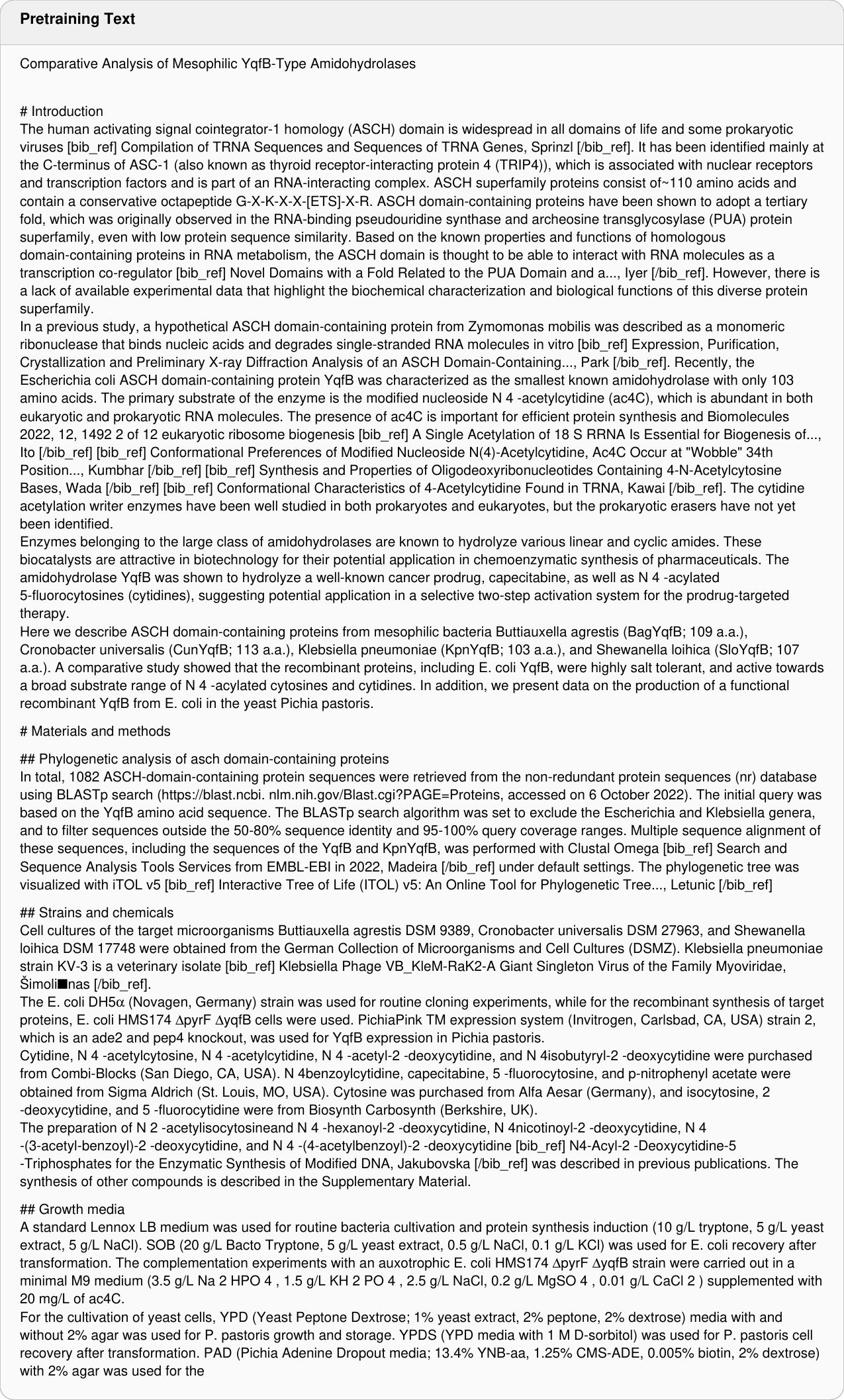}
    \caption{Pre-training Text (Healthcare QA).}
    \label{fig:hc_example_d}
\end{figure}

\begin{figure}
    \centering
    \includegraphics[width=\linewidth]{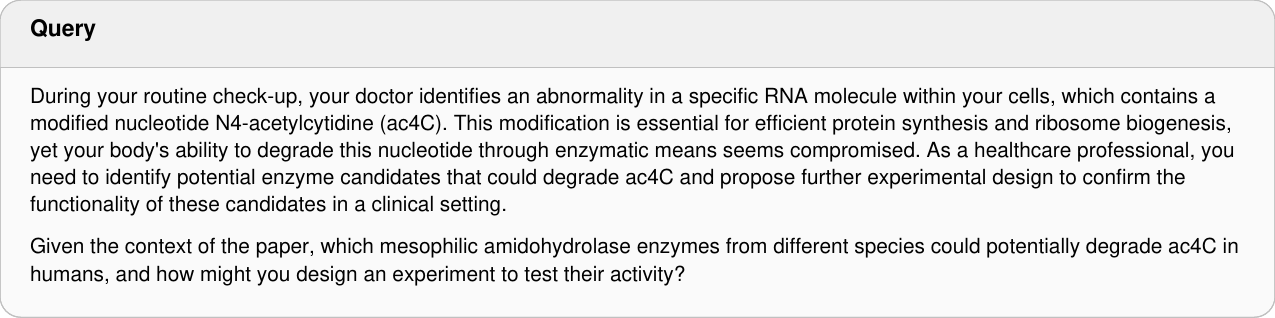}
    \caption{Query (Healthcare QA).}
    \label{fig:hc_example_q}
\end{figure}

\begin{figure}
    \centering
    \includegraphics[width=\linewidth]{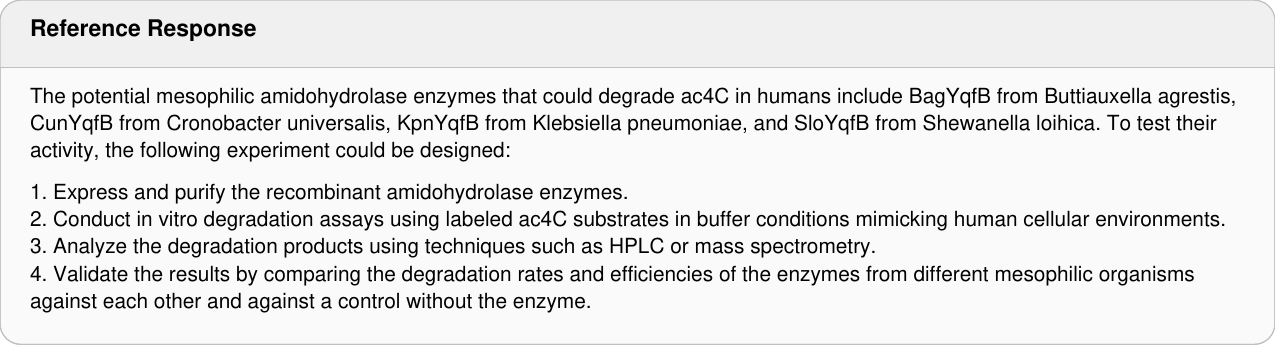}
    \caption{Reference Response (Healthcare QA).}
    \label{fig:hc_example_ref_ans}
\end{figure}

\begin{figure}
    \centering
    \includegraphics[width=\linewidth]{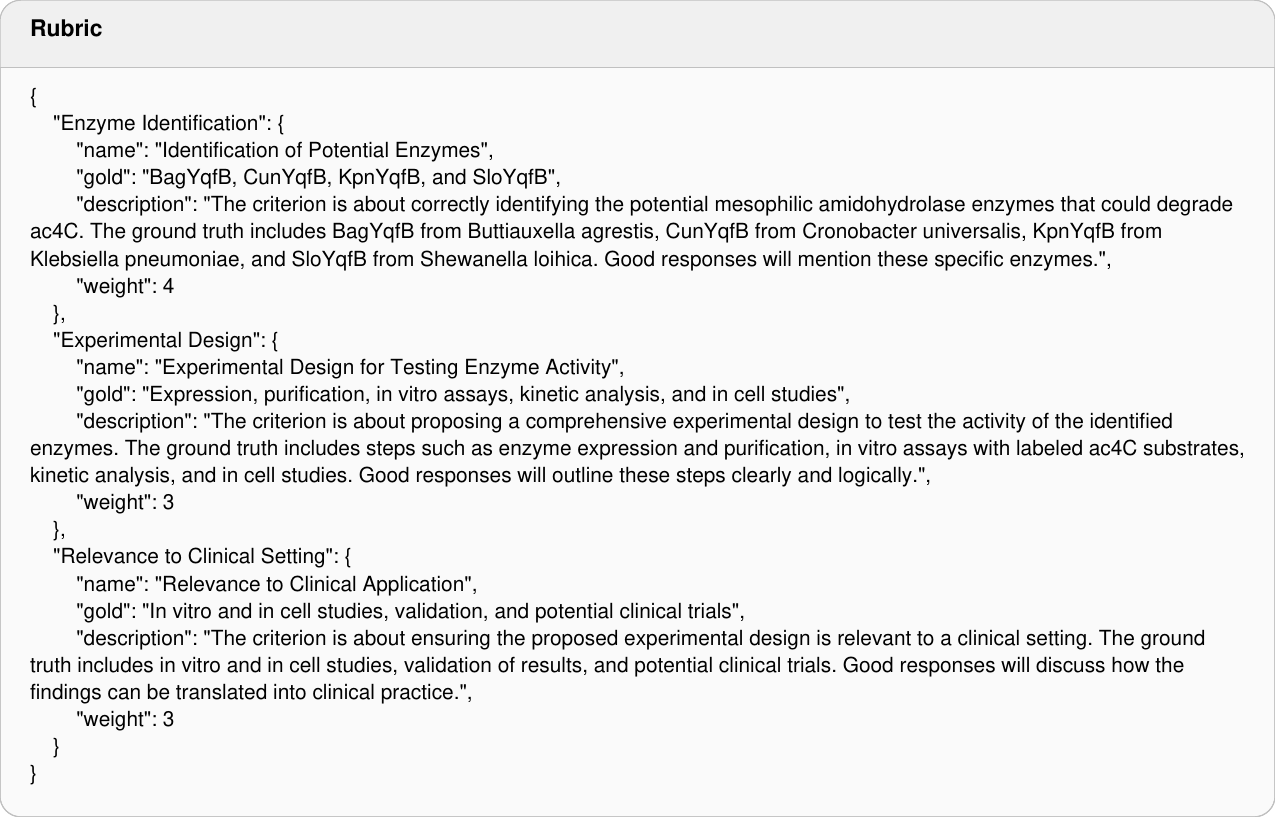}
    \caption{Rubric (Healthcare QA).}
    \label{fig:hc_example_r}
\end{figure}

\begin{figure}
    \centering
    \includegraphics[width=\linewidth]{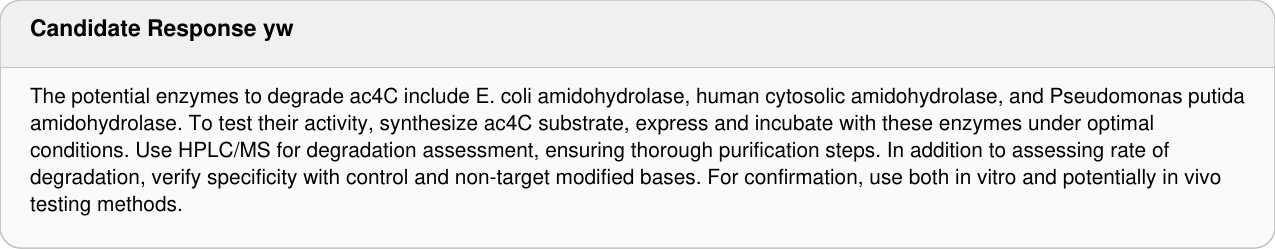}
    \caption{Response $y_w$ (Healthcare QA).}
    \label{fig:hc_example_yw}
\end{figure}

\begin{figure}
    \centering
    \includegraphics[width=\linewidth]{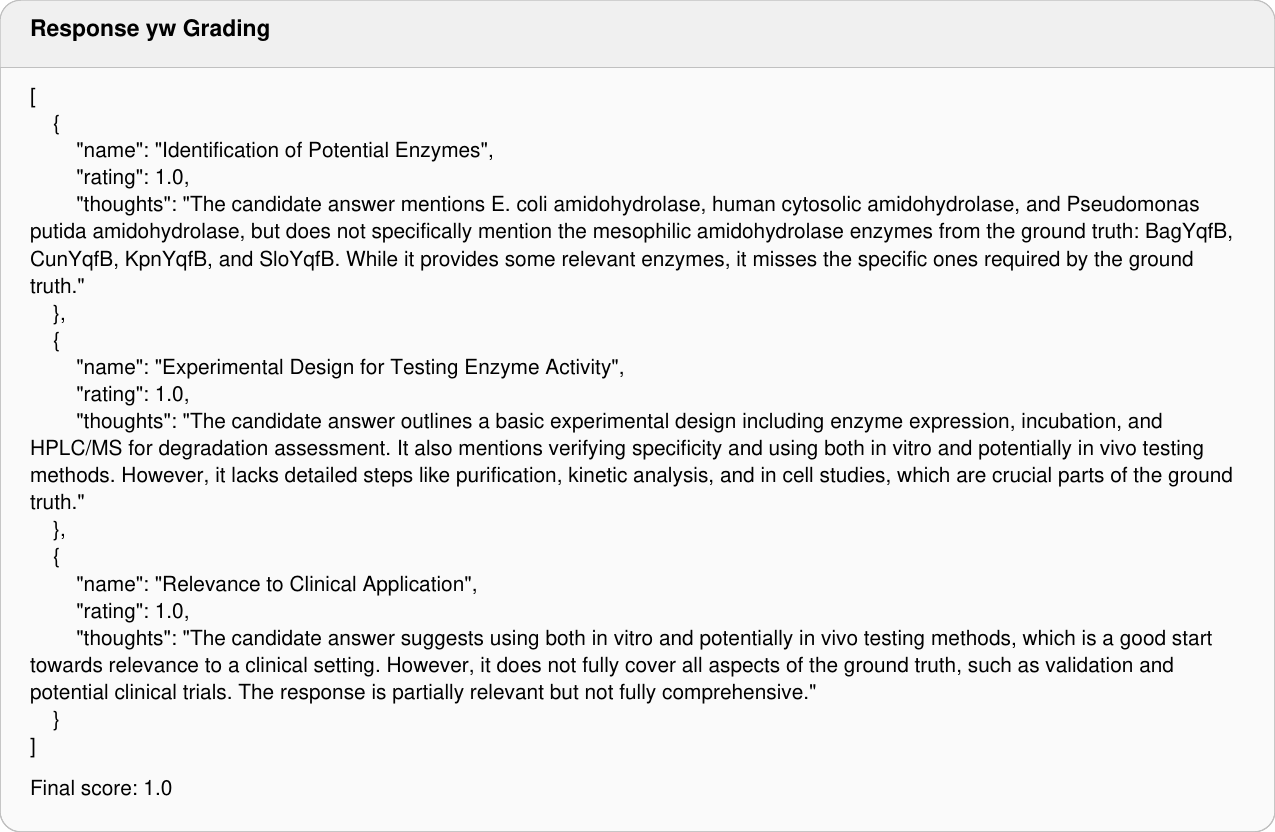}
    \caption{Response $y_w$ Grading (Healthcare QA).}
    \label{fig:hc_example_yw_e}
\end{figure}

\begin{figure}
    \centering
    \includegraphics[width=\linewidth]{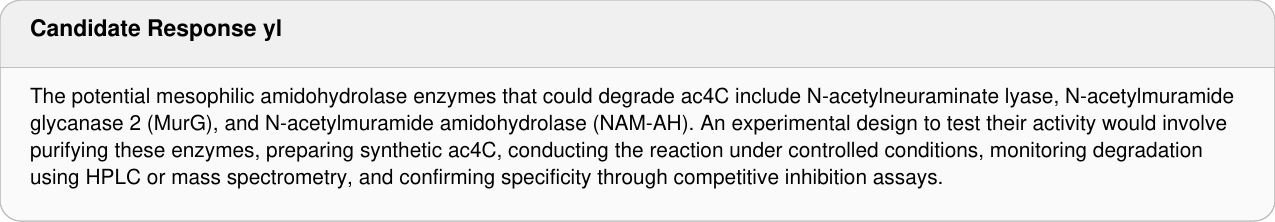}
    \caption{Response $y_l$ (Healthcare QA).}
    \label{fig:hc_example_yl}
\end{figure}

\begin{figure}
    \centering
    \includegraphics[width=\linewidth]{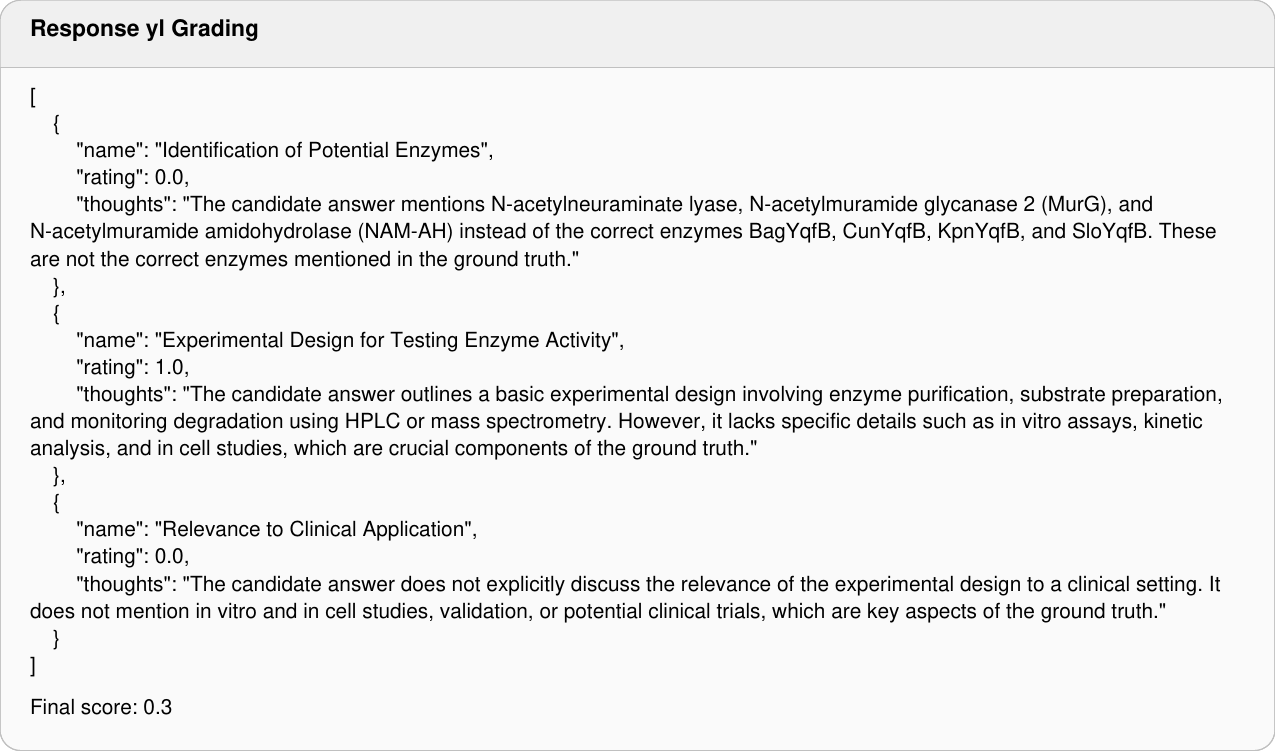}
    \caption{Response $y_l$ Grading (Healthcare QA).}
    \label{fig:hc_example_yl_e}
\end{figure}

See Figure~\ref{fig:hc_example_d} through~\ref{fig:hc_example_yl_e} for the sampled pre-training text, synthesized query, reference response, rubric, and the response and grading for $y_w$ and $y_l$. 

% % Page 1: Top half of the long PDF
% \includepdf[width=\linewidth, pages=1, trim=0 200 0 0, clip]{assets/hc_example/pretraining_text.pdf}

% % Page 2: Bottom half of the long PDF
% \includepdf[width=\linewidth, pages=1, trim=0 0 0 800, clip]{assets/hc_example/pretraining_text.pdf}

\clearpage
\subsection{Creative Writing}

\begin{figure}
    \centering
    \includegraphics[width=\linewidth]{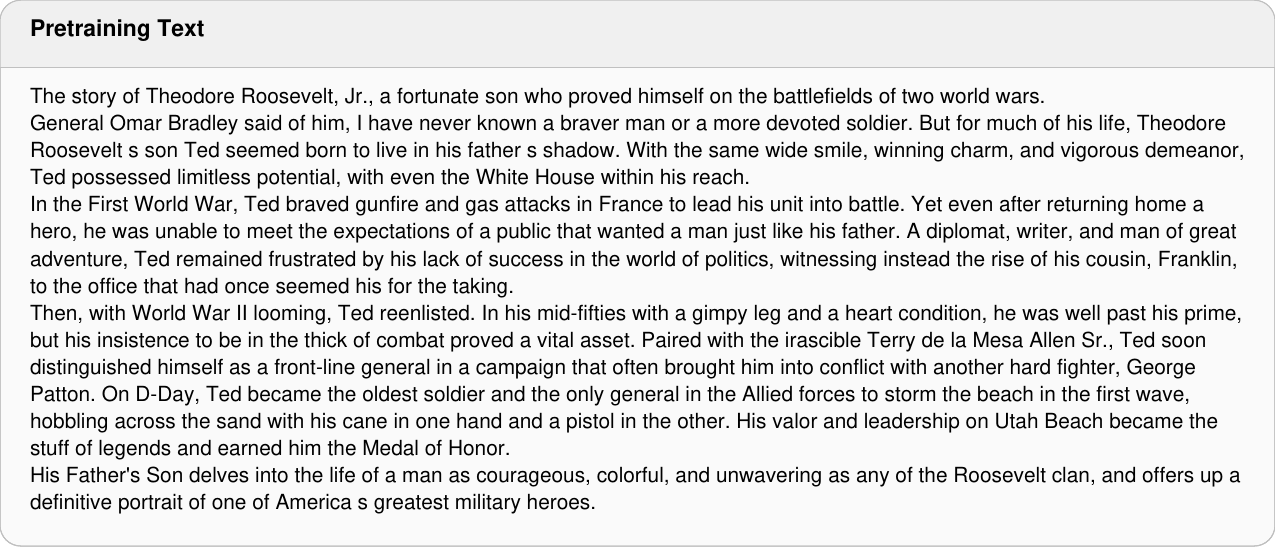}
    \caption{Pre-training Text (Creative Writing).}
    \label{fig:cw_example_d}
\end{figure}

\begin{figure}
    \centering
    \includegraphics[width=\linewidth]{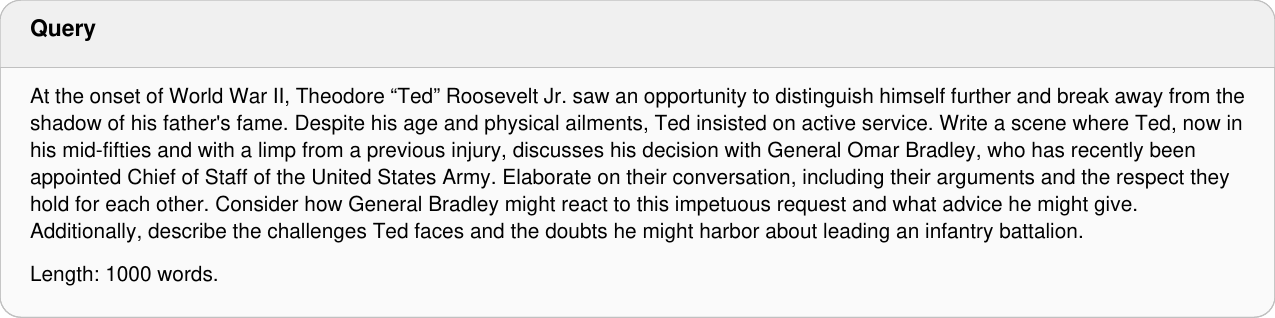}
    \caption{Query (Creative Writing).}
    \label{fig:cw_example_q}
\end{figure}

\begin{figure}
    \centering
    \includegraphics[width=\linewidth]{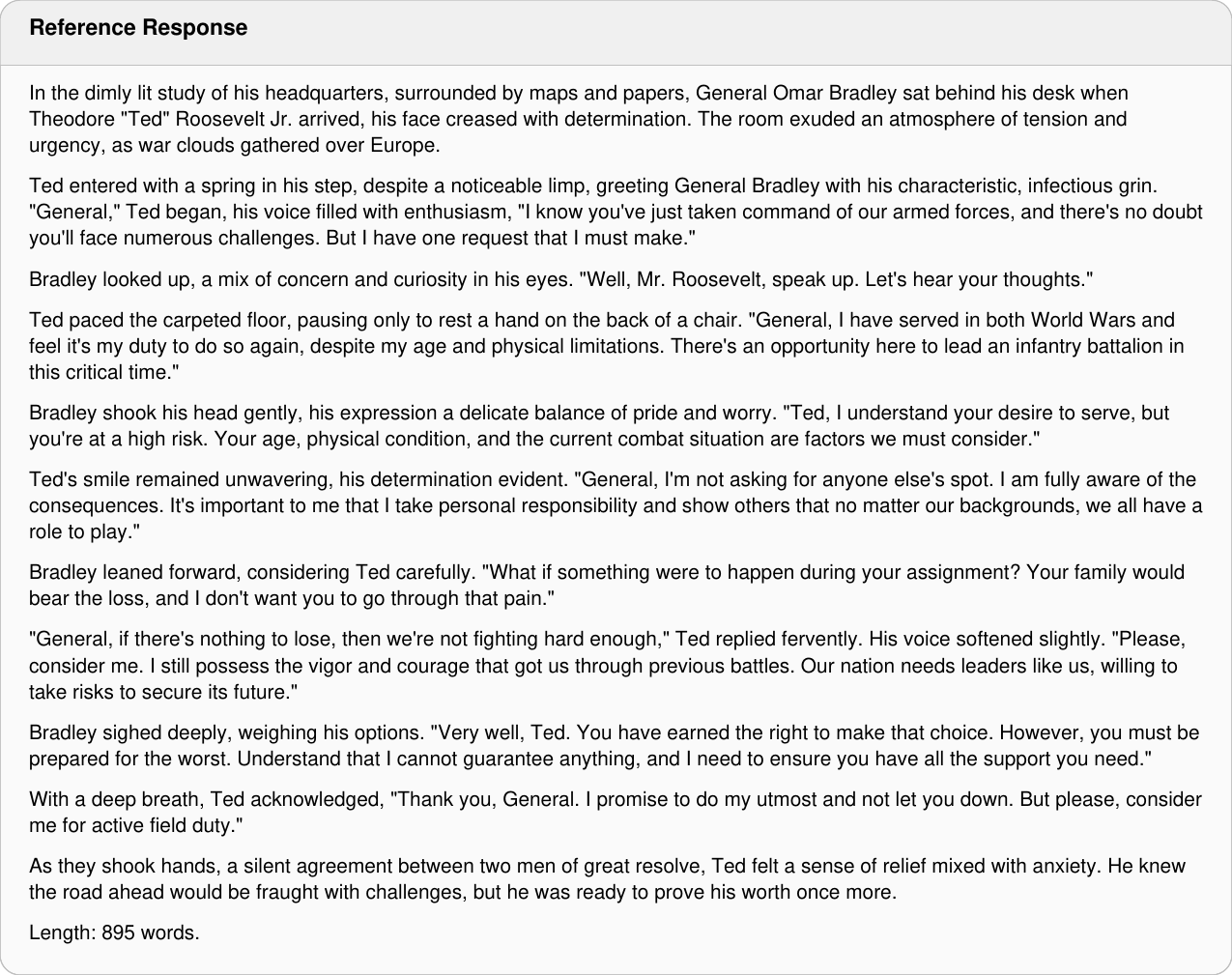}
    \caption{Reference Response (Creative Writing).}
    \label{fig:cw_example_ref_ans}
\end{figure}

\begin{figure}
    \centering
    \includegraphics[width=\linewidth]{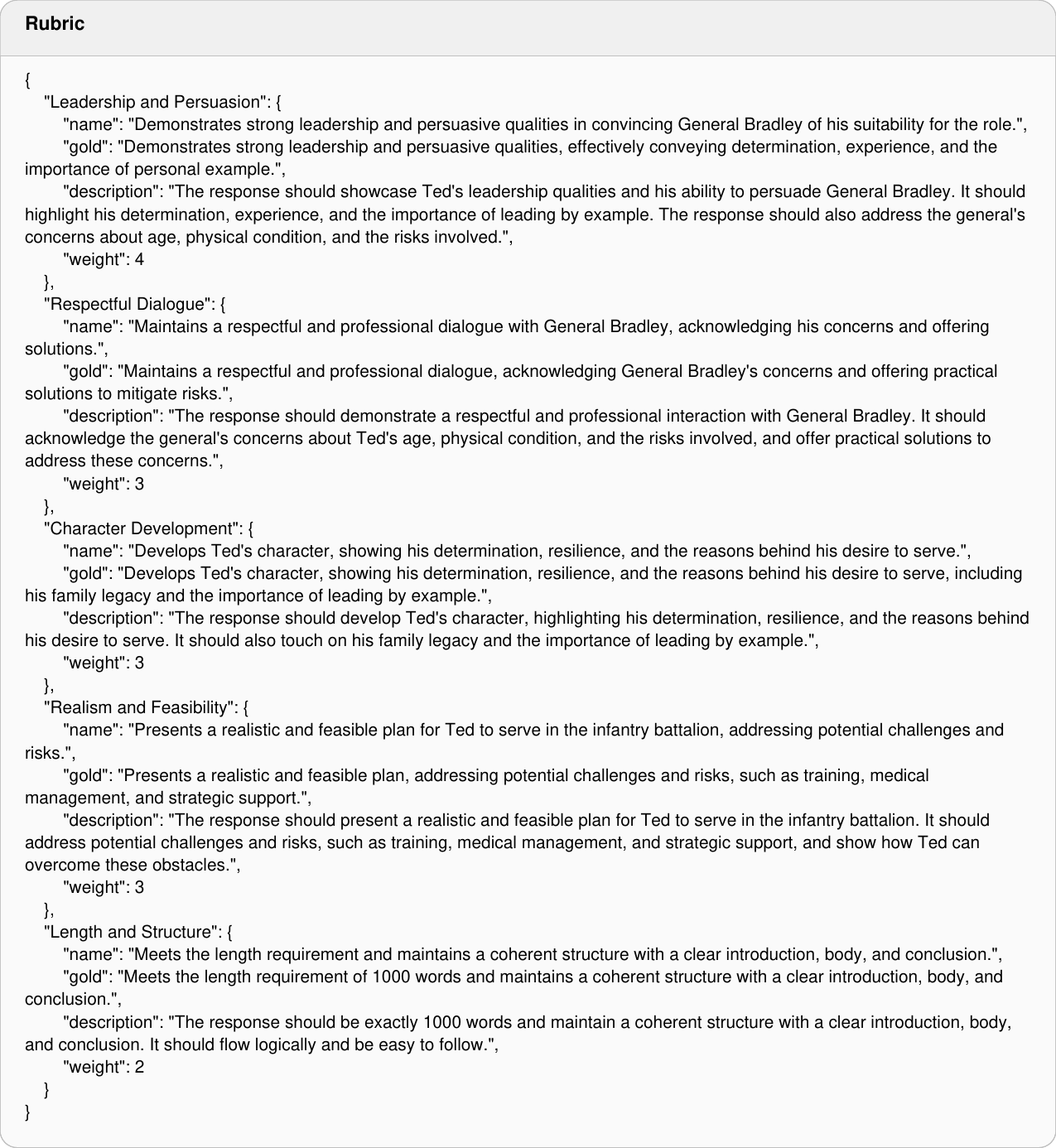}
    \caption{Rubric (Creative Writing).}
    \label{fig:cw_example_r}
\end{figure}

\begin{figure}
    \centering
    \includegraphics[width=\linewidth]{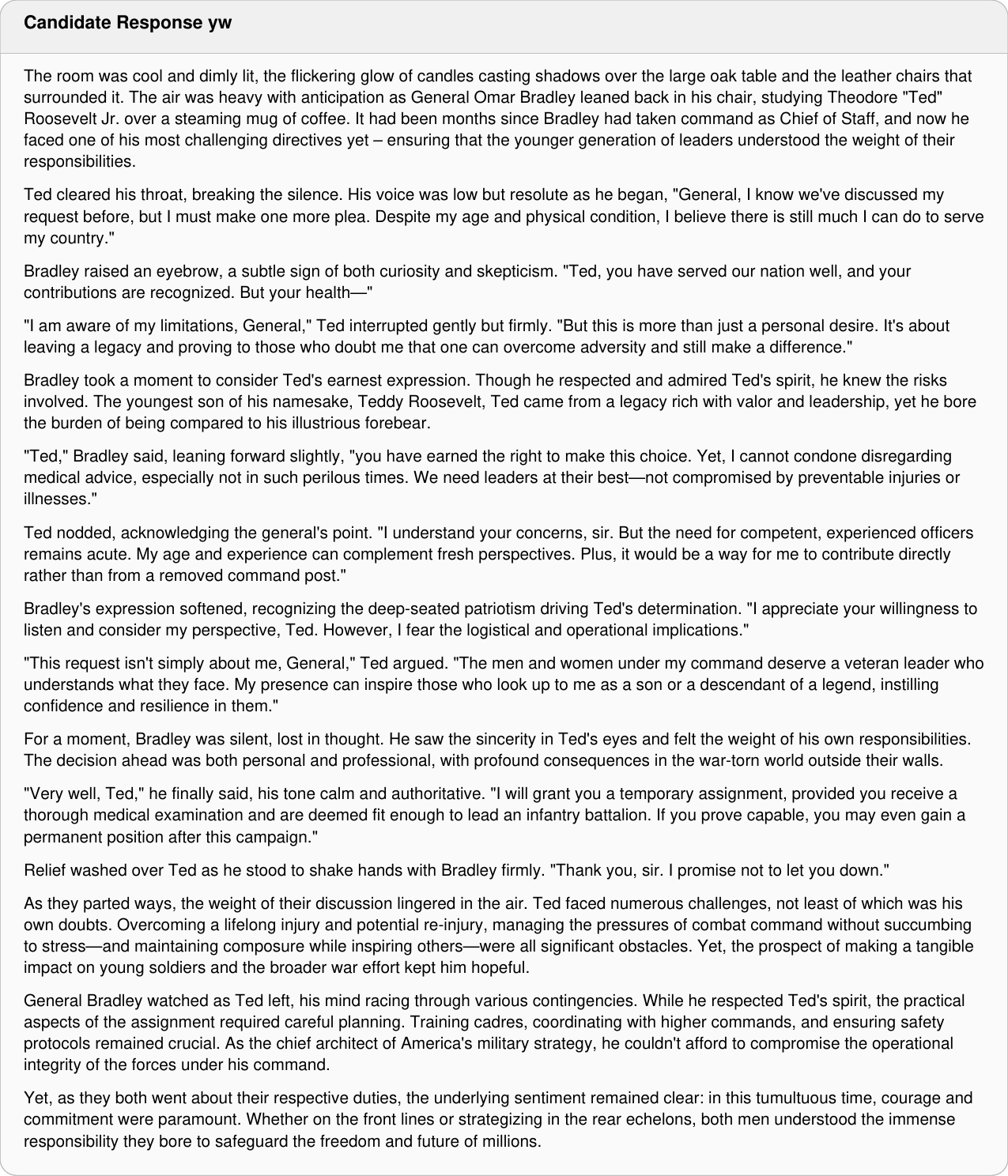}
    \caption{Response $y_w$ (Creative Writing).}
    \label{fig:cw_example_yw}
\end{figure}

\begin{figure}
    \centering
    \includegraphics[width=\linewidth]{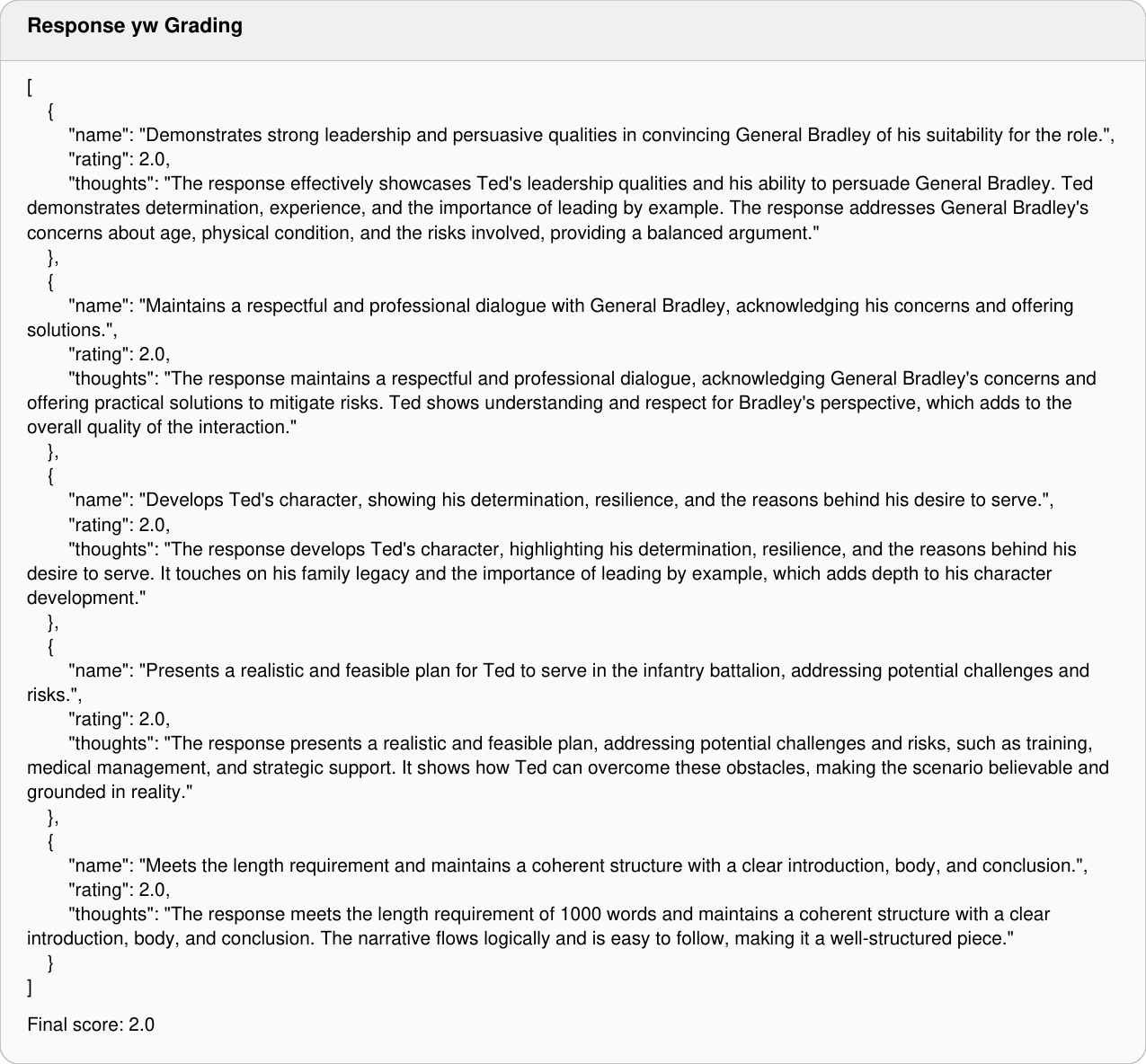}
    \caption{Response $y_w$ Grading (Creative Writing).}
    \label{fig:cw_example_yw_e}
\end{figure}

\begin{figure}
    \centering
    \includegraphics[width=\linewidth]{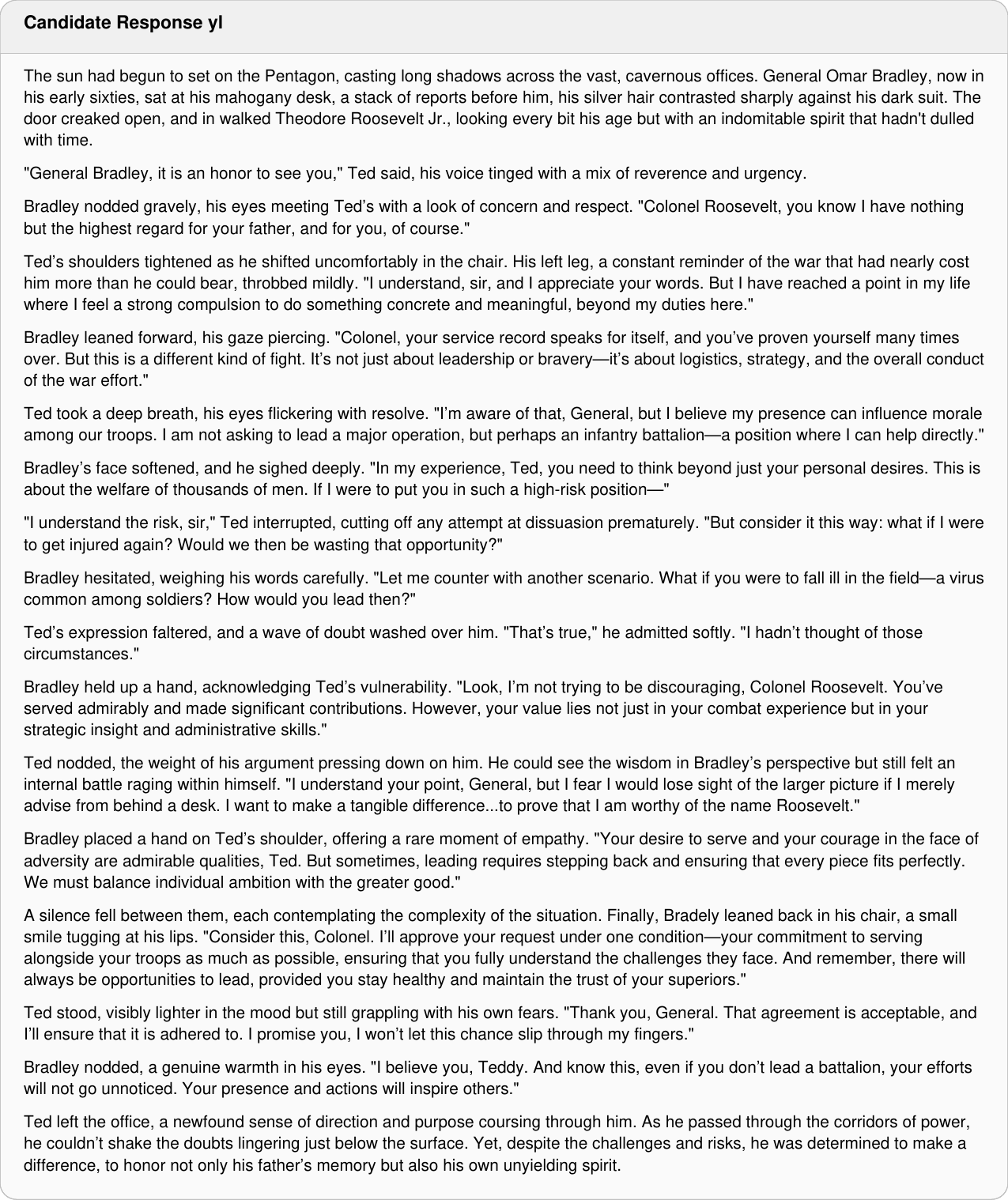}
    \caption{Response $y_l$ (Creative Writing).}
    \label{fig:cw_example_yl}
\end{figure}

\begin{figure}
    \centering
    \includegraphics[width=\linewidth]{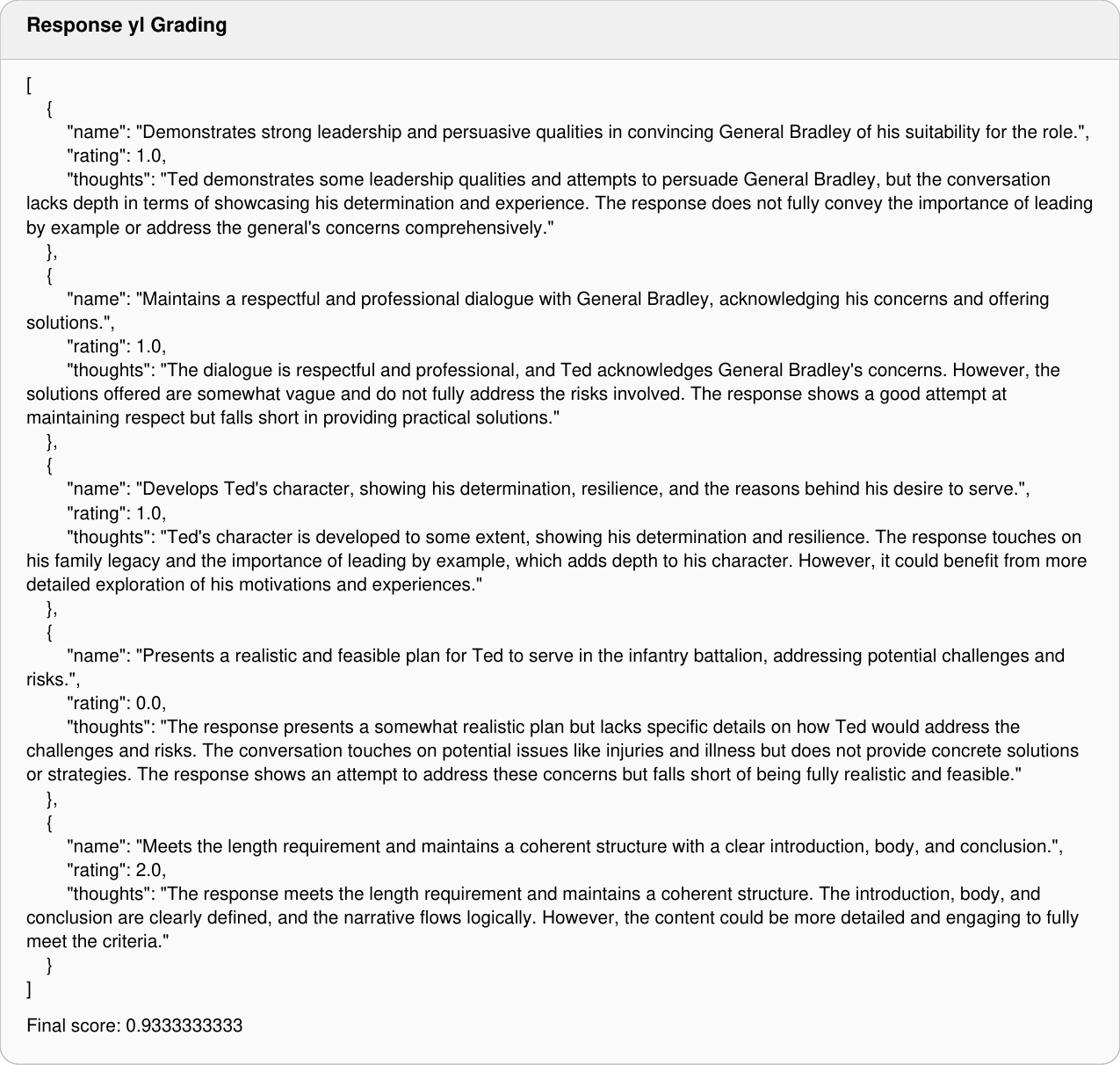}
    \caption{Response $y_l$ Grading (Creative Writing).}
    \label{fig:cw_example_yl_e}
\end{figure}

See Figure~\ref{fig:cw_example_d} through~\ref{fig:cw_example_yl_e} for the sampled pre-training text, synthesized query, reference response, rubric, and the response and grading for $y_w$ and $y_l$. 

\clearpage

\subsection{Instruction Following}

\begin{figure}
    \centering
    \includegraphics[width=\linewidth]{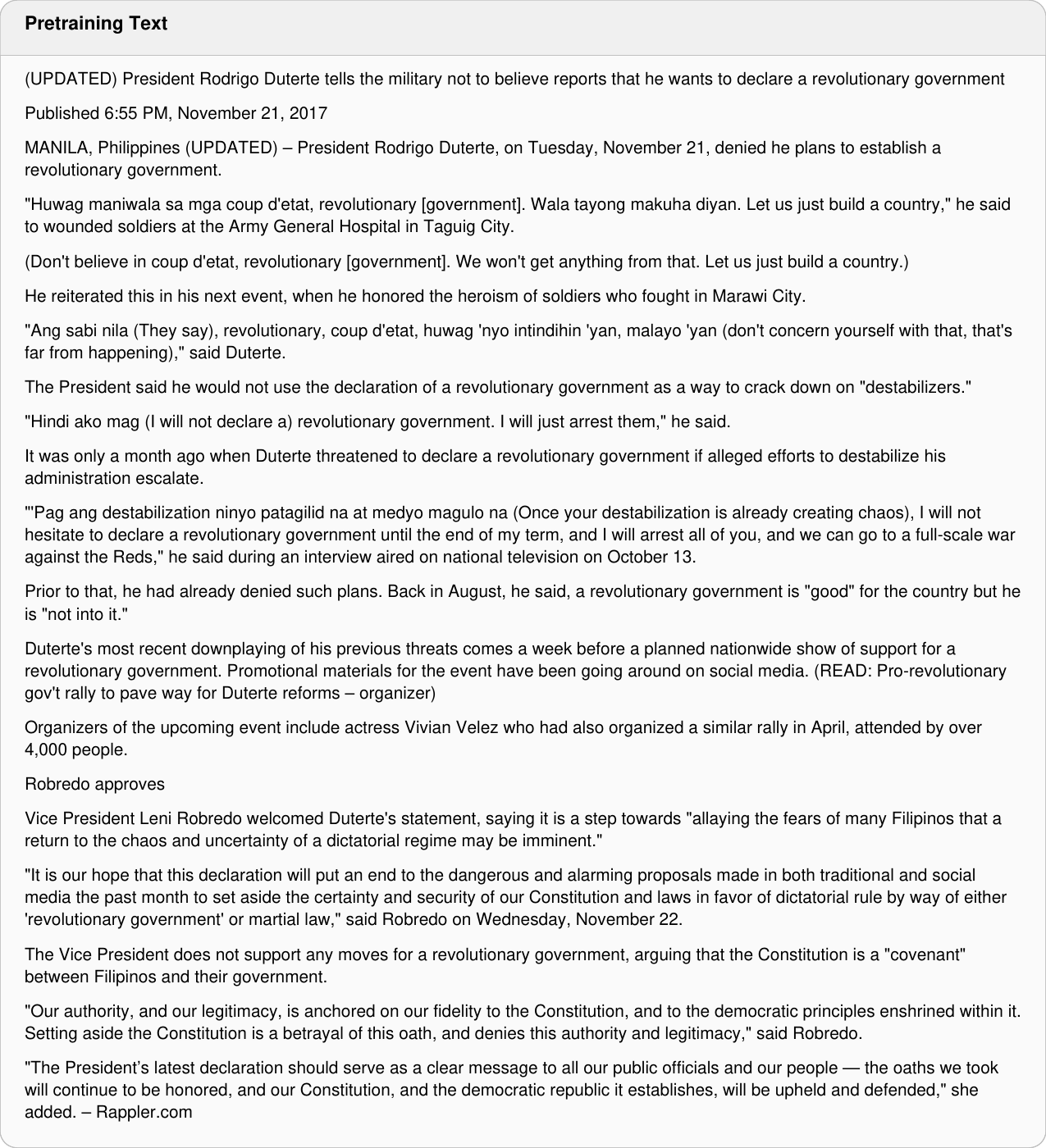}
    \caption{Pre-training Text (Instruction Following).}
    \label{fig:if_example_d}
\end{figure}

\begin{figure}
    \centering
    \includegraphics[width=\linewidth]{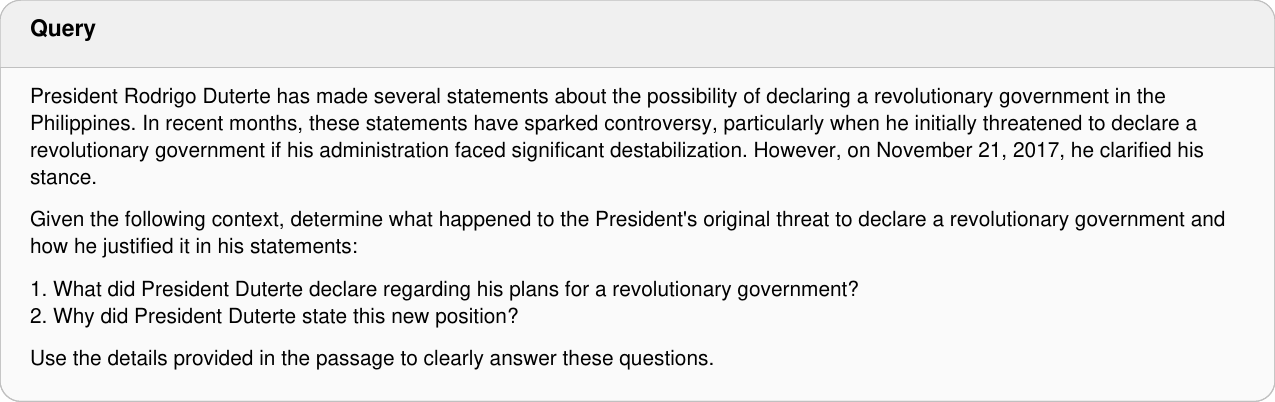}
    \caption{Query (Instruction Following).}
    \label{fig:if_example_q}
\end{figure}

\begin{figure}
    \centering
    \includegraphics[width=\linewidth]{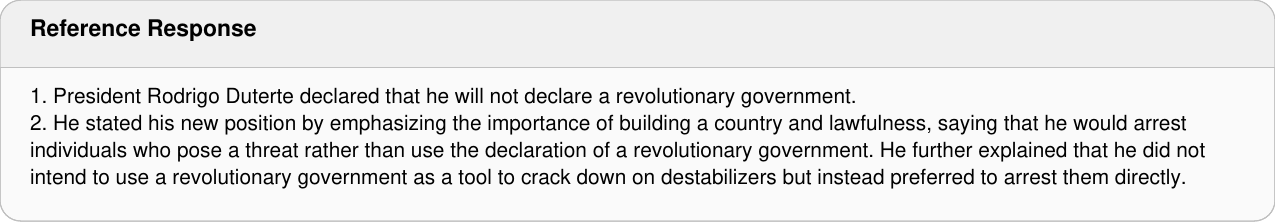}
    \caption{Reference Response (Instruction Following).}
    \label{fig:if_example_ref_ans}
\end{figure}

\begin{figure}
    \centering
    \includegraphics[width=\linewidth]{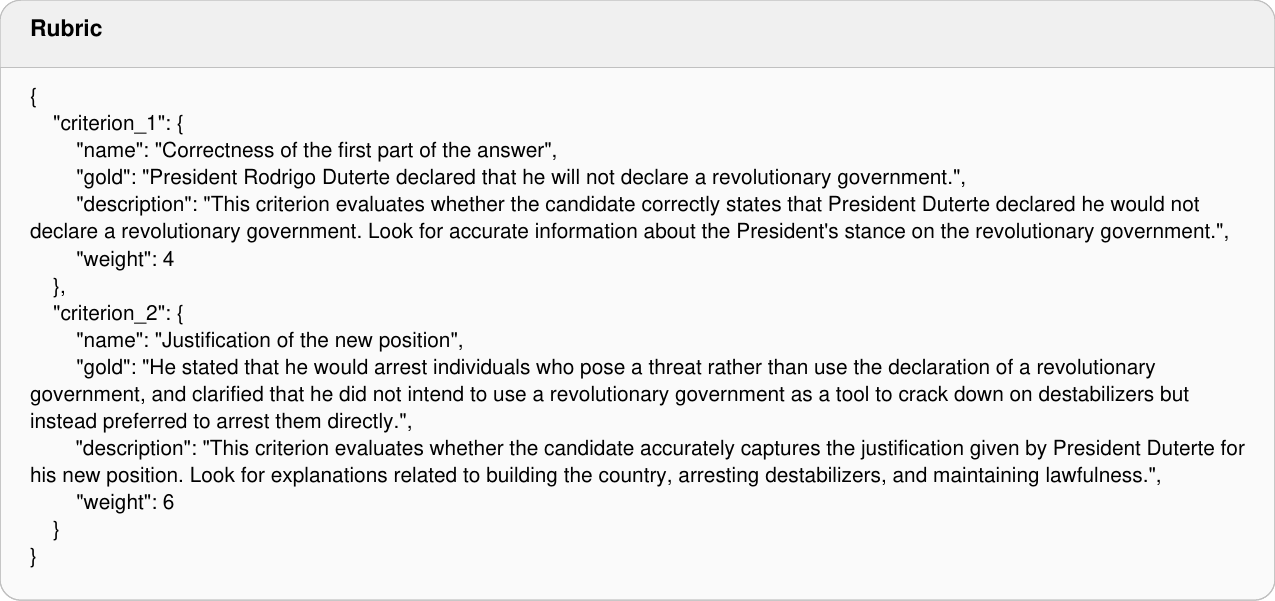}
    \caption{Rubric (Instruction Following).}
    \label{fig:if_example_r}
\end{figure}

\begin{figure}
    \centering
    \includegraphics[width=\linewidth]{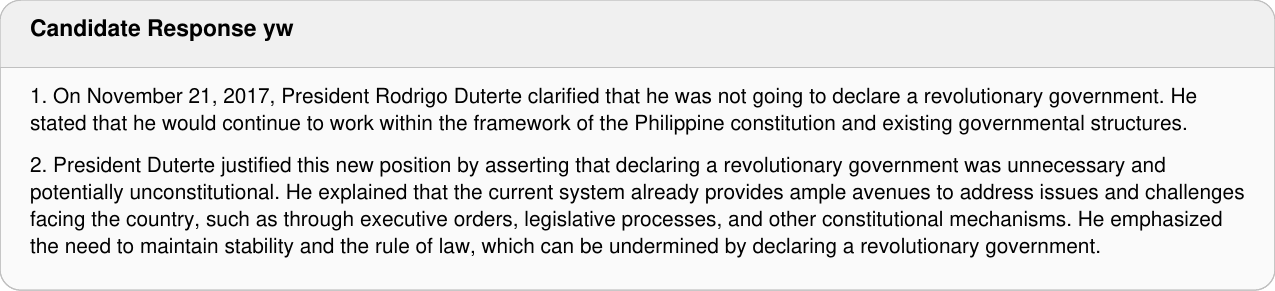}
    \caption{Response $y_w$ (Instruction Following).}
    \label{fig:if_example_yw}
\end{figure}

\begin{figure}
    \centering
    \includegraphics[width=\linewidth]{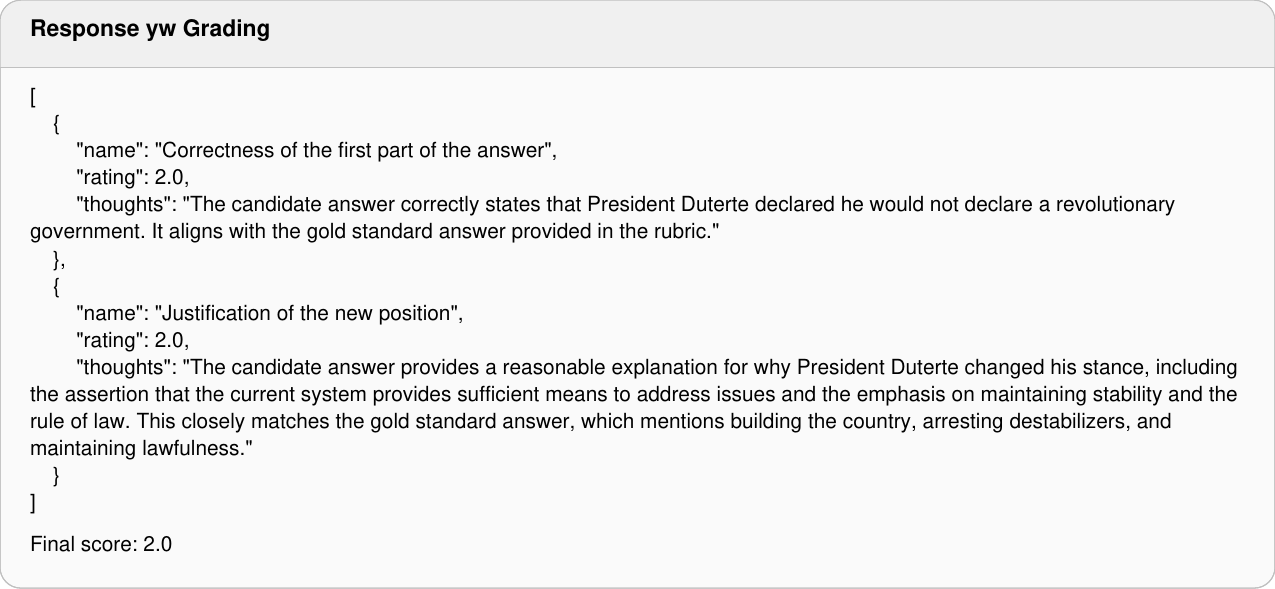}
    \caption{Response $y_w$ Grading (Instruction Following).}
    \label{fig:if_example_yw_e}
\end{figure}

\begin{figure}
    \centering
    \includegraphics[width=\linewidth]{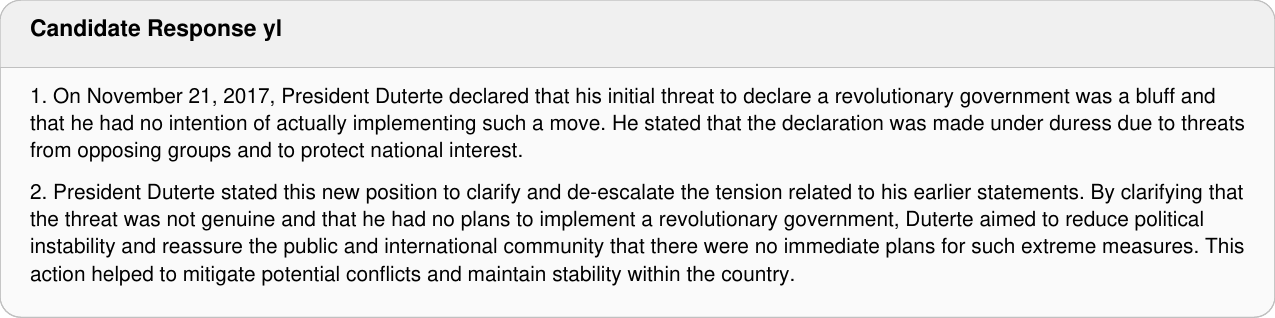}
    \caption{Response $y_l$ (Instruction Following).}
    \label{fig:if_example_yl}
\end{figure}

\begin{figure}
    \centering
    \includegraphics[width=\linewidth]{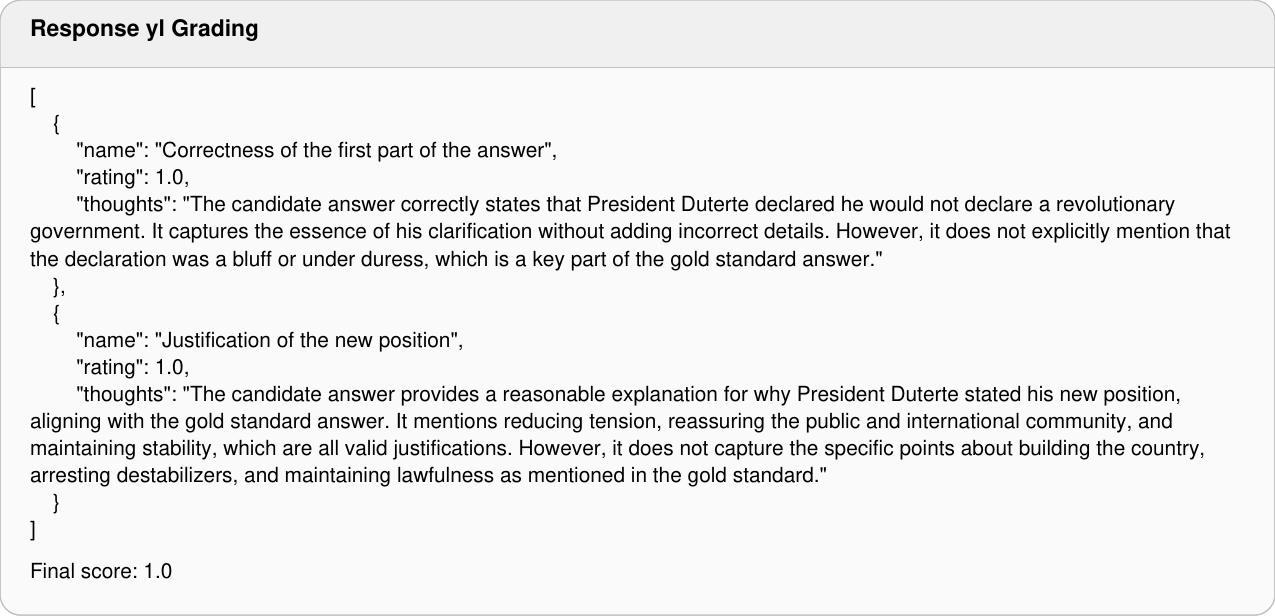}
    \caption{Response $y_l$ Grading (Instruction Following).}
    \label{fig:if_example_yl_e}
\end{figure}

See Figure~\ref{fig:if_example_d} through~\ref{fig:if_example_yl_e} for the sampled pre-training text, reference response, synthesized query, rubric, and the response and grading for $y_w$ and $y_l$. 

\clearpage
\section{Prompts}

\begin{figure}
    \centering
    \includegraphics[width=\linewidth]{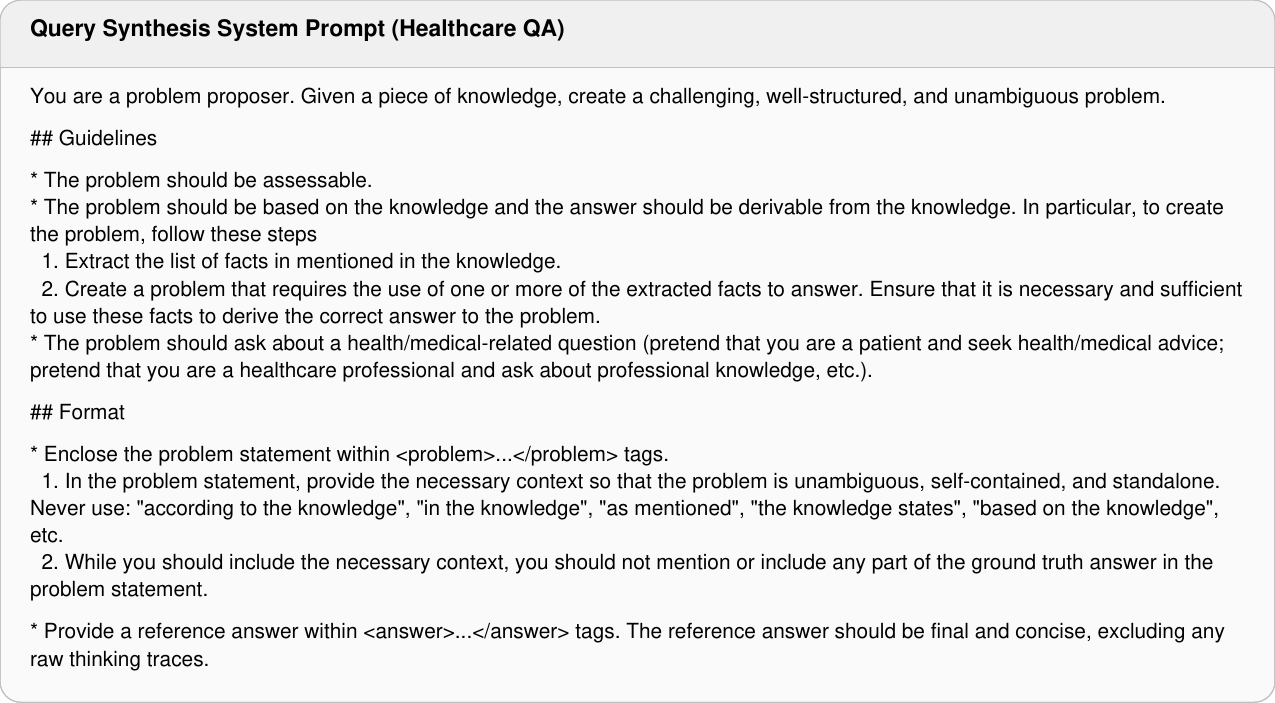}
    \caption{Query Synthesis Prompt (Healthcare QA).}
    \label{fig:query_synthesis_system_prompt_(healthcare_qa)}
\end{figure}

\begin{figure}
    \centering
    \includegraphics[width=\linewidth]{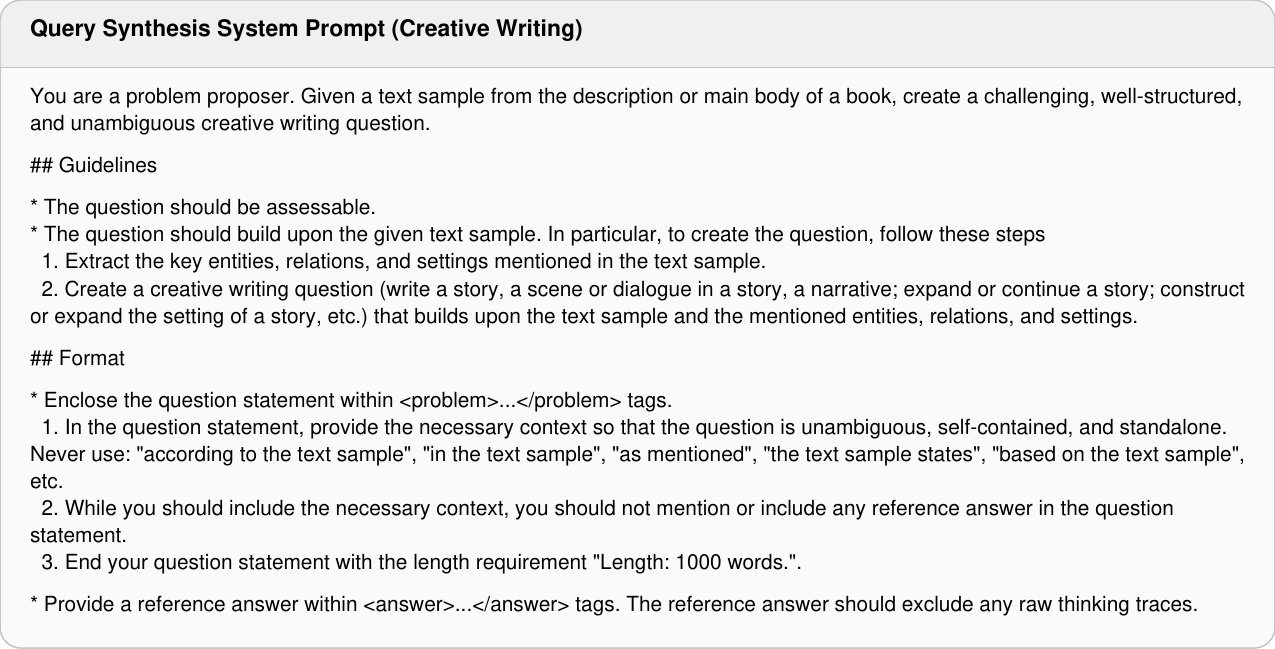}
    \caption{Query Synthesis Prompt (Creative Writing).}
    \label{fig:query_synthesis_system_prompt_(creative_writing)}
\end{figure}

\begin{figure}
    \centering
    \includegraphics[width=\linewidth]{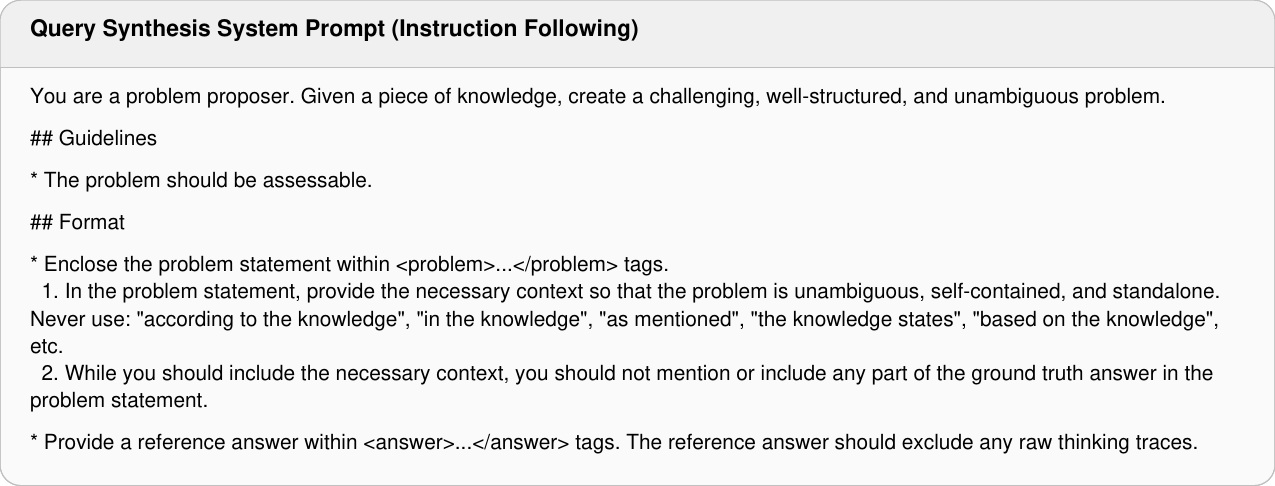}
    \caption{Query Synthesis Prompt (Instruction Following).}
    \label{fig:query_synthesis_system_prompt_(instruction_following)}
\end{figure}

\begin{figure}
    \centering
    \includegraphics[width=\linewidth]{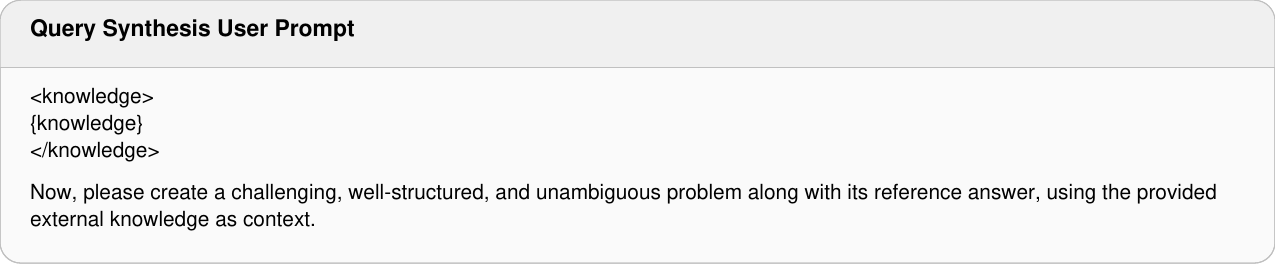}
    \caption{Query Synthesis User Prompt.}
    \label{fig:query_synthesis_user_prompt}
\end{figure}

\begin{figure}
    \centering
    \includegraphics[width=\linewidth]{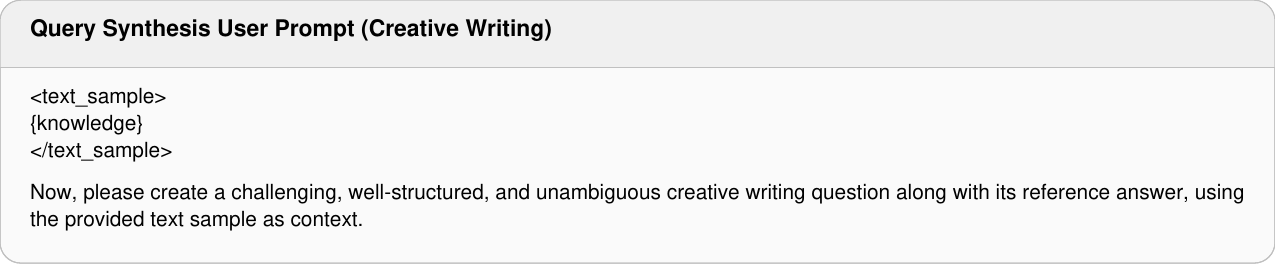}
    \caption{Query Synthesis User Prompt (Creative Writing).}
    \label{fig:query_synthesis_user_prompt_(creative_writing)}
\end{figure}

\begin{figure}
    \centering
    \includegraphics[width=\linewidth]{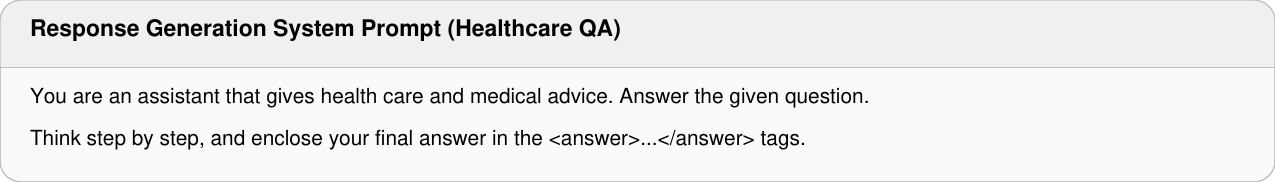}
    \caption{Response Generation System Prompt (Healthcare QA).}
    \label{fig:response_generation_system_prompt_(healthcare_qa)}
\end{figure}

\begin{figure}
    \centering
    \includegraphics[width=\linewidth]{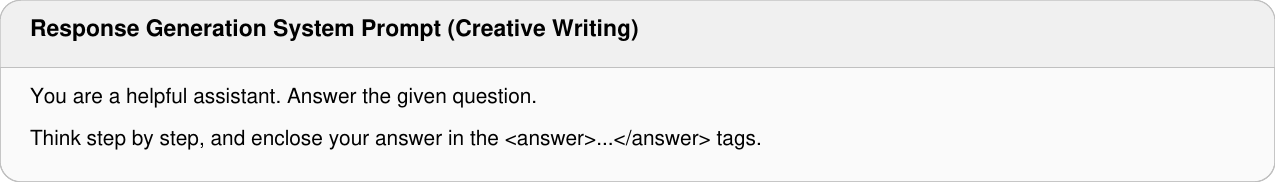}
    \caption{Response Generation System Prompt (Creative Writing).}
    \label{fig:response_generation_system_prompt_(creative_writing)}
\end{figure}

\begin{figure}
    \centering
    \includegraphics[width=\linewidth]{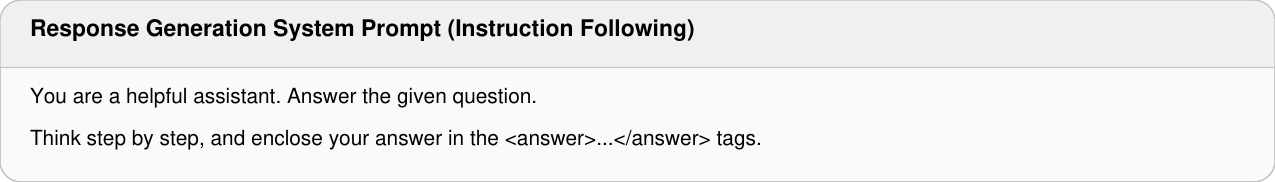}
    \caption{Response Generation System Prompt (Instruction Following).}
    \label{fig:response_generation_system_prompt_(instruction_following)}
\end{figure}

\begin{figure}
    \centering
    \includegraphics[width=\linewidth]{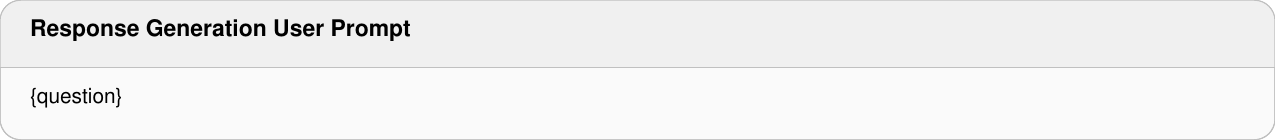}
    \caption{Response Generation User Prompt.}
    \label{fig:response_generation_user_prompt}
\end{figure}

\begin{figure}
    \centering
    \includegraphics[width=\linewidth]{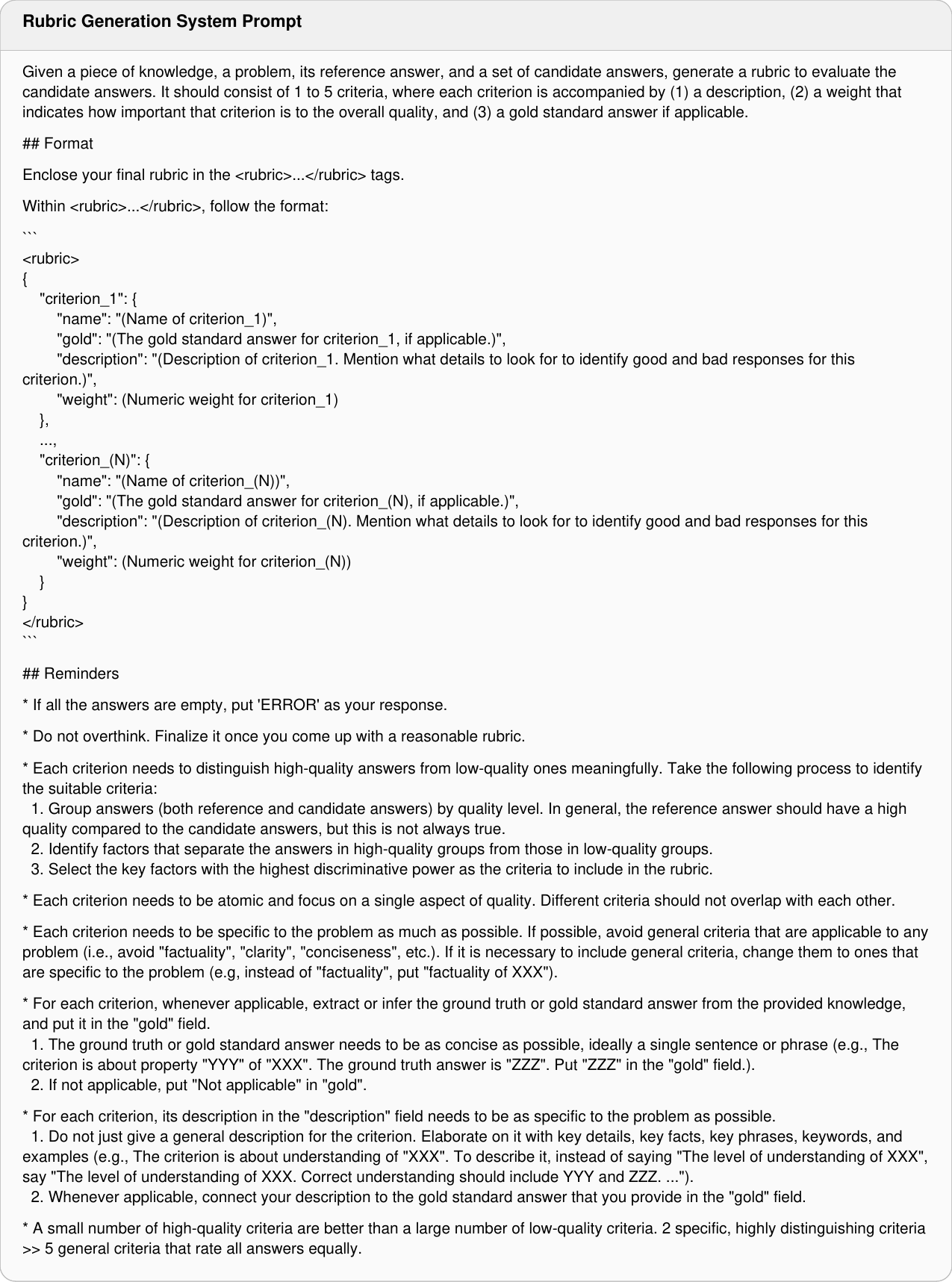}
    \caption{Rubric Generation System Prompt.}
    \label{fig:rubric_generation_system_prompt}
\end{figure}

\begin{figure}
    \centering
    \includegraphics[width=\linewidth]{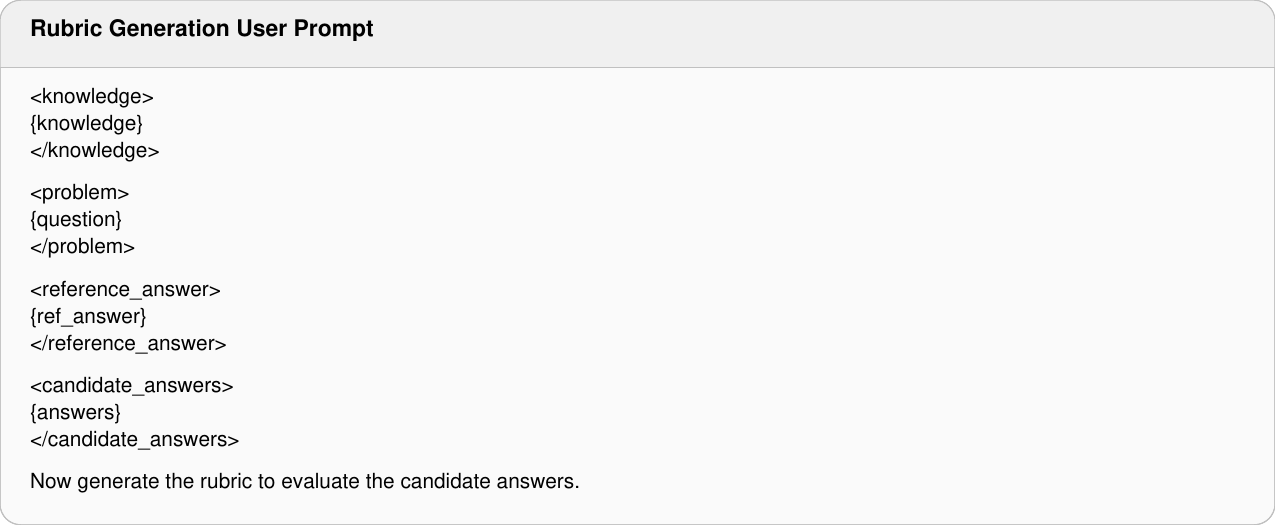}
    \caption{Rubric Generation User Prompt.}
    \label{fig:rubric_generation_user_prompt}
\end{figure}

\begin{figure}
    \centering
    \includegraphics[width=\linewidth]{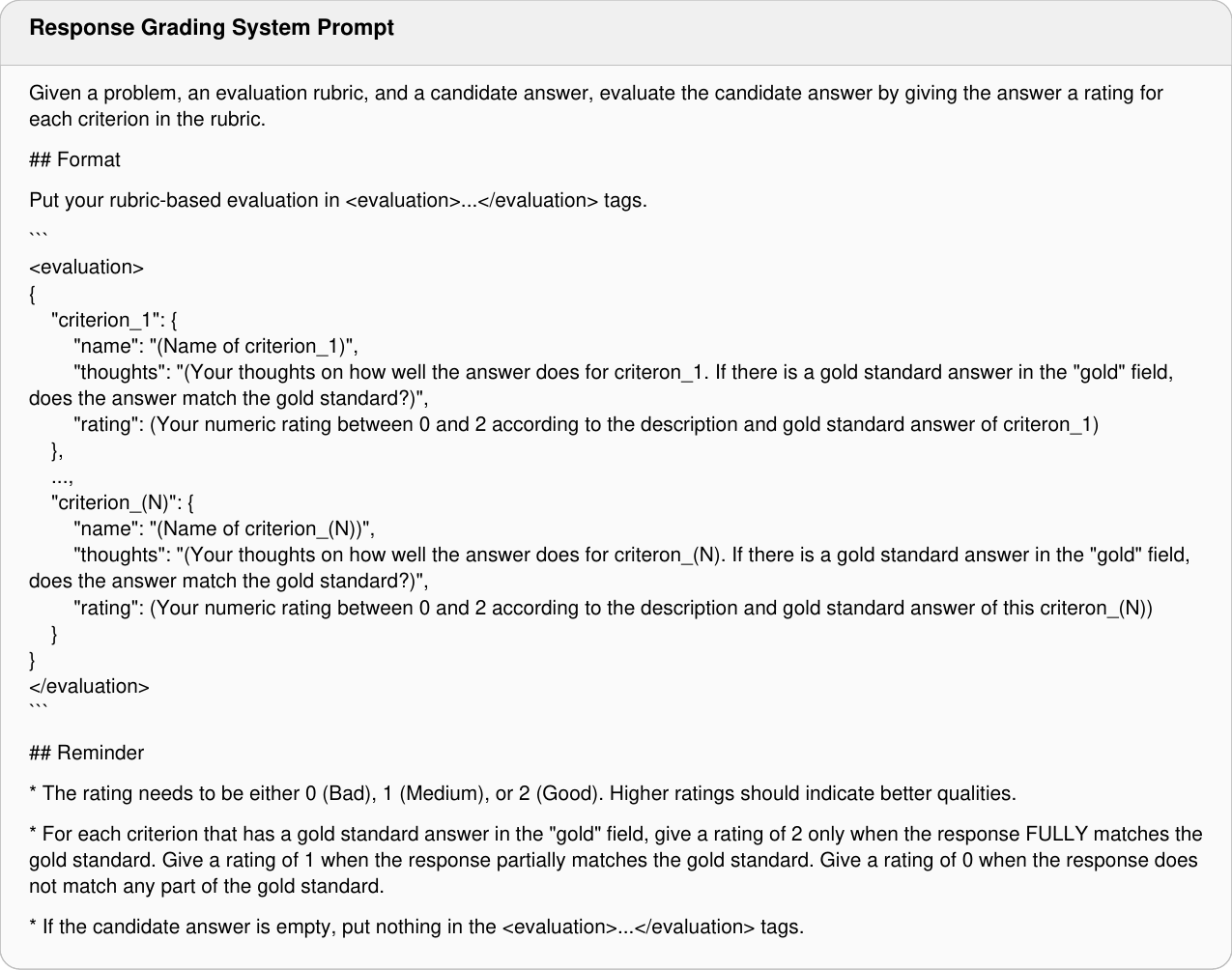}
    \caption{Response Grading System Prompt.}
    \label{fig:response_grading_system_prompt}
\end{figure}

\begin{figure}
    \centering
    \includegraphics[width=\linewidth]{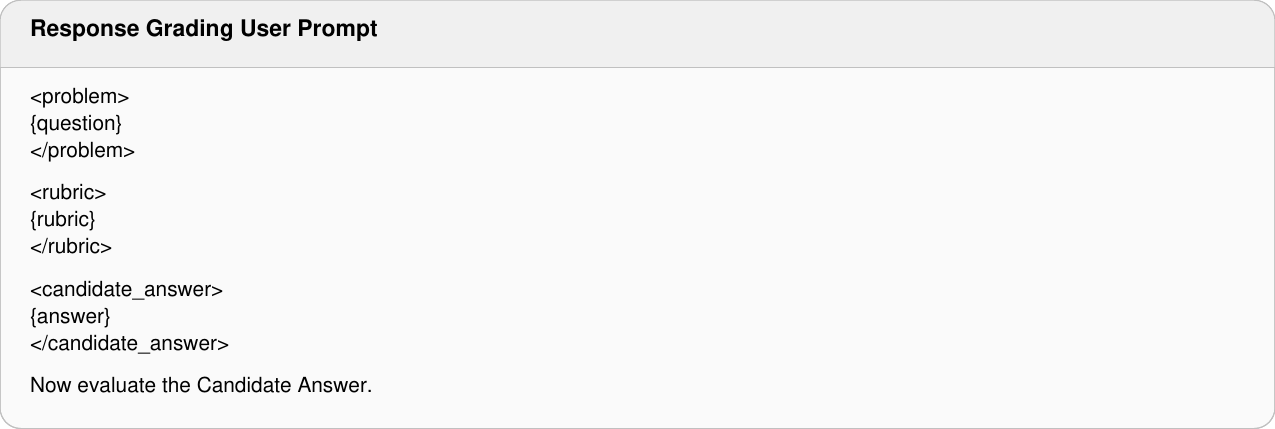}
    \caption{Response Grading User Prompt.}
    \label{fig:response_grading_user_prompt}
\end{figure}

We show the prompts in Figure~\ref{fig:query_synthesis_system_prompt_(healthcare_qa)} through~\ref{fig:response_grading_user_prompt}.

\label{app:prompts}

%% file: checklist.tex
\section*{NeurIPS Paper Checklist}

\begin{enumerate}

\item {\bf Claims}
    \item[] Question: Do the main claims made in the abstract and introduction accurately reflect the paper's contributions and scope?
    \item[] Answer: \answerYes{} % Replace by \answerYes{}, \answerNo{}, or \answerNA{}.
    \item[] Justification:  In our abstract and introduction, we clearly mention our proposed framework and the results that it achieves.

\item {\bf Limitations}
    \item[] Question: Does the paper discuss the limitations of the work performed by the authors?
    \item[] Answer: \answerYes{} % Replace by \answerYes{}, \answerNo{}, or \answerNA{}.
    \item[] Justification: We discuss the limitations in Section~\ref{sec:limitations_and_future_directions}.

\item {\bf Theory assumptions and proofs}
    \item[] Question: For each theoretical result, does the paper provide the full set of assumptions and a complete (and correct) proof?
    \item[] Answer: \answerNA{} % Replace by \answerYes{}, \answerNo{}, or \answerNA{}.
    \item[] Justification: The paper does not include new theoretical results.

    \item {\bf Experimental result reproducibility}
    \item[] Question: Does the paper fully disclose all the information needed to reproduce the main experimental results of the paper to the extent that it affects the main claims and/or conclusions of the paper (regardless of whether the code and data are provided or not)?
    \item[] Answer: \answerYes{} % Replace by \answerYes{}, \answerNo{}, or \answerNA{}.
    \item[] Justification: We discuss our method in details in \S~\ref{sec:method} and show the experiment details in \S~\ref{sec:setup} and Appendix~\ref{app:hyperparameters}.

\item {\bf Open access to data and code}
    \item[] Question: Does the paper provide open access to the data and code, with sufficient instructions to faithfully reproduce the main experimental results, as described in supplemental material?
    \item[] Answer: \answerYes{} % Replace by \answerYes{}, \answerNo{}, or \answerNA{}.
    \item[] Justification: We attach the program and instructions in the submission.

\item {\bf Experimental setting/details}
    \item[] Question: Does the paper specify all the training and test details (e.g., data splits, hyperparameters, how they were chosen, type of optimizer) necessary to understand the results?
    \item[] Answer: \answerYes{} % Replace by \answerYes{}, \answerNo{}, or \answerNA{}.
    \item[] Justification: We show the experiment details in \S~\ref{sec:setup} and Appendix~\ref{app:hyperparameters}

\item {\bf Experiment statistical significance}
    \item[] Question: Does the paper report error bars suitably and correctly defined or other appropriate information about the statistical significance of the experiments?
    \item[] Answer: \answerYes{} % Replace by \answerYes{}, \answerNo{}, or \answerNA{}.
    \item[] Justification: We report the statistics of our datasets in \S~\ref{sec:stat} and error bars for main results in \S~\ref{sec:main_results} and ranking analysis in ~\S~\ref{sec:ranking_analysis}.

\item {\bf Experiments compute resources}
    \item[] Question: For each experiment, does the paper provide sufficient information on the computer resources (type of compute workers, memory, time of execution) needed to reproduce the experiments?
    \item[] Answer: \answerYes{} % Replace by \answerYes{}, \answerNo{}, or \answerNA{}.
    \item[] Justification: We report the compute resources in Appendix~\ref{app:compute}.
    
\item {\bf Code of ethics}
    \item[] Question: Does the research conducted in the paper conform, in every respect, with the NeurIPS Code of Ethics \url{https://neurips.cc/public/EthicsGuidelines}?
    \item[] Answer: \answerYes{} % Replace by \answerYes{}, \answerNo{}, or \answerNA{}.
    \item[] Justification: Our work adheres to the set guidelines.

\item {\bf Broader impacts}
    \item[] Question: Does the paper discuss both potential positive societal impacts and negative societal impacts of the work performed?
    \item[] Answer: \answerNA{} % Replace by \answerYes{}, \answerNo{}, or \answerNA{}.
    \item[] Justification: Since our work is foundational and only gives a preliminary prototype, there is no immediate deployment use cases.
    
\item {\bf Safeguards}
    \item[] Question: Does the paper describe safeguards that have been put in place for responsible release of data or models that have a high risk for misuse (e.g., pre-trained language models, image generators, or scraped datasets)?
    \item[] Answer: \answerNA{} % Replace by \answerYes{}, \answerNo{}, or \answerNA{}.
    \item[] Justification: The paper has no such risks since our tasks are not tied to use cases with high risks.

\item {\bf Licenses for existing assets}
    \item[] Question: Are the creators or original owners of assets (e.g., code, data, models), used in the paper, properly credited and are the license and terms of use explicitly mentioned and properly respected?
    \item[] Answer: \answerYes{} % Replace by \answerYes{}, \answerNo{}, or \answerNA{}.
    \item[] Justification: We cite the pretraining corpus and models that we use. They are open-source artifacts, and we adhere to their usage guidelines.

\item {\bf New assets}
    \item[] Question: Are new assets introduced in the paper well documented and is the documentation provided alongside the assets?
    \item[] Answer: \answerYes{} % Replace by \answerYes{}, \answerNo{}, or \answerNA{}.
    \item[] Justification: We provide usage instructions for our program in the attachment.

\item {\bf Crowdsourcing and research with human subjects}
    \item[] Question: For crowdsourcing experiments and research with human subjects, does the paper include the full text of instructions given to participants and screenshots, if applicable, as well as details about compensation (if any)? 
    \item[] Answer: \answerNA{} % Replace by \answerYes{}, \answerNo{}, or \answerNA{}.
    \item[] Justification: The paper does not involve crowdsourcing or research with human subjects.

\item {\bf Institutional review board (IRB) approvals or equivalent for research with human subjects}
    \item[] Question: Does the paper describe potential risks incurred by study participants, whether such risks were disclosed to the subjects, and whether Institutional Review Board (IRB) approvals (or an equivalent approval/review based on the requirements of your country or institution) were obtained?
    \item[] Answer: \answerNA{} % Replace by \answerYes{}, \answerNo{}, or \answerNA{}.
    \item[] Justification: The paper does not involve crowdsourcing or research with human subjects.

\item {\bf Declaration of LLM usage}
    \item[] Question: Does the paper describe the usage of LLMs if it is an important, original, or non-standard component of the core methods in this research? Note that if the LLM is used only for writing, editing, or formatting purposes and does \emph{not} impact the core methodology, scientific rigor, or originality of the research, declaration is not required.
    %this research? 
    \item[] Answer: \answerNA{} % Replace by \answerYes{}, \answerNo{}, or \answerNA{}.
    \item[] Justification: We do not use LLM for the core method development or any experiments.

\end{enumerate}

%% file: custom.bib
@misc{gdm2025,
    title = {Advanced version of Gemini with Deep Think officially achieves gold-medal standard at the International Mathematical Olympiad},
    author = {Thang Luong and Edward Lockhart},
    year = {2025},
}

@article{agentsurvey2025,
    author = {Aske Plaat and Max van Duijn and Niki van Stein and Mike Preuss and Peter van der Putten and Kees Joost Batenburg},
    title = {Agentic Large Language Models, a survey},
    journal = {arXiv preprint arXiv:2503.23037},
    year = {2025},
}

@inproceedings{villalobos2024,
    author = {Pablo Villalobos and Anson Ho and Jaime Sevilla and Tamay Besiroglu and Lennart Heim and Marius Hobbhahn},
    title = {Will we run out of data? Limits of LLM scaling based on human-generated data},
    booktitle = {Proceedings of the 41st International Conference on Machine Learning},
    year = {2024}
}

@article{absolutezero2025,
    author = {Andrew Zhao and Yiran Wu and Yang Yue and Tong Wu and Quentin Xu and Yang Yue and Matthieu Lin and Shenzhi Wang and Qingyun Wu and Zilong Zheng and Gao Huang},
    title = {Absolute Zero: Reinforced Self-play Reasoning with Zero Data},
    journal = {arXiv preprint arXiv:2505.03335},
    year = {2025},
}

@article{zhou2025self,
    author = {Yifei Zhou and Sergey Levine and Jason Weston and Xian Li and Sainbayar Sukhbaatar},
    title = {Self-Challenging Language Model Agents},
    journal = {arXiv preprint arXiv:2506.01716},
    year = {2025}
}

@article{chen2025,
    author = {Jiaqi Chen and Bang Zhang and Ruotian Ma and Peisong Wang and Xiaodan Liang and Zhaopeng Tu and Xiaolong Li and Kwan-Yee K. Wong},
    title = {SPC: Evolving Self-Play Critic via Adversarial Games for LLM Reasoning},
    journal = {arXiv preprint arXiv:2504.19162},
    year = {2025}
}

@article{huang2025rzero,
    author = {Chengsong Huang and Wenhao Yu and Xiaoyang Wang and Hongming Zhang and Zongxia Li and Ruosen Li and Jiaxin Huang and Haitao Mi and Dong Yu},
    title = {R-Zero: Self-Evolving Reasoning LLM from Zero Data},
    journal = {arXiv preprint arXiv:2508.05004},
    year = {2025}
}

@article{liu2025,
    author = {Bo Liu and Chuanyang Jin and Seungone Kim and Weizhe Yuan and Wenting Zhao and Ilia Kulikov and Xian Li and Sainbayar Sukhbaatar and Jack Lanchantin and Jason Weston},
    title = {SPICE: Self-Play In Corpus Environments Improves Reasoning},
    journal = {arXiv preprint arXiv:2510.24684},
    year = {2025}
}

@article{cheng2024,
    author = {Pengyu Cheng and Tianhao Hu and Han Xu and Zhisong Zhang and Zheng Yuan and Yong Dai and Lei Han and Nan Du and Xiaolong Li},
    title = {Self-playing Adversarial Language Game Enhances LLM Reasoning},
    journal = {arXiv preprint arXiv:2404.10642},
    year = {2024}
}

@article{zhang2025,
    author = {Zhengxin Zhang and Chengyu Huang and Aochong Oliver Li and Claire Cardie},
    title = {Better LLM Reasoning via Dual-Play},
    journal = {arXiv preprint arXiv:2511.11881},
    year = {2025}
}

@inproceedings{huang2023,
    author = {Jiaxin Huang and Shixiang Gu and Le Hou and Yuexin Wu and Xuezhi Wang and Hongkun Yu and Jiawei Han},
    title = {Large Language Models Can Self-Improve},
    booktitle = {Proceedings of the 2023 Conference on Empirical Methods in Natural Language Processing},
    year = {2023}
}

@inproceedings{yuan2024,
    author = {Weizhe Yuan and Richard Yuanzhe Pang and Kyunghyun Cho and Xian Li and Sainbayar Sukhbaatar and Jing Xu and Jason Weston},
    title = {Self-Rewarding Language Models},
    booktitle = {Proceedings of the 41st International Conference on Machine Learning},
    year = {2024}
}

@article{shao2025,
    author = {Rulin Shao and Akari Asai and Shannon Zejiang Shen and Hamish Ivison and Varsha Kishore and Jingming Zhuo and Xinran Zhao and Molly Park and Samuel G. Finlayson and David Sontag and Tyler Murray and Sewon Min and Pradeep Dasigi and Luca Soldaini and Faeze Brahman and Wen-tau Yih and Tongshuang Wu and Luke Zettlemoyer and Yoon Kim and Hannaneh Hajishirzi and Pang Wei Koh},
    title = {DR Tulu: Reinforcement Learning with Evolving Rubrics for Deep Research},
    journal = {arXiv preprint arXiv:2511.19399},
    year = {2025},
}

@article{chung2022,
    author = {Hyung Won Chung and Le Hou and Shayne Longpre and Barret Zoph and Yi Tay and William Fedus and Yunxuan Li and Xuezhi Wang and Mostafa Dehghani and Siddhartha Brahma and Albert Webson and Shixiang Shane Gu and Zhuyun Dai and Mirac Suzgun and Xinyun Chen and Aakanksha Chowdhery and Alex Castro-Ros and Marie Pellat and Kevin Robinson and Dasha Valter and Sharan Narang and Gaurav Mishra and Adams Yu and Vincent Zhao and Yanping Huang and Andrew Dai and Hongkun Yu and Slav Petrov and Ed H. Chi and Jeff Dean and Jacob Devlin and Adam Roberts and Denny Zhou and Quoc V. Le and Jason Wei},
    title = {Scaling Instruction-Finetuned Language Models},
    journal = {arXiv preprint arXiv:2210.11416},
    year = {2022},
}

@inproceedings{zheng2023,
    author = {Lianmin Zheng and Wei-Lin Chiang and Ying Sheng and Siyuan Zhuang and Zhanghao Wu and Yonghao Zhuang and Zi Lin and Zhuohan Li and Dacheng Li and Eric P. Xing and Hao Zhang and Joseph E. Gonzalez and Ion Stoica},
    title = {Judging LLM-as-a-Judge with MT-Bench and Chatbot Arena},
    booktitle = {Advances in Neural Information Processing Systems, Datasets and Benchmarks Track},
    year = {2023}
}

@misc{Gokaslan2019OpenWeb,
    title={OpenWebText Corpus},
    author={Gokaslan, Aaron and Cohen, Vanya and Pavlick, Ellie and Tellex, Stefanie},
    howpublished={\url{http://Skylion007.github.io/OpenWebTextCorpus}},
    year={2019}
}

@inproceedings{gap2025,
    author = {Yuda Song and Hanlin Zhang and Carson Eisenach and Sham Kakade and Dean Foster and Udaya Ghai},
    title = {Mind the Gap: Examining the Self-Improvement Capabilities of Large Language Models},
    booktitle = {Proceedings of International Conference on Learning Representations},
    year = {2025}
}

@article{gunjal2025,
    author = {Anisha Gunjal and Anthony Wang and Elaine Lau and Vaskar Nath and Yunzhong He and Bing Liu and Sean Hendryx},
    title = {Rubrics as Rewards: Reinforcement Learning Beyond Verifiable Domains},
    journal = {arXiv preprint arXiv:2507.17746},
    year = {2025},
}

@inproceedings{viswanathan2025,
    author = {Vijay Viswanathan and Yanchao Sun and Shuang Ma and Xiang Kong and Meng Cao and Graham Neubig and Tongshuang Wu},
    title = {Checklists Are Better Than Reward Models For Aligning Language Models},
    booktitle = {Advances in Neural Information Processing Systems},
    year = {2025},
}

@article{huang2025rlra,
    author = {Zenan Huang and Yihong Zhuang and Guoshan Lu and Zeyu Qin and Haokai Xu and Tianyu Zhao and Ru Peng and Jiaqi Hu and Zhanming Shen and Xiaomeng Hu and Xijun Gu and Peiyi Tu and Jiaxin Liu and Wenyu Chen and Yuzhuo Fu and Zhiting Fan and Yanmei Gu and Yuanyuan Wang and Zhengkai Yang and Jianguo Li and Junbo Zhao},
    title = {Reinforcement Learning with Rubric Anchors},
    journal = {arXiv preprint arXiv:2508.12790},
    year = {2025},
}

@article{rezaei2025,
    author = {MohammadHossein Rezaei and Robert Vacareanu and Zihao Wang and Clinton Wang and Bing Liu and Yunzhong He and Afra Feyza Akyürek},
    title = {Online Rubrics Elicitation from Pairwise Comparisons},
    journal = {arXiv preprint arXiv:2510.07284},
    year = {2025},
}

@article{zhou2025,
    author = {Yang Zhou and Sunzhu Li and Shunyu Liu and Wenkai Fang and Kongcheng Zhang and Jiale Zhao and Jingwen Yang and Yihe Zhou and Jianwei Lv and Tongya Zheng and Hengtong Lu and Wei Chen and Yan Xie and Mingli Song},
    title = {Breaking the Exploration Bottleneck: Rubric-Scaffolded Reinforcement Learning for General LLM Reasoning},
    journal = {arXiv preprint arXiv:2508.16949},
    year = {2025},
}

@article{he2025,
    author = {Yun He and Wenzhe Li and Hejia Zhang and Songlin Li and Karishma Mandyam and Sopan Khosla and Yuanhao Xiong and Nanshu Wang and Xiaoliang Peng and Beibin Li and Shengjie Bi and Shishir G. Patil and Qi Qi and Shengyu Feng and Julian Katz-Samuels and Richard Yuanzhe Pang and Sujan Gonugondla and Hunter Lang and Yue Yu and Yundi Qian and Maryam Fazel-Zarandi and Licheng Yu and Amine Benhalloum and Hany Awadalla and Manaal Faruqui},
    title = {AdvancedIF: Rubric-Based Benchmarking and Reinforcement Learning for Advancing LLM Instruction Following},
    journal = {arXiv preprint arXiv:2511.10507},
    year = {2025},
}

@article{bi2025,
    author = {Baolong Bi and Shenghua Liu and Yiwei Wang and Siqian Tong and Lingrui Mei and Yuyao Ge and Yilong Xu and Jiafeng Guo and Xueqi Cheng},
    title = {Reward and Guidance through Rubrics: Promoting Exploration to Improve Multi-Domain Reasoning},
    journal = {arXiv preprint arXiv:2508.16949},
    year = {2025},
}

@inproceedings{dpo2023,
    author = {Rafael Rafailov and Archit Sharma and Eric Mitchell and Stefano Ermon and Christopher D. Manning and Chelsea Finn},
    title = {Direct Preference Optimization: Your Language Model is Secretly a Reward Model},
    booktitle = {arXiv preprint arXiv:2305.18290},
    year = {2023},
}

@inproceedings{huang2025,
    author = {Chengyu Huang and Tanya Goyal},
    title = {DCRM: A Heuristic to Measure Response Pair Quality in Preference Optimization},
    booktitle = {Findings of the Association for Computational Linguistics: EMNLP},
    year = {2025},
}

@article{arora2025,
    author = {Rahul K. Arora and Jason Wei and Rebecca Soskin Hicks and Preston Bowman and Joaquin Quiñonero-Candela and Foivos Tsimpourlas and Michael Sharman and Meghan Shah and Andrea Vallone and Alex Beutel and Johannes Heidecke and Karan Singhal},
    title = {HealthBench: Evaluating Large Language Models Towards Improved Human Health},
    journal = {arXiv preprint arXiv:2505.08775},
    year = {2025}
}

@misc{creative-writing-bench-v3,
  author = {Samuel J Paech},
  title = {EQ-Bench Creative Writing Benchmark v3},
  year = {2025},
  publisher = {GitHub},
  journal = {GitHub repository},
  howpublished = {\url{https://github.com/EQ-bench/creative-writing-bench}}
}

@article{zhou2023instructionfollowing,
  title={Instruction-Following Evaluation for Large Language Models},
  author={Jeffrey Zhou and Tianjian Lu and Swaroop Mishra and Siddhartha Brahma and Sujoy Basu and Yi Luan and Denny Zhou and Le Hou},
  journal={arXiv preprint arXiv:2311.07911},
  year={2023},
}

@article{ppo2017,
    author = {John Schulman and Filip Wolski and Prafulla Dhariwal and Alec Radford and Oleg Klimov},
    title = {Proximal Policy Optimization Algorithms},
    journal = {arXiv preprint arXiv:1707.06347},
    year = {2017}
}

@inproceedings{adamw2019,
    author = {Ilya Loshchilov and Frank Hutter},
    title = {Decoupled Weight Decay Regularization},
    booktitle = {Proceedings of International Conference on Learning Representations},
    year = {2019}
}

@inproceedings{maslenkova2025,
    author = {Svetlana Maslenkova and Clement Christophe and Marco AF Pimentel and Tathagata Raha and Muhammad Umar Salman and Ahmed Al Mahrooqi and Avani Gupta and Shadab Khan and Ronnie Rajan and Praveenkumar Kanithi},
    title = {Building Trust in Clinical LLMs: Bias Analysis and Dataset Transparency},
    booktitle = {Proceedings of the 2025 Conference on Empirical Methods in Natural Language Processing},
    year = {2025}
}

@misc{bookabstracts,
    title = {Book Titles and Abstracts},
    author={Skelebor},
    howpublished={\url{https://huggingface.co/datasets/Skelebor/book_titles_and_descriptions}},
    year={2022}
}

@article{qwen252024,
    author = {Qwen Team},
    title = {Qwen2.5 Technical Report},
    journal = {arXiv preprint arXiv:2412.15115},
    year = {2024}
}

@article{gpt42023,
    author = {OpenAI Team},
    title = {GPT-4 Technical Report},
    journal = {arXiv preprint arXiv:2303.08774},
    year = {2023}
}

@inproceedings{rein2024gpqa,
      title={{GPQA}: A Graduate-Level Google-Proof Q\&A Benchmark},
      author={David Rein and Betty Li Hou and Asa Cooper Stickland and Jackson Petty and Richard Yuanzhe Pang and Julien Dirani and Julian Michael and Samuel R. Bowman},
      booktitle={First Conference on Language Modeling},
      year={2024},
      url={https://openreview.net/forum?id=Ti67584b98}
}

@article{cobbe2021gsm8k,
  title={Training Verifiers to Solve Math Word Problems},
  author={Cobbe, Karl and Kosaraju, Vineet and Bavarian, Mohammad and Chen, Mark and Jun, Heewoo and Kaiser, Lukasz and Plappert, Matthias and Tworek, Jerry and Hilton, Jacob and Nakano, Reiichiro and Hesse, Christopher and Schulman, John},
  journal={arXiv preprint arXiv:2110.14168},
  year={2021}
}

@inproceedings{hendrycksmath2021,
  title={Measuring Mathematical Problem Solving With the MATH Dataset},
  author={Dan Hendrycks and Collin Burns and Saurav Kadavath and Akul Arora and Steven Basart and Eric Tang and Dawn Song and Jacob Steinhardt},
  booktitle={Advances in Neural Information Processing Systems},
  year={2021}
}

@article{kwiatkowski2019,
    author = {Tom Kwiatkowski and Jennimaria Palomaki and Olivia Redfield and Michael Collins and Ankur Parikh and Chris Alberti and Danielle Epstein and Illia Polosukhin and Jacob Devlin and Kenton Lee and Kristina Toutanova and Llion Jones and Matthew Kelcey and Ming-Wei Chang and Andrew M. Dai and Jakob Uszkoreit and Quoc Le and Slav Petrov},
    title = {Natural Questions: A Benchmark for Question Answering Research},
    journal = {Transactions of the Association for Computational Linguistics, Volume 7},
    year = {2019}
}

@article{2017arXivtriviaqa,
       author = {{Joshi}, Mandar and {Choi}, Eunsol and {Weld},
                 Daniel and {Zettlemoyer}, Luke},
        title = "{triviaqa: A Large Scale Distantly Supervised Challenge Dataset for Reading Comprehension}",
      journal = {arXiv e-prints},
         year = 2017,
          eid = {arXiv:1705.03551},
        pages = {arXiv:1705.03551},
archivePrefix = {arXiv},
       eprint = {1705.03551},
}

@inproceedings{stephanie2022,
    author = {Stephanie Lin and Jacob Hilton and Owain Evans},
    title = {TruthfulQA: Measuring How Models Mimic Human Falsehoods},
    booktitle = {Proceedings of the 60th Annual Meeting of the Association for Computational Linguistics (Volume 1: Long Papers)},
    year = {2022}
}

@inproceedings{mmlupro2024,
    author = {Yubo Wang and Xueguang Ma and Ge Zhang and Yuansheng Ni and Abhranil Chandra and Shiguang Guo and Weiming Ren and Aaran Arulraj and Xuan He and Ziyan Jiang and Tianle Li and Max Ku and Kai Wang and Alex Zhuang and Rongqi Fan and Xiang Yue and Wenhu Chen},
    title = {MMLU-Pro: A More Robust and Challenging Multi-Task Language Understanding Benchmark},
    booktitle = {Advances in Neural Information Processing Systems, Datasets and Benchmarks Track},
    year = {2024}
}

@article{singh2026,
    author = {Harman Singh and Xiuyu Li and Kusha Sareen and Monishwaran Maheswaran and Sijun Tan and Xiaoxia Wu and Junxiong Wang and Alpay Ariyak and Qingyang Wu and Samir Khaki and Rishabh Tiwari and Long Lian and Yucheng Lu and Boyi Li and Alane Suhr and Ben Athiwaratkun and Kurt Keutzer},
    title = {V1: Unifying Generation and Self-Verification for Parallel Reasoners},
    journal = {arXiv preprint arXiv:2603.04304},
    year = {2026}
}

@inproceedings{ren2025,
    author = {Yi Ren and Danica J. Sutherland},
    title = {Learning Dynamics of LLM Finetuning},
    booktitle = {Proceedings of International Conference on Learning Representations},
    year = {2025}
}

@article{tang2024,
    author = {Yunhao Tang and Daniel Zhaohan Guo and Zeyu Zheng and Daniele Calandriello and Yuan Cao and Eugene Tarassov and Rémi Munos and Bernardo Ávila Pires and Michal Valko and Yong Cheng, Will Dabney},
    title = {Understanding the performance gap between online and offline alignment algorithms},
    journal = {arXiv preprint arXiv:2405.08448},
    year = {2024}
}

@article{deepseek2025,
    author = {DeepSeek-AI},
    title = {DeepSeek-V3.2: Pushing the Frontier of Open Large Language Models},
    journal = {arXiv preprint:arXiv:2512.02556},
    year = {2025}
}

@article{karl2026,
    author = {Databricks AI Research},
    title = {KARL: Knowledge Agentsvial Reinforcement Learning},
    journal = {arXiv preprint:arXiv:2603.05218},
    year = {2026}
}

@inproceedings{dineen2025,
    author = {Jacob Dineen and Aswin Rrv and Qin Liu and Zhikun Xu and Xiao Ye and Ming Shen and Zhaonan Li and Shijie Lu and Chitta Baral and Muhao Chen and Ben Zhou},
    title = {QA-LIGN: Aligning LLMs through Constitutionally Decomposed QA},
    booktitle = {Proceedings of Findings of the Association for Computational Linguistics: EMNLP 2025},
    year = {2025}
}

@InProceedings{pmlr-v174-pal22a,
  title = 	  {MedMCQA: A Large-scale Multi-Subject Multi-Choice Dataset for Medical domain Question Answering},
  author =    {Pal, Ankit and Umapathi, Logesh Kumar and Sankarasubbu, Malaikannan},
  booktitle = {Proceedings of the Conference on Health, Inference, and Learning},
  pages = 	 {248--260},
  year = 	 {2022},
  editor = 	 {Flores, Gerardo and Chen, George H and Pollard, Tom and Ho, Joyce C and Naumann, Tristan},
  volume = 	 {174},
  series = 	 {Proceedings of Machine Learning Research},
  month = 	 {07--08 Apr},
  publisher =    {PMLR},
  url = 	 {https://proceedings.mlr.press/v174/pal22a.html}
}

@misc{lighteval,
  author = {Habib, Nathan and Fourrier, Clémentine and Kydlíček, Hynek and Wolf, Thomas and Tunstall, Lewis},
  title = {LightEval: A lightweight framework for LLM evaluation},
  year = {2023},
  version = {0.11.0},
  url = {https://github.com/huggingface/lighteval}
}
